\documentclass{article}
\usepackage{ijcai24}

\usepackage{times}
\usepackage{soul}
\usepackage{url}
\usepackage[hidelinks]{hyperref}
\usepackage[utf8]{inputenc}
\usepackage[small]{caption}
\usepackage{graphicx}
\usepackage{amsmath}
\usepackage{amsthm}
\usepackage{booktabs}
\usepackage{algorithm}
\usepackage{algorithmic}
\usepackage[switch]{lineno}
\usepackage{xcolor}
\usepackage[normalem]{ulem}
\useunder{\uline}{\ul}{}

\usepackage{multirow}
\usepackage{amssymb}
\usepackage{tabularx}
\usepackage{hyperref}
\usepackage[numbers,sort&compress]{natbib}
\definecolor{ccr}{RGB}{10,110,150}
\hypersetup{
    colorlinks=true,
    linkcolor=ccr,
    anchorcolor=ccr,
    citecolor=ccr,
    urlcolor=ccr
}
\usepackage[numbers]{natbib}
\setcitestyle{comma}

\title{From Individual to Society: A Survey on Social Simulation Driven by Large Language Model-based Agents}

\author{
Xinyi Mou$^1$\thanks{These authors contributed equally.}\and
Xuanwen Ding$^2$\footnotemark[1]\and
Qi He$^1$\footnotemark[1]\and
Liang Wang$^3$\footnotemark[1]\and\\
Jingcong Liang$^1$\and
Xinnong Zhang$^1$\and
Libo Sun$^1$\and
Jiayu Lin$^1$\and\\
Jie Zhou$^2$\and
Xuanjing Huang$^1$\And
Zhongyu Wei$^{1,4}$\thanks{Corresponding author.}
\\
\affiliations
$^1$Fudan University\\
$^2$East China Normal University\\
$^3$Harbin Institute of Technology, Shenzhen\\
$^4$Shanghai Innovation Institute\\
\emails
 \href{mailto:zywei@fudan.edu.cn}{\texttt{zywei@fudan.edu.cn}}
}

\begin{document}

\maketitle

\begin{abstract}
Traditional sociological research often relies on human participation, which, though effective, is expensive, challenging to scale, and with ethical concerns. Recent advancements in large language models (LLMs) highlight their potential to simulate human behavior, enabling the replication of individual responses and facilitating studies on many interdisciplinary studies. In this paper, we conduct a comprehensive survey of this field, illustrating the recent progress in simulation driven by LLM-empowered agents. We categorize the simulations into three types: (1) \textit{Individual Simulation}, which mimics specific individuals or demographic groups; (2) \textit{Scenario Simulation}, where multiple agents collaborate to achieve goals within specific contexts; and (3) \textit{Society Simulation}, which models interactions within agent societies to reflect the complexity and variety of real-world dynamics. These simulations follow a progression, ranging from detailed individual modeling to large-scale societal phenomena. We provide a detailed discussion of each simulation type, including the architecture or key components of the simulation, the classification of objectives or scenarios and the evaluation method. Afterward, we summarize commonly used datasets and benchmarks. Finally, we discuss the trends across these three types of simulation. A repository for the related sources is at {\url{https://github.com/FudanDISC/SocialAgent}}.
\end{abstract}

\section{Introduction}
Social science investigates human behavior and social structures to understand how societies function. Traditional sociological research heavily relies on human participation to conduct experiments and gather data. Questionnaires~\cite{granovetter1973strength,katz2015social} and psychological experiments~\cite{asch1951effects,milgram1963behavioral} are commonly used to test theoretical hypotheses, understand social phenomena, and predict collective outcomes. While these methods can provide highly authentic data, they are expensive, challenging to scale, and involve certain ethical risks.

Recently, large language models (LLMs) have demonstrated impressive capabilities in human-level reasoning and planning~\cite{wei2022chain,kojima2022large,xi2023rise,yao2024tree,Wang_2024}. They can perceive the environment, make decisions, and take corresponding actions, showcasing their potential as autonomous agents that can serve as human substitutes. In appropriate settings, LLM-driven agents can accurately simulate responses from corresponding individuals by leveraging their role-playing abilities~\cite{shao2023characterllmtrainableagentroleplaying,chen2024persona}, a property known as algorithmic fidelity~\cite{Argyle_2023,chaudhary2024large}. This characteristic makes LLM-driven agents highly valuable in simulating human behavior. By reproducing individual response patterns in specific scenarios, LLM-driven agents help researchers to better understand, validate, and predict human reactions.

\begin{figure*}[!t]
    \centering
    \includegraphics[width=\linewidth]{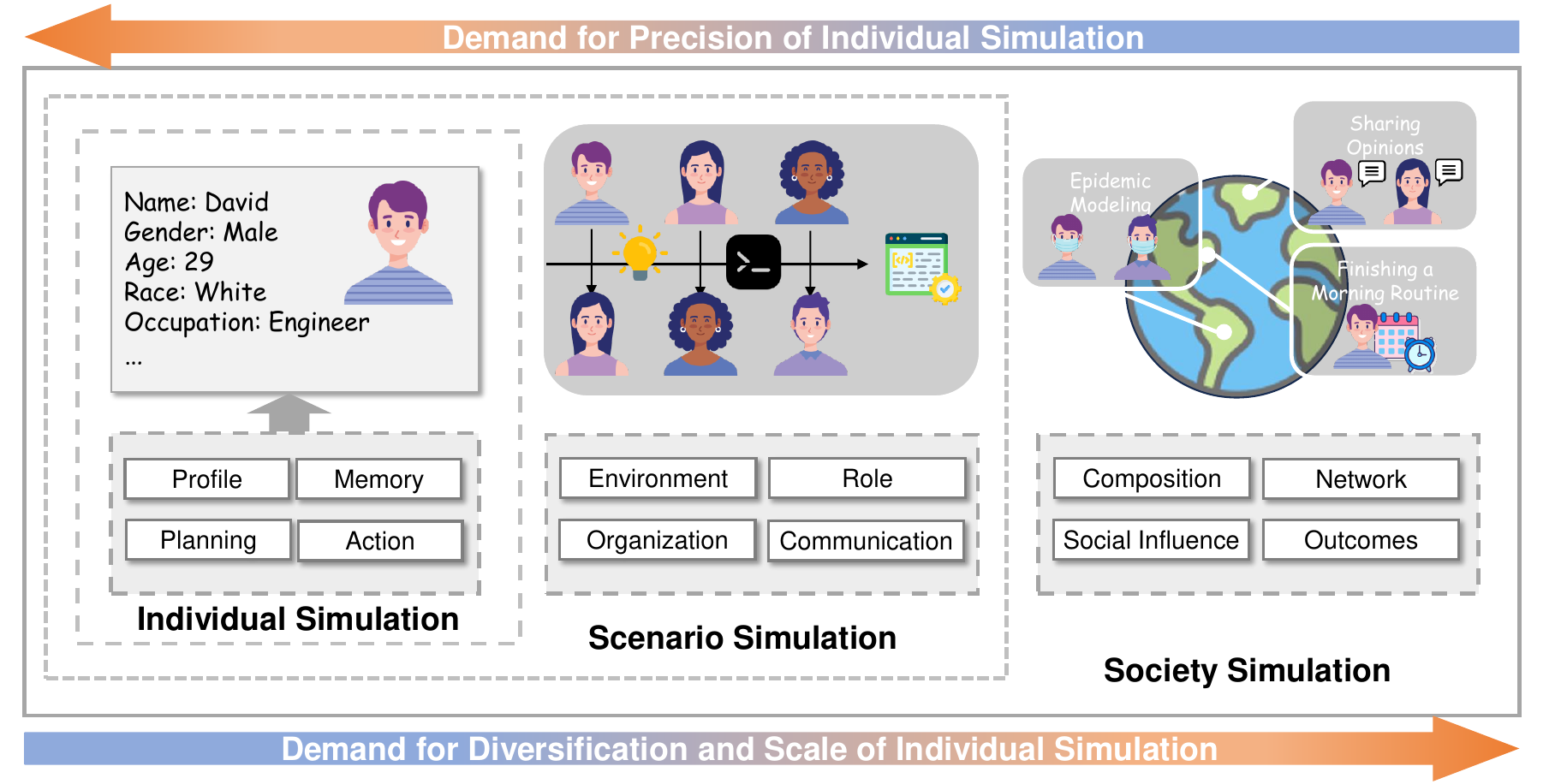}
    \caption{Illustration of simulations empowered by LLM-driven agents. We categorize the simulations into individual simulation, scenario simulation and society simulation. From left to right, the diversity and scale of individual modeling generally increase. Conversely, from right to left, the granularity of individual modeling becomes more refined.}
    \label{fig:intro}
\end{figure*}

Just as individuals do not exist independently within society, in addition to separate individual agents, interactions between multiple agents have also been widely studied to solve specific problems or simulate complex dynamics in the real world~\cite{guo2024large,gao2024large}. On one hand, LLMs can be specialized as agents with detailed knowledge and skills, leveraging collective intelligence to solve complex problems, such as software development~\cite{qian2023communicative,hong2023metagpt}, automatic diagnosis~\cite{li2024agent,fan2024ai} and judicial decision-making~\cite{he2024simucourt}. In this case, multiple autonomous agents collaborate on planning, discussion, and decision-making, reflecting the cooperative nature of human groups when solving problems. On the other hand, simple interactions between multiple agents can lead to the emergence of complex collective behaviors or patterns~\cite{schelling1971dynamic,hegselmann2005opinion,chuang2023computational}, thereby replicating complex social dynamics in the real world, such as opinion dynamics~\cite{chuang2023simulating,mou2024unveiling,liu2024skepticism} and macroeconomics phenomena~\cite{li2024econagent}.  Such simulations provide valuable tools for understanding, analyzing, and predicting complex phenomena that may be difficult or impractical to observe directly in real life, offering strong support for decision-making in areas such as policy-making and social management.

This research field is rapidly expanding, with papers focusing on various aspects. Considering the purpose of simulation and the varying demands for diversity, scale, and accuracy in individual modeling, we categorize the existing work into three types, as illustrated in Figure~\ref{fig:intro}: 

\begin{enumerate}
    \item \textbf{Individual Simulation}: leveraging LLM-based agents to mimic specific individuals or groups of people sharing common demographic characteristics~\cite{wang2024rolellmbenchmarkingelicitingenhancing,shao2023characterllmtrainableagentroleplaying,chen2024persona}. This line of research focuses on the replication of features of a single person, e.g., personality, and has not involved multi-agent interactions.
    
    \item \textbf{Scenario Simulation}: organizing a group of agents in a concentrated scenario, driven by specific goals or tasks, such as software development~\cite{qian2023communicative,hong2023metagpt}, question answering~\cite{du2023improving} and paper reviewing~\cite{d2024marg}. Such simulations are usually focused on small-scale agents within specific scenarios,  emphasizing the collective wisdom of agents with specialized expertise.
    
    \item \textbf{Society Simulation}: simulating more complex and diverse behaviors in the agent society to explore social dynamics in real-world applications. Such simulations could test social science theories within a small scope~\cite{chuang2024wisdom} or populate virtual spaces and communities with large-scale realistic social phenomena~\cite{park2023generative,yang2024oasis}. The composition of individuals in such simulations is more complex and diverse.
\end{enumerate}

These three types of simulations exhibit a progressive relationship. Individual simulation models a specific person or a type of person, serving as the foundation for scenario simulation and society simulation. Theoretically, society simulation can encompass a chaotic world composed of countless sub-scenarios, though current work focuses on specific scenarios.

Although this field has seen rapid growth, with some surveys summarizing agent architectures~\cite{xi2023rise,Wang_2024,gao2024large} or certain aspects of single-agent ability or multi-agent systems~\cite{chen2024persona,guo2024large,liu2024large}, there is an absence of a systematic review to summarize the work from the individual to society, providing a comprehensive blueprint for this field. This motivates us to present this survey, aiming to contribute to the research and development of simulations driven by LLM-based agents, as well as a wider range of interdisciplinary studies. To comprehensively describe our landscape, we organize our survey as follows. After a brief introduction to the background in ~\S~\ref{sec:bg}, we begin in ~\S~\ref{sec:indi} by detailing how to conduct individual simulation through discussions of (1) the architecture of a single agent, (2) construction method of individual simulation, (3) the classification of objectives, and (4) the evaluation of individual simulation. Next, in ~\S~\ref{sec:task}, we summarize scenario simulation, including (1) the elements that constitute a scenario simulation system, (2) the classification of scenarios, and (3) the evaluation of scenario simulation, exploring how multiple agents collaborate to achieve objectives within a single scenario. Following this, in ~\S~\ref{sec:soc}, we introduce society simulation, examining how multi-agent systems can construct complex social dynamics through (1) the social construction elements of society simulation, (2) the classification of society simulation scenarios, and (3) the evaluation of society simulation. In ~\S~\ref{sec:data}, we summarize existing datasets and benchmarks. Based on the earlier sections, we analyze trends in these three aspects in ~\S~\ref{sec:trend} and present the conclusion in ~\S~\ref{sec:con}.

\begin{figure*}[t]
    \centering
    \includegraphics[width=\linewidth]{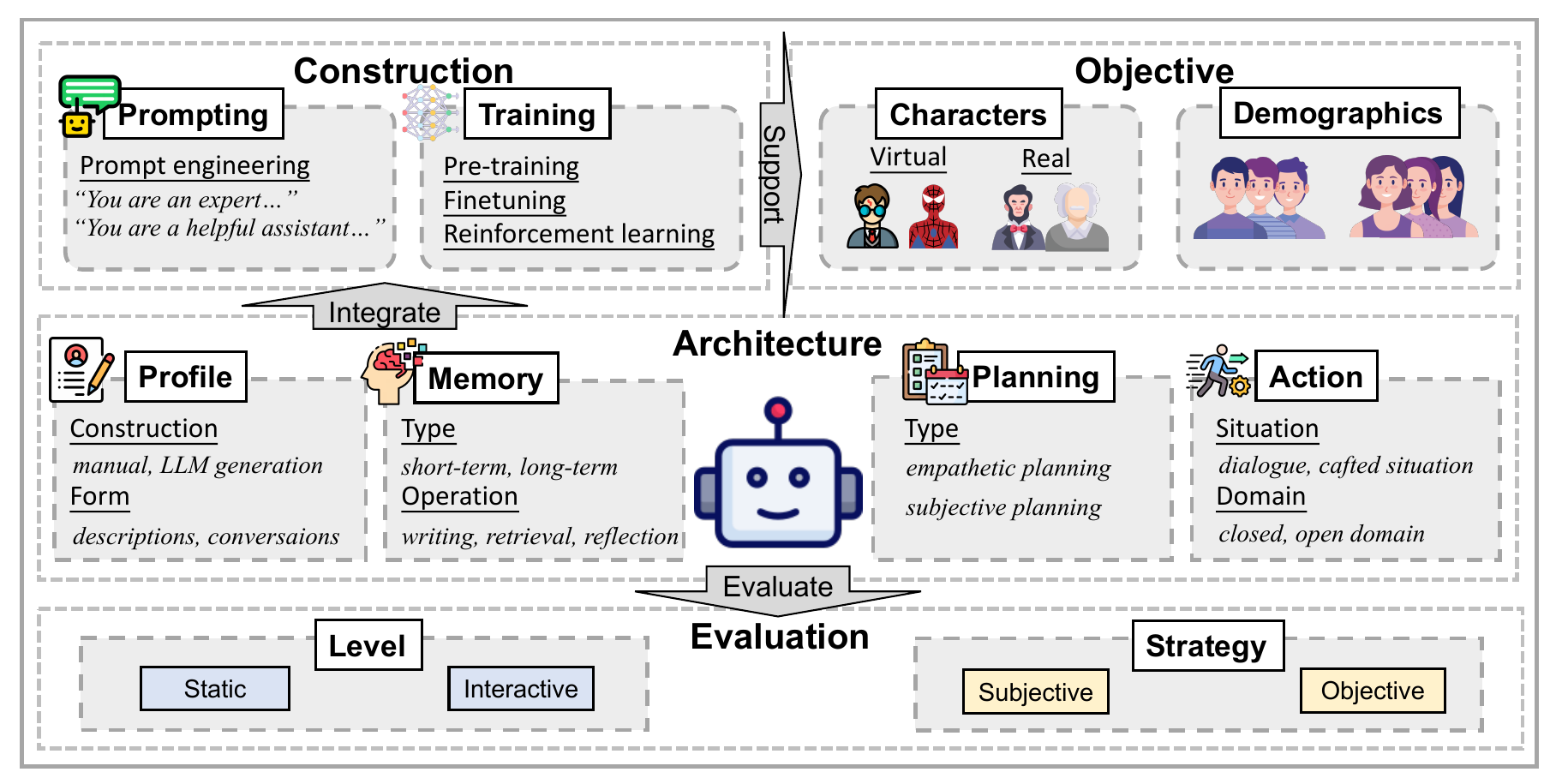}
    \caption{Illustration of individual simulation blueprint. An individual agent is typically composed of an {\ul architecture} with modules involving profile, memory, planning, and action through  {\ul construction} method, prompting or training, to simulate specific {\ul objectives} like characters or demographics . Individual simulation can be {\ul evaluated} statically and interactively with different dimensions being observed.}
    \label{fig:fm_indi}
\end{figure*}

\section{Background}~\label{sec:bg}
\subsection{Large Language Model-based Agents}
Benefiting from the large-scale parameters and pre-training on vast amounts of data, the recently emerging large language models have shown great potential in achieving human-like intelligence~\cite{brown2020language,kojima2022large,achiam2023gpt}. This has sparked a rise in the research of LLM-empowered agents, where the key idea is to equip the LLMs with human capabilities such as memory~\cite{fischer2023reflective,wang2023user}, planning~\cite{yao2022react,hao2023reasoning} and tool usage~\cite{parisi2022talm,schick2024toolformer}. The memory module enables agents to store and operate historical information to facilitate future actions. Memory of different structures~\cite{park2023generative,shinn2024reflexion} and formats~\cite{hu2023chatdb,zhong2024memorybank} have been integrated into LLM-based agents. The planning module helps agents to decompose complex tasks into subtasks, where various planning strategies~\cite{wei2022chain,yao2022react} are adopted. The tool-usage module allows agents to make use of external tools or resources~\cite{yao2022react,ruan2023tptu} to solve tasks. Overall, these modules assist agents in operating more effectively in complex and diverse environments.

\subsection{Multi-agent Systems}
To realize complex scenarios, a single agent is never enough. A system where interaction between multiple agents is involved is referred to as a multi-agent system (MAS). The agents may have a common goal, such as working together to accomplish a task~\cite{qian2023communicative,hong2023metagpt} or solve a problem~\cite{du2023improving}, or they may just have self-interested goals that can cause them to compete for limited resources~\cite{hua2023war}. In a multi-agent system, each agent may be assigned distinct roles and skills, as well as distinct tasks. These agents can be organized in various ways, such as layered or centralized structures~\cite{qian2024scaling,hao2023chatllm,li2024culturepark}, and can communicate through different methods~\cite{chen2024beyond,phamlet,marro2024scalable}. These factors significantly influence the effectiveness and efficiency of multi-agent interactions.


\section{Individual Simulation}~\label{sec:indi}
\begin{table*}[!htp]
\aboverulesep0pt
\belowrulesep0pt
\extrarowheight1pt
\centering
\footnotesize
\tabcolsep3pt
\resizebox*{!}{\dimexpr\textheight-3\baselineskip\relax}{\begin{tabularx}{\linewidth}{c|>{\centering\arraybackslash}X|c|c|c|c|>{\centering\arraybackslash}m{8em}}
    \toprule
    \multirow[m]{2}{*}{\textbf{Objectives}} 
    & \multirow[m]{2.5}{2.5em}{\textbf{Paper}} & \multicolumn{4}{c|}{\textbf{Aritecture}} & \multirow[m]{2.5}{*}{\textbf{Construtction}} \\
    \cmidrule{3-6}
    &  & \textbf{Profile} & \textbf{Memory} & \textbf{Planning} & \textbf{Action Domain} &  \\
    \midrule
    \multirow[m]{19}{*}{\textbf{Characters}} & 
    Brahman et al.
    \cite{brahman2021letcharacterstellstory} & Dialogue/Description & Short-term & - & Open/Closed  & Parametric  \\
    \cmidrule{2-7}
    & Chen et al. \cite{chen2023largelanguagemodelsmeet} & Dialogue/Description & Short-term & - & Open  & \multicolumn{1}{m{7em}}{Parametric /Nonparametric }\\
    \cmidrule{2-7}
    & Schwitzgebel et al. \cite{schwitzgebel2023creatinglargelanguagemodel} & Dialogue & Short-term & - & Open  & Parametric  \\
    \cmidrule{2-7}
    & Generative Agents \cite{park2023generativeagentsinteractivesimulacra} & Description & Short/Long-term & - & Open  & Nonparametric  \\
    \cmidrule{2-7}
    & Agrawal et al. \cite{agrawal-etal-2023-multimodal} & Dialogue/Description & Short-term & - & Open  & Parametric  \\
    \cmidrule{2-7}
    & ChatHaruhi \cite{li2023chatharuhirevivinganimecharacter} & Dialogue & Short-term & - & Open  & Parametric  \\
    \cmidrule{2-7}
    & LiveChat \cite{gao2023livechatlargescalepersonalizeddialogue} & Dialogue/Description & Short/Long-term & - & Open/Closed  & Parametric  \\
    \cmidrule{2-7}
    & RoleLLM \cite{wang2024rolellmbenchmarkingelicitingenhancing} & Description/Dialogue & Short-term & - & Open/Closed  & Parametric  \\
    \cmidrule{2-7}
    & CharacterLLM \cite{shao2023characterllmtrainableagentroleplaying} & Description & Short-term & Subjective & Open  & Parametric  \\
    \cmidrule{2-7}
    & InCharacter \cite{wang2024incharacterevaluatingpersonalityfidelity} & - & Short-term & - & Open/Closed  & - \\
    \cmidrule{2-7}
    & CharacterGLM \cite{zhou2023characterglmcustomizingchineseconversational} & Description/Dialogue & Short-term & - & Open  & Parametric  \\
    \cmidrule{2-7}
    & RoleEval \cite{shen2024roleevalbilingualroleevaluation} & Description & Short-term & - & Closed  & Parametric  \\
    \cmidrule{2-7}
    & CharacterEval \cite{tu2024characterevalchinesebenchmarkroleplaying} & Dialogue & Short-term & - & Open  & Nonparametric  \\
    \cmidrule{2-7}
    & Neeko \cite{yu2024neekoleveragingdynamiclora} & Description & Short-term & - & Open  & Parametric  \\
    \cmidrule{2-7}
    & Character is Destiny \cite{xu2024characterdestinylargelanguage} & Description & Short/Long-term & - & Closed  & Nonparametric  \\
    \cmidrule{2-7}
    & Yuan et al. \cite{yuan2024evaluatingcharacterunderstandinglarge} & Description & Short-term & - & Open/Closed  & Nonparametric  \\
    \cmidrule{2-7}
    & Capturing Minds \cite{ran2024capturingmindsjustwords} & Description/Dialogue & Short/Long-term & Subjective & Open/Closed  & Parametric  \\
    \cmidrule{2-7}
    & MMRole \cite{dai2024mmrolecomprehensiveframeworkdeveloping} & Description & Short-term & - & Open  & Parametric  \\
    \cmidrule{2-7}
    & Yu et al. \cite{yu2024dialogueprofiledialoguealignmentframework} & Dialogue & Short-term & - & Open  & Parametric  \\
    \cmidrule{2-7}
    & Rational sensibility \cite{sun2023rational} & - & Short-term & Empathetic & Closed & Parametric \\
    \midrule
     \multirow[m]{35}{6em}{\textbf{Demographics}}& Karra et al.\cite{karra2023estimatingpersonalitywhiteboxlanguage} & Dialogue/Description & Short-term & - & Closed  & Parametric  \\
     \cmidrule{2-7}
    & Jiang et al. \cite{jiang2023evaluatinginducingpersonalitypretrained} & Description & Short-term & - & Closed  & Nonparametric  \\
    \cmidrule{2-7}
    & Liu et al. \cite{Liu_2022} & Description & Short/Long-term & - & Open  & Parametric  \\
    \cmidrule{2-7}
    & Out of One, Many \cite{Argyle_2023} & Description & Short-term & - & Open  & Nonparametric  \\
    \cmidrule{2-7}
    & Simulated Economic Agents \cite{horton2023largelanguagemodelssimulated} & Description & Short-term & - & Closed  & Nonparametric  \\
    \cmidrule{2-7}
    & The wall street neophyte \cite{xie2023wallstreetneophytezeroshot} & Description & Short-term & Empathetic & Closed  & Nonparametric  \\
    \cmidrule{2-7}
    & Toxicity in ChatGPT \cite{deshpande2023toxicitychatgptanalyzingpersonaassigned} & Description & Short-term & - & Open  & Nonparametric  \\
    \cmidrule{2-7}
    & Song et al. \cite{song2023largelanguagemodelsdeveloped} & Description & Short-term & - & Closed  & Nonparametric  \\
    \cmidrule{2-7}
    & Marked Personas \cite{cheng2023markedpersonasusingnatural} & Description & Short-term & - & Open  & Nonparametric  \\
    \cmidrule{2-7}
    & Wang et al. \cite{wang2024userbehaviorsimulationlarge} & Description & Short/Long-term & - & Open  & Nonparametric  \\
    \cmidrule{2-7}
    & Serapio-García et al. \cite{serapiogarcía2023personalitytraitslargelanguage} & Description & Short-term & - & Open  & Nonparametric  \\
    \cmidrule{2-7}
    & Huang et al. \cite{huang2024emotionallynumbempatheticevaluating} & Description & Short-term & - & Closed  & Nonparametric  \\
    \cmidrule{2-7}
    & CharacterChat \cite{tu2023characterchatlearningconversationalai} & Description & Short/Long-term & - & Open  & Nonparametric  \\
    \cmidrule{2-7}
    & Conversational health agents \cite{abbasian2024conversationalhealthagentspersonalized} & Description & Short/Long-term & Empathetic & Open  & Nonparametric  \\
    \cmidrule{2-7}
    & Chen et al. \cite{chen2024moneymouthisevaluating} & Description & Short/Long-term & - & Closed  & Nonparametric  \\
    \cmidrule{2-7}
    & EconAgent \cite{li2024econagentlargelanguagemodelempowered} & Description & Short/Long-term & - & Open  & Nonparamaetric  \\
    \cmidrule{2-7}
    & Shea et al. \cite{shea2023buildingpersonaconsistentdialogue} & Dialogue & Short-term & - & Open  & Parametric  \\
    \cmidrule{2-7}
    & Be Selfish, But Wisely \cite{chawla2023selfishwiselyinvestigatingimpact} & Dialogue & Short-term & - & Open  & Parametric  \\
    \cmidrule{2-7}
    & Chain of Empathy \cite{10.19066/COGSCI.2024.35.1.002} & - & Short-term & Empathetic & Open  & Nonparametric  \\
    \cmidrule{2-7}
    & Bias Runs Deep \cite{gupta2024biasrunsdeepimplicit} & Description & Short-term & - & Open  & Nonparametric  \\
    \cmidrule{2-7}
    & Li et al. \cite{li2024steerabilitylargelanguagemodels} & Dialogue & Short-term & - & Open  & Parametric  \\
    \cmidrule{2-7}
    & Xie et al. \cite{xie2024largelanguagemodelagents} & Description & Short-term & Subjective & Closed  & Nonparametric  \\
    \cmidrule{2-7}
    & Lee et al. \cite{Lee_2024} & Description & Short-term & - & Closed  & Nonparametric  \\
    \cmidrule{2-7}
    & CultureLLM \cite{li2024culturellmincorporatingculturaldifferences} & Dialogue & Short-term & - & Open  & Parametric  \\
    \cmidrule{2-7}
    & ControlLM \cite{weng2024controllmcraftingdiversepersonalities} & - & Short/Long-term & - & Open  & Nonparamatric  \\
    \cmidrule{2-7}
    & Random Silicon Sampling \cite{sun2024randomsiliconsamplingsimulating} & Description & Short-term & - & Closed  & Nonparametric  \\
    \cmidrule{2-7}
    & Bisbee et al. \cite{Bisbee_Clinton_Dorff_Kenkel_Larson_2024} & Description & Short-term & - & Closed  & Nonparametric  \\
    \cmidrule{2-7}
    & PersonaHub \cite{ge2024scalingsyntheticdatacreation} & Description & Short-term & - & Open  & Parametric  \\
    \cmidrule{2-7}
    & Qu et al. \cite{Qu2024PerformanceAB} & Description & Short-term & - & Closed  & Nonparametric  \\
    \cmidrule{2-7}
    & Interactive Agents \cite{qiu2024interactiveagentssimulatingcounselorclient} & Description & Short-term & - & Open  & Nonparametric  \\
    \bottomrule
\end{tabularx}}
\caption{A list of representative works of individual simulation.}
\label{tab:indi}
\end{table*}

Individual simulation focuses on designing a modular architecture that integrates individualized data for the construction of agents and simulating the specific objective with high fidelity.
In this section, we first outline the basic architecture of the agent in the individual simulation with four key components in~\S\ref{subsec:indi_arch}. Then, two construction methods are discussed in~\S\ref{subsec:indi_cons} to implement the integration of individualized data into objectives introduced in~\S\ref{subsec:indi_obj}. The evaluation methods are examined from different perspectives in~\S\ref{subsec:indi_eval}. The overall framework is presented in Figure~\ref{fig:fm_indi} and representative works are summarized in Table~\ref{tab:indi}.


\subsection{Architecture}\label{subsec:indi_arch}
To effectively accomplish individual simulation, it is essential to construct an agent architecture that can accurately replicate the features of the individual. This requires a balance between theoretical abstraction and practical implementation to capture the complexity of human behaviors. Typically, this architecture is modularized into four core components: \textbf{profile}, \textbf{memory}, \textbf{planning}, and \textbf{action}.

\subsubsection{Profile}

Profile differentiates the unique characteristics of simulated individuals, encompassing attributes, behaviors, and constraints. The profiles differ in the ways of construction and their forms.

\paragraph{Profile Construction}

Profile construction refers to the process of collecting individual-related information, which can be categorized into \textit{manual modification} and \textit{LLM generation}. 
\textit{{\ul Manual modification}} takes advantage of publicly available data to create high-quality profiles through a human-guided process. According to the collected sources, manual modification can also be classified into three categories:  handcrafting, online communities, and historical works. Handcrafting manually organized some coarse strength information, such as well-known characters~\cite{wang2023humanoidagentsplatformsimulating} and specific personalities~\cite{deshpande2023toxicitychatgptanalyzingpersonaassigned,cheng2023markedpersonasusingnatural}, while online communities construct profiles built on the web data like Wikipedia~\cite{shao2023characterllmtrainableagentroleplaying} and social media~\cite{gao2023livechatlargescalepersonalizeddialogue}, where the profile implicitly exists in conversations and materials.
In addition, literary works serve as additional descriptions that reflect the author's thoughts~\cite{schwitzgebel2023creatinglargelanguagemodel} and characters in the storyline~\cite{brahman2021letcharacterstellstory,li2023chatharuhirevivinganimecharacter}. 
\textit{{\ul LLM generation}} automatically generates the expected persona-based information profiles by prompting LLMs with essential individual details~\cite{tu2023characterchatlearningconversationalai,wang2024rolellmbenchmarkingelicitingenhancing,wang2024incharacterevaluatingpersonalityfidelity}. This method explores diverse profiles with ease, while the quality needs human supervision with caution.


\paragraph{Profile Form}

Profile form defines the format of individual information, which can be categorized into  \textit{descriptions} and \textit{conversations}.
\textit{{\ul Descriptions}} directly describe basic individual information or identity with details like name, age, and gender~\cite{jang2022customizedconversationcustomizedconversation,wang2023humanoidagentsplatformsimulating}. While descriptions can intuitively reflect the basic attributes of an individual, deeper contextual information can also be ignored.
On the contrary, \textit{{\ul conversations}} implicitly reflect the character profile through dialogue.
A substantial amount of conversational data is derived from sources such as films, literary works, and scripts~\cite{li2021dialoguehistorymatterspersonalized,brahman2021letcharacterstellstory,jandaghi2023faithfulpersonabasedconversationaldataset,yu2024dialogueprofiledialoguealignmentframework}. 
Considering the extensive commonsense knowledge learned by LLMs in the pre-training stage, recent works leverage LLMs to generate individual dialogues~\cite{li2023chatharuhirevivinganimecharacter,ge2024scalingsyntheticdatacreation}, which defines the artistic genre through six essential elements to generate detailed drama scripts~\cite{wu2024roleplaydramainteractionllmsolution} and imitates speaking styles through context learning~\cite{wang2024rolellmbenchmarkingelicitingenhancing,yu2024neekoleveragingdynamiclora}.


\subsubsection{Memory}
Memory is designed to store perceived or generated information, helping agents maintain consistency and continuity of behavior and overcome the limited context window of LLMs. Considering the complexity of memory, researchers struggle to design more efficient memory types and operations.

\paragraph{Memory Type}
Based on the temporal span of stored content, memory can be commonly divided into two types, namely \textit{short-term memory} and \textit{long-term memory}. \textit{{\ul Short-term memory}} records the instant local information that the agent perceives, which can be further divided into simulation contents and simulation supplements. Simulation contents include essential interaction data like user instructions~\cite{schwitzgebel2023creatinglargelanguagemodel,deshpande2023toxicitychatgptanalyzingpersonaassigned}, dialogue history~\cite{xiang2023languagemodelsmeetworld,huang2024embodiedgeneralistagent3d}, and user/environment responses~\cite{xie2023wallstreetneophytezeroshot}. Simulation supplements provide additional environmental information including scene descriptions~\cite{xie2023wallstreetneophytezeroshot,agrawal-etal-2023-multimodal} and scene-related experiences~\cite{shao2023characterllmtrainableagentroleplaying,xu2024characterdestinylargelanguage}, which navigate agents through the simulation to perform tasks appropriately.
\textit{{\ul Long-term memory}} stores persistent global information, preventing deviations from intended goals, which holds extensive individual-specific information stably, including past experiences and behaviors, current knowledge, and skills~\cite{li2024econagentlargelanguagemodelempowered,xu2024characterdestinylargelanguage}. With the proposal of using the vector database as the long-term memory hub, the management, retrieval, and organization of memory is more effective~\cite{lin2023agentsimsopensourcesandboxlarge}.

\paragraph{Memory Operation}

Memory operations stand for the continuous updating and utilization of memory by the agent. The common memory operations include three types, namely \textit{memory writing}, \textit{memory retrieval}, and \textit{memory reflection}.

\textit{{\ul Memory writing}} aims to incorporate the relevant historical content into the memory. This process mirrors human memory formation, where useful information is retained for future retrieval. The memories to be written vary from user-specific dialogue history~\cite{li2021dialoguehistorymatterspersonalized}, new skills~\cite{wang2023voyageropenendedembodiedagent}, to selected papers and other forms~\cite{wang2024surveyagentconversationalpersonalizedefficient}.


\textit{{\ul Memory retrieval}} serves to extract valuable content from memory based on customized requirements. The overall performance of the individual simulation highly relies on the effectiveness of memory retrieval since simulations are sensitive to the context.
Traditional retrieval technologies rely on similarity such as keyword matching~\cite{chen2024socialbenchsocialityevaluationroleplaying} and embedding vectors~\cite{lin2023agentsimsopensourcesandboxlarge}, while recent works introduce the retrieval model to select the most relevant information~\cite{chen2023chatcottoolaugmentedchainofthoughtreasoning,salemi2024lamplargelanguagemodels}.

\textit{{\ul Memory reflection}} mirrors the human ability to reconsider past behaviors and opinions. Specifically, it helps the agent to organize, refine, and elevate memories into more abstract and insightful concepts. Generative Agents~\cite{park2023generativeagentsinteractivesimulacra} maintains a comprehensive record of agents' experiences with a tree-structured reflection process to optimize memory usage. ProAgent~\cite{zhang2024proagentbuildingproactivecooperative} incorporates memory reflection with validation and belief correction to improve the agent’s planning and decision-making. Voyager~\cite{wang2023voyageropenendedembodiedagent} allows agents to reflect on their behavior and update their skill libraries through self-verification. Although the application scenarios of memory reflection are still limited, it shows great improvement in enhancing performance and increasing the depth of simulations, especially in complex environments.

\subsubsection{Planning}

Planning is the process of deciding on a series of actions aimed at achieving specific goals. Traditional planning tasks typically focus on solving particular problems, such as mathematical reasoning~\cite{wang2023plan} or embodied tasks~\cite{wu2023embodied, song2023llmplannerfewshotgroundedplanning}. At the individual simulation level, however, agents are expected to go beyond mere problem-solving. They should also be able to simulate personalized thinking and emotional responses during interactions with specific individuals. This extends planning into two additional categories: empathetic planning and subjective planning.

\paragraph{Empathetic planning} Empathetic planning refers to an agent's ability to infer and perceive the behavior and emotions of others before taking action. It involves using Chain-of-Thought (CoT) reasoning to understand the situations of others and make adaptive decisions or judgments~\cite{xie2023wallstreetneophytezeroshot, 10.19066/COGSCI.2024.35.1.002, sun2023rational}. This allows the agent to tailor its actions based on the emotional and behavioral context, guiding the acquisition of personalized feedback.

\paragraph{Subjective planning} Subjective planning refers to the actions an agent takes based on its own thoughts and feelings, in line with its predefined role or identity.
This can involve utilizing inner monologues from simulated characters to fine-tune LLMs~\cite{shao2023characterllmtrainableagentroleplaying,ran2024capturingmindsjustwords} or using CoT to guide LLMs to express themselves according to their own beliefs~\cite{xie2024largelanguagemodelagents}. 
This form of planning is driven by the agent's internal state, rather than by external stimuli or the needs of others.

\subsubsection{Action}\label{sec:action}

Action refers to the direct interaction between LLMs and their environment. Action encompasses two key aspects: the action situation, which describes the context in which actions occur, and the action domain, which defines the requirements for action space. Action serves as the interface for simulating human behavior, allowing LLMs to execute tasks that mimic real-world actions and responses. This interaction enables a deeper understanding of human-like decision-making and execution in various scenarios.

\paragraph{Action Situation}
With individual simulations focusing on more and more diverse and complicated situations, various action situations spring out accordingly, ranging from dialogue~\cite{cho2022personalizeddialoguegeneratorimplicit}, games~\cite{light2023avalonbenchevaluatingllmsplaying}, real word~\cite{xiang2023languagemodelsmeetworld}, etc. Typically, action situations can be divided into \textit{simple dialogues} and \textit{crafted situations}.

\textit{{\ul Simple dialogues}} are few-turn conversations without restricted environments, such as constructing dialogues between two characters~\cite{brahman2021letcharacterstellstory}. Recent researches utilize simple dialogues to induce potential attributes within the models, involving personality~\cite{karra2023estimatingpersonalitywhiteboxlanguage,jiang2023evaluatinginducingpersonalitypretrained}, traits~\cite{serapiogarcía2023personalitytraitslargelanguage} and toxicity~\cite{deshpande2023toxicitychatgptanalyzingpersonaassigned}. Other works conduct evaluations of persona with interviewing~\cite{wang2024incharacterevaluatingpersonalityfidelity} or questionnaire~\cite{pan2023llmspossesspersonalitymaking} with simple dialogues to facilitate their experiment.

\textit{{\ul Crafted situations}} are elaborately designed environments including detailed rules and surrounding descriptions. Common situations like games are modified from simple dialogues. They leverage game rules to provide a settled virtual topic for both users and agents to play in, especially in the board role-playing games~\cite{light2023avalonbenchevaluatingllmsplaying,light2023avalonbenchevaluatingllmsplaying}~\cite{chalamalasetti2023clembenchusinggameplay}. Besides, researchers have developed a more delicate environment called sandbox~\cite{chen2024socialbenchsocialityevaluationroleplaying}, which not only includes rules but establish an objective environment. To further enrich the individual simulation situation, some authors add some elements existing in scripts like facial expression, tiny movements~\cite{agrawal-etal-2023-multimodal,wu2024roleplaydramainteractionllmsolution}, and nuanced information from environment images~\cite{dai2024mmrolecomprehensiveframeworkdeveloping}.

\paragraph{Action Domain} The Action domain can be commonly divided into \textit{close domain} and \textit{open domain} based on the restriction of action space.

\textit{{\ul Closed domain}} simulation occurs when the available action space is limited. In simple situations such as completing questionnaires testing~\cite{karra2023estimatingpersonalitywhiteboxlanguage}, making decisions from a set of options~\cite{horton2023largelanguagemodelssimulated}, or rating with predefined standards~\cite{wang2024incharacterevaluatingpersonalityfidelity}, the action space of LLMs is determined by researchers ahead of simulation to make responses predictable. In practical scenarios, LLMs are required to choose tools~\cite{chen2023chatcottoolaugmentedchainofthoughtreasoning,yuan2023skill} or select specific functions to complete concrete tasks, like recommending, browsing, and compiling.
Individual simulation with agents in closed-domain tasks can improve human work efficiency, extending beyond entertainment purposes.

\textit{{\ul Open domain}} simulation places few restrictions on actions, allowing LLMs to generate responses freely. This approach more closely resembles real-world conditions, but also demands higher standards for individual simulation. Among various open-domain tasks, taking actions through conversation is a popular method for simulating individual behavior~\cite{brahman2021letcharacterstellstory,li2023chatharuhirevivinganimecharacter,zhou2023characterglmcustomizingchineseconversational,yu2024neekoleveragingdynamiclora}, in which the varied settings stimulate LLMs’ potential for individual simulation and allow researchers to oversee simulations across diverse and nuanced dimensions. Another growing method of open-domain simulation is scenario-based interaction, where LLMs are assigned roles and are required to interact in crated situations like sandbox~\cite{wang2023voyageropenendedembodiedagent,lin2023agentsimsopensourcesandboxlarge} or established game settings~\cite{light2023avalonbenchevaluatingllmsplaying,chalamalasetti2023clembenchusinggameplay}.

\subsection{Construction}\label{subsec:indi_cons}

Construction indicates the process of integrating individual data into the established model of LLMs, which aligns the design model and the individual, thus creating the simulating LLMs. Generally, construction methods are distinguished into two types, namely \textbf{nonparametric prompting} and \textbf{parametric training}.

\subsubsection{Nonparametric Prompting}

Nonparametric prompting, i.e. prompt engineering, is a method of interacting with LLMs by designing and optimizing input prompts. In some individual simulations, the description-based profile is implemented by a system prompt. Researchers often create system prompts that begin with \textit{``You are a..."} to assign models specific demographic features and roles~\cite{deshpande2023toxicitychatgptanalyzingpersonaassigned}. Besides, LLM outputs are enhanced in some works through few-shot prompting by providing specific examples to inject detailed information and improve response quality. Moreover, incorporating problem-specific details directly within prompt structures can significantly enhance the effectiveness of the simulation.

Short-term memory is often implemented by nonparametric prompting. For situation-based individual simulations, environment descriptions and behavior rules are typically conveyed through prompt engineering~\cite{chalamalasetti2023clembenchusinggameplay}. Since situational information is generally objective and must be followed, emphasizing this information directly in the input is a rather effective method for constructing simulations. However, due to the context window limitations of LLMs, the quality of the profile prompt significantly restricts prompt-based individual simulations. 
Moreover, the preset template configurations as the \textit{``assistant"} within 
LLMs pose a major challenge for prompt engineering in individual simulations~\cite{tu2023characterchatlearningconversationalai}.

\subsubsection{Parametric Training}
Parametric training modifies the model by directly updating the LLM parameters with given data. The training methods can be generally categorized into pre-training, finetuning, and reinforcement learning.

\paragraph{Pre-training} The pre-training method in individual simulation focuses on aligning the original LLMs with basic individual-related data and setting up a fundamental knowledge of individuals for LLMs. The targets of training datasets vary in recent studies, including individual descriptions~\cite{salemi2024lamplargelanguagemodels}, literature summaries~\cite{brahman2021letcharacterstellstory}, and philosophical works or utterances~\cite{schwitzgebel2023creatinglargelanguagemodel}.

\paragraph{Finetuning} The finetuning method is designed for adapting LLMs for individual simulation in specific tasks and situations. Researchers collect and modify supervised instruction datasets tailored for specific situations and fine-tune their models to equip them with the corresponding capabilities. Using persona-enhanced datasets is an effective method to regulate the models' behavior in individual simulation, which is constructed by adding instruction tuning samples of the simulated individual’s behavior~\cite{ran2024capturingmindsjustwords,ge2024scalingsyntheticdatacreation}. 
LoRA finetuning method can integrate multiple characters into a single model~\cite{yu2024neekoleveragingdynamiclora,sun2024identity}. In multimodal finetuning scenarios, both visual and textual information are considered to significantly enhance LLMs' simulation behavior in multimodal contexts~\cite{salemi2024lamplargelanguagemodels,dai2024mmrolecomprehensiveframeworkdeveloping}. Compared to prompt engineering, finetuning leverages large datasets more effectively and reduces the limitations imposed by the pre-training phase of LLMs.

\paragraph{Reinforcement Learning} The reinforcement learning method is used to refine models in dynamic environments with the goal of maximizing cumulative rewards. In simulations involving conversations and dialogues, the quality of the LLM's responses directly influences the rewards it receives~\cite{bai2022traininghelpfulharmlessassistant,shea2023buildingpersonaconsistentdialogue,jang2023personalizedsoupspersonalizedlarge}, which encourages the model to learn the appropriate ways to respond in dialogues.
By modifying the reward function, researchers can influence the model's preference and thus manage to mimic the personas of the simulated individuals~\cite{chawla2023selfishwiselyinvestigatingimpact}. As individual simulations become more diverse and complex, reinforcement learning plays a crucial role in improving the dynamic behavior of simulated LLMs.

\subsection{Simulation Objectives}\label{subsec:indi_obj}

The simulation objectives of individual simulation for various purposes can be divided into two categories: (1) \textbf{Demographics}: a group of people who share the same characteristics, such as psychological traits (e.g., INTJ) or identity-related features (e.g., farmers). (2) \textbf{Characters}: a specific individual, whether real or virtual, who is widely recognized by groups of people.

\subsubsection{Demographics}
Demographic individuals refer to a group of people who share the same features. In an abstract sense, demographics can be understood as the centroid of an embedding space that represents common opinions and beliefs, essentially clustering individual embeddings for classification purposes~\cite{li2024steerabilitylargelanguagemodels}. Demographic simulation involves assigning an identity, such as ``student," to LLMs and guiding the simulators to perform specific tasks. Early demographic simulations have focused on investigating the internal demographic attributes within pre-trained models~\cite{Liu_2022,li2024evaluatingpsychologicalsafetylarge}, laying the groundwork for further simulations. Additionally, these simulations are used to reflect opinion surveys~\cite{Lee_2024} or evaluate preferences and biases~\cite{Qu2024PerformanceAB,lee2024exploringsocialdesirabilityresponse} of particular groups. With the ability to scale synthetic dialogue~\cite{cho2023crowdmeetspersonacreating,shen2024roleevalbilingualroleevaluation,ge2024scalingsyntheticdatacreation} involving specific personas, demographic simulations can also contribute to societal simulation studies~\cite{chen2024socialbenchsocialityevaluationroleplaying}. In most cases, demographic simulation is implemented through nonparametric prompting. Many researchers in this field focus on designing tasks, such as questionnaires or social experiments~\cite{horton2023largelanguagemodelssimulated}, to fully tap into the simulating potential of LLMs.

\subsubsection{Characters}
Characters are distinct individuals who differ from one another. They may be ordinary platform users, well-known public figures, or fictional characters from novels. Researchers favor these characters because they enhance the expertise of LLMs in specific domains and challenge the learning capabilities of these models. From Haruhi and Li Yunlong~\cite{li2023chatharuhirevivinganimecharacter} to Beethoven~\cite{xu2024characterdestinylargelanguage}, individual simulations select their protagonists from both real and virtual worlds. 

\paragraph{Real Characters} Real characters, typically famous figures, are associated with high-quality data from platforms like Wikipedia and social media, making it easier to establish objective profiles and evaluate simulations. Many LLMs focus on historical figures, celebrities across various periods and backgrounds~\cite{shao2023characterllmtrainableagentroleplaying,mou2024unifying}, characters from online encyclopedias~\cite{tu2024characterevalchinesebenchmarkroleplaying}, and popular livestreamers on Douyin~\cite{gao2023livechatlargescalepersonalizeddialogue}. Since LLMs often have prior knowledge of these individuals, creating their profiles is relatively straightforward. Real and simulated characters are also used to test LLM simulation capabilities, such as in philosopher simulations~\cite{schwitzgebel2023creatinglargelanguagemodel}.
\paragraph{Virtual Characters} Virtual characters are fictional roles created in novels, movies, and video games. Advancements in virtual character simulation can significantly benefit entertainment sectors like the gaming industry and theme parks. Many researchers have drawn inspiration from famous fictional characters, such as Harry Potter~\cite{chen2023largelanguagemodelsmeet}, Sun Wukong~\cite{zhou2023characterglmcustomizingchineseconversational}, and Tong Xiangyu~\cite{li2024personalllmagentsinsights}. Additionally, some experiments design virtual characters~\cite{light2023avalonbenchevaluatingllmsplaying} with specific attributes or objectives. However, despite the attention virtual character simulation attracts, developing virtual individual LLMs presents challenges, particularly in ensuring the quality and reliability of their datasets. Most simulations of virtual characters are designed for interactive conversations, enhancing user experience in various entertaining scenarios.

\begin{figure*}[t]
    \centering
    \includegraphics[width=\linewidth]{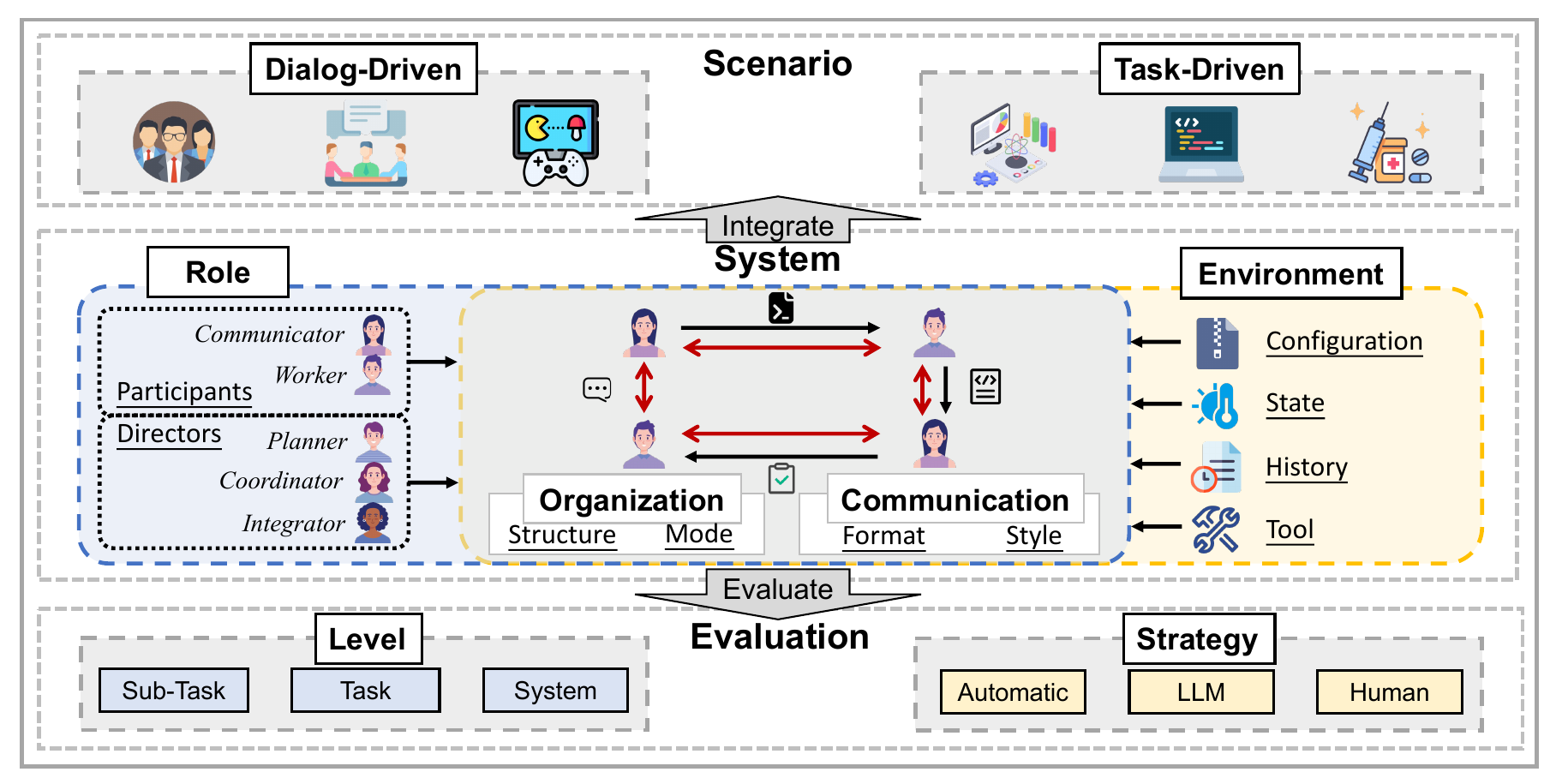}
    \caption{Illustration of scenario simulations. Given a specific scenario, building a multi-agent {\ul system} involves modeling environment, roles, organization, and communication with detailed modules or mechanisms adjusted to the targeted scenario being supported. After simulating the {\ul scenario}, the desired output, typically the result of a task or problem, is obtained and {\ul evaluated} using different levels and strategies.}
    \label{fig:fm_task}
\end{figure*}

\subsection{Evaluation}\label{subsec:indi_eval}

To measure the performance of individual simulations, provide insights into their feasibility, and guide improvements to simulation architectures, researchers have developed diverse evaluation standards and methods, ranging from simple to complex approaches. These methods can be categorized into \textbf{static evaluation} and \textbf{interactive evaluation}.

\subsubsection{Static Evaluation}

Static evaluation refers to the dialogue-based assessment of LLMs by directly inducing their generation and measuring their quality. It can be categorized into subjective evaluation, which involves assessments by both LLMs and human evaluators, and objective evaluation, which utilizes mathematical tools for analysis.

\paragraph{Subjective Evaluation} Subjective evaluation refers to assessments conducted by humans or LLMs based on subjective standards. It often involves leveraging conversations with varying forms and contexts. Interview techniques are widely adopted~\cite{wang2024rolellmbenchmarkingelicitingenhancing,wang2024incharacterevaluatingpersonalityfidelity} because they can effectively prompt LLMs to generate expected responses. Other approaches, such as utterance imitation~\cite{deshpande2023toxicitychatgptanalyzingpersonaassigned}, are also favored in some research. Once dialogues are generated, some studies utilize advanced LLMs to evaluate the output on a given scale~\cite{wang2024incharacterevaluatingpersonalityfidelity,li2024personalllmagentsinsights,yu2024neekoleveragingdynamiclora}, considering performance dimensions. These dimensions range from psychology-based metrics, such as the Big Five Personality Traits (BFI) and Myers-Briggs Type Indicator (MBTI), to language-based factors like grammar and tone. Human annotators are often involved in experiments to provide human reference points ~\cite{park2023generativeagentsinteractivesimulacra,abbasian2024conversationalhealthagentspersonalized,baek2024knowledgeaugmentedlargelanguagemodels}.
\par
\paragraph{Objective Evaluation}Objective evaluation refers to assessments based on objective indicators, utilizing mathematical and statistical tools. It takes advantage of mathematical tools to grade the generation of simulating LLMs. Examination commonly involves option choosing(or questionnaire)~\cite{karra2023estimatingpersonalitywhiteboxlanguage}, ranking~\cite{gao2023livechatlargescalepersonalizeddialogue} and question answering~\cite{jang2022customizedconversationcustomizedconversation}. Accuracy~\cite{xiang2023languagemodelsmeetworld,li2024steerabilitylargelanguagemodels}, F1 score, recall~\cite{branch2022evaluatingsusceptibilitypretrainedlanguage,ahn-etal-2023-mpchat} are used in option choosing and ranking. In the examination of generation(question answering), text sequence related tools such as perplexity~\cite{li2016personabasedneuralconversationmodel,cho2022personalizeddialoguegeneratorimplicit,agrawal-etal-2023-multimodal}, ROUGE-L~\cite{Liu_2022,chen2023largelanguagemodelsmeet,xiang2023languagemodelsmeetworld} and BLUE~\cite{li2016personabasedneuralconversationmodel,Liu_2022,branch2022evaluatingsusceptibilitypretrainedlanguage,gao2023livechatlargescalepersonalizeddialogue} are broadly used in the evaluation, especially those with a reference version~\cite{chen2023largelanguagemodelsmeet}. Objective Examination is a more credible method of evaluating the performance of LLMs in individual simulation. However, it is highly restricted, and occasionally, specific objective tools must be developed to facilitate the evaluation of simulation in given dimensions.

\subsubsection{Interactive Evaluation}

Interactive evaluation refers to a circumstance-based assessment that creates a detailed interactive environment to measure the ability of individual simulations in complex scenarios. It is commonly applied in areas such as game performance~\cite{chalamalasetti2023clembenchusinggameplay,light2023avalonbenchevaluatingllmsplaying}, task completion~\cite{chen2023chatcottoolaugmentedchainofthoughtreasoning,wang2023chatcoderchatbasedrefinerequirement,farn2023tooltalkevaluatingtoolusageconversational}, and nuanced role-playing~\cite{chawla2023selfishwiselyinvestigatingimpact,jandaghi2023faithfulpersonabasedconversationaldataset}. Three key features of interactive evaluation are the carefully designed environment, real-time interactive external responses, and multi-stage assessments. Information about the crafted environment has been introduced in \S\ref{sec:action}. Real-time interactive external responses refer to the feedback from the external environment in reaction to the outputs of simulating LLMs. Agent-environment interactions construct multiple dialogues between the LLMs and the environment. These interactions help reveal the LLMs' capabilities in complex contexts, leading to more dynamic simulations. Single-aspect measurements are insufficient for interactive evaluation, so many studies adopt evaluated objectives that range from specific actions to hybrid actions~\cite{wang2024surveyagentconversationalpersonalizedefficient}, or from single-turn interactions to multi-turn dialogues~\cite{shao2023characterllmtrainableagentroleplaying}. Other studies assess generation quality, focusing on aspects such as accuracy relative to ground truth, nuanced simulations like tone imitation~\cite{wang2024rolellmbenchmarkingelicitingenhancing,huang2024embodiedgeneralistagent3d}, and self-reporting consistency~\cite{liu2023agentbenchevaluatingllmsagents}. In interactive evaluation, researchers prioritize not only accuracy but also the degree to which the simulation resembles real-world scenarios.

\section{Scenario Simulation}~\label{sec:task}

\begin{table*}[!htp]
\setlength{\tabcolsep}{2pt}
\centering
\resizebox{\textwidth}{!}{\begin{tabular}{c|c|c|c|c|c|c|c|c|c|c|c}
    \toprule
    \multirow{2}{*}{\textbf{Scenario}}                                          & \multirow{2}{*}{\textbf{Task}}                                                                   & \multirow{2}{*}{\textbf{Paper}}                      & \multicolumn{4}{c|}{\textbf{Environment}}                                   & \multicolumn{3}{c|}{\textbf{Director Role}}                   & \multirow{2}{*}{\textbf{Organization}} & \multirow{2}{*}{\textbf{Communication}} \\ \cline{4-10}
                                                                                &                                                                                                  &                                                      & \textbf{Configuration} & \textbf{State} & \textbf{History} & \textbf{Tools} & \textbf{Planner} & \textbf{Coordinator} & \textbf{Integrator} &                                        &                                         \\ \midrule
    \multirow{29}{*}{\begin{tabular}[c]{@{}c@{}}Dialog- \\ Driven\end{tabular}} & \multirow{5}{*}{\begin{tabular}[c]{@{}c@{}}Social \\ Interaction\end{tabular}}                   & Sotopia~\cite{zhou2023sotopia}                       & \checkmark             & \checkmark     & \checkmark       &                &                  &                      &                     & static,single                          & UNL                                     \\ \cline{3-12} 
                                                                                &                                                                                                  & Elicitron~\cite{ataei2024elicitron}                  & \checkmark             &                & \checkmark       &                &                  &                      &                     & static,multi                           & UNL                                     \\ \cline{3-12}
                                                                                &                                                                                                  & APAM~\cite{yang2024social}                           & \checkmark             &                & \checkmark       &                &                  &                      &                     & static,single                          & UNL                                     \\ \cline{3-12}
                                                                                &                                                                                                  & SimuLife++~\cite{yan2024social}                      & \checkmark             &                & \checkmark       &                &                  &                      &                     & static,single                          & UNL                                     \\ \cline{3-12}
                                                                                &                                                                                                  & Self-Emotion~\cite{zhang2024self}                    & \checkmark             &                & \checkmark       &                &                  & \checkmark           &                     & dynamic,single                         & UNL                                     \\ \cline{2-12}
                                                                                & \multirow{9}{*}{\begin{tabular}[c]{@{}c@{}}Question \\ Answering\end{tabular}}                   & ICL-AIF~\cite{fu2023improving}                       & \checkmark             &                & \checkmark       &                &                  & \checkmark           &                     & static,single                          & UNL                                     \\ \cline{3-12}
                                                                                &                                                                                                  & FORD~\cite{xiong2023examining}                       & \checkmark             &                & \checkmark       &                &                  & \checkmark           &                     & static,multi                           & UNL                                     \\ \cline{3-12}
                                                                                &                                                                                                  & du et al.~\cite{du2023improving}                     & \checkmark             &                & \checkmark       &                &                  &                      &                     & static,single                          & UNL                                     \\ \cline{3-12}
                                                                                &                                                                                                  & MAD~\cite{liang2023encouraging}                      & \checkmark             &                & \checkmark       &                &                  & \checkmark           &                     & static,single                          & UNL                                     \\ \cline{3-12}
                                                                                &                                                                                                  & ChatEval~\cite{chan2023chateval}                     & \checkmark             &                & \checkmark       &                &                  &                      & \checkmark          & static,single                          & UNL                                     \\ \cline{3-12}
                                                                                &                                                                                                  & AutoGen~\cite{wu2023autogen}                         & \checkmark             &                & \checkmark       & \checkmark     &                  &                      &                     & dynamic,single                         & UNL                                     \\ \cline{3-12}
                                                                                &                                                                                                  & AmazonHistoryPrice~\cite{xia2024measuring}           & \checkmark             &                & \checkmark       &                &                  &                      &                     & static,single                          & UNL                                     \\ \cline{3-12}
                                                                                &                                                                                                  & DoG~\cite{ma2024debate}                              & \checkmark             &                & \checkmark       & \checkmark     &                  &                      &                     & dynamic,single                         & UNL                                     \\ \cline{3-12}
                                                                                &                                                                                                  & ChatLLM~\cite{hao2023chatllm}                        & \checkmark             &                &                  &                &                  &                      &                     & static,single                          & UNL                                     \\ \cline{2-12}
                                                                                & \multirow{15}{*}{Game}                                                                           & xu et al.\cite{xu2023exploring}                      & \checkmark             &                & \checkmark       &                &                  &                      &                     & static,multi                           & UNL                                     \\ \cline{3-12}
                                                                                &                                                                                                  & ReCon~\cite{wang2023avalon}                          & \checkmark             &                & \checkmark       &                &                  &                      &                     & static,multi                           & UNL                                     \\ \cline{3-12}
                                                                                &                                                                                                  & MachineSoM~\cite{zhang2023exploring}                 & \checkmark             &                & \checkmark       &                &                  &                      &                     & dynamic,single                         & UNL                                     \\ \cline{3-12}
                                                                                &                                                                                                  & AvalonBench~\cite{light2023text}                     & \checkmark             &                & \checkmark       &                &                  &                      &                     & static,multi                           & UNL                                     \\ \cline{3-12}
                                                                                &                                                                                                  & lan et al.~\cite{lan2023llm}                         & \checkmark             &                & \checkmark       &                &                  &                      &                     & static,multi                           & UNL                                     \\ \cline{3-12}
                                                                                &                                                                                                  & xu et al.~\cite{xu2023language}                      & \checkmark             &                & \checkmark       &                &                  &                      &                     & static,multi                           & UNL                                     \\ \cline{3-12}
                                                                                &                                                                                                  & ThinkThrice~\cite{wu2023deciphering}                 & \checkmark             &                & \checkmark       &                &                  &                      &                     & dynamic,single                         & UNL                                     \\ \cline{3-12}
                                                                                &                                                                                                  & CodeAct~\cite{shi2023cooperation}                    & \checkmark             &                & \checkmark       &                &                  &                      &                     & static,multi                           & UNL                                     \\ \cline{3-12}
                                                                                &                                                                                                  & wu et al.~\cite{wu2024enhance}                       & \checkmark             &                & \checkmark       &                &                  &                      &                     & static,multi                           & UNL                                     \\ \cline{3-12}
                                                                                &                                                                                                  & WWQA~\cite{du2024helmsman}                           & \checkmark             &                & \checkmark       & \checkmark     &                  &                      & \checkmark          & static,multi                           & UNL                                     \\ \cline{3-12}
                                                                                &                                                                                                  & PLAYER~\cite{zhu2024player}                          & \checkmark             &                & \checkmark       &                &                  &                      &                     & dynamic,multi                          & UNL                                     \\ \cline{3-12}
                                                                                &                                                                                                  & GITM~\cite{zhu2023ghost}                             & \checkmark             & \checkmark     & \checkmark       &                & \checkmark       &                      &                     & static,multi                           & UNL                                     \\ \cline{3-12}
                                                                                &                                                                                                  & sreedhar et al.~\cite{sreedhar2024simulating}        & \checkmark             & \checkmark     & \checkmark       &                &                  &                      &                     & static,single                          & UNL                                     \\ \cline{3-12}
                                                                                &                                                                                                  & AmongAgents~\cite{chi2024amongagents}                & \checkmark             & \checkmark     & \checkmark       &                &                  &                      &                     & static,multi                           & UNL                                     \\ \cline{3-12}
                                                                                &                                                                                                  & S-Agents~\cite{chen2024s}                            & \checkmark             & \checkmark     & \checkmark       &                & \checkmark       & \checkmark           &                     & dynamic,single                         & UNL                                     \\ \cline{1-12}
    \multirow{42}{*}{\begin{tabular}[c]{@{}c@{}}Task- \\ Driven\end{tabular}}   & \multirow{16}{*}{\begin{tabular}[c]{@{}c@{}}Foundational \\ and Applied \\ Science\end{tabular}} & VIDS~\cite{hassan2023chatgpt}                        & \checkmark             &                & \checkmark       &                &                  &                      &                     & dynamic,multi                          & UNL                                     \\ \cline{3-12}
                                                                                &                                                                                                  & DR-CoT~\cite{wu2023large}                            & \checkmark             &                & \checkmark       &                &                  &                      &                     & static,single                          & UNL                                     \\ \cline{3-12}
                                                                                &                                                                                                  & ChatGPT Research Group~\cite{zheng2023chatgpt}       & \checkmark             & \checkmark     & \checkmark       &                &                  & \checkmark           &                     & dynamic,multi                          & UNL                                     \\ \cline{3-12}
                                                                                &                                                                                                  & MedAgents~\cite{tang2023medagents}                   & \checkmark             &                & \checkmark       &                &                  &                      & \checkmark          & dynamic,multi                          & UNL,SL                                  \\ \cline{3-12}
                                                                                &                                                                                                  & MARG~\cite{d2024marg}                                & \checkmark             &                & \checkmark       &                &                  & \checkmark           &                     & static,multi                           & UNL                                     \\ \cline{3-12}
                                                                                &                                                                                                  & AI Hospital~\cite{fan2024ai}                         & \checkmark             &                & \checkmark       &                &                  &                      & \checkmark          & static,multi                           & UNL,SL                                  \\ \cline{3-12}
                                                                                &                                                                                                  & REVIEWER2~\cite{gao2024reviewer2}                    & \checkmark             &                &                  &                &                  &                      &                     & static,multi                           & UNL                                     \\ \cline{3-12}
                                                                                &                                                                                                  & CosmoAgent~\cite{jin2024if}                          & \checkmark             &                & \checkmark       &                &                  & \checkmark           &                     & dynamic,single                         & UNL                                     \\ \cline{3-12}
                                                                                &                                                                                                  & FPS~\cite{liu2024skepticism}                         & \checkmark             &                & \checkmark       &                &                  &                      &                     & dynamic,single                         & UNL                                     \\ \cline{3-12}
                                                                                &                                                                                                  & ResearchAgent~\cite{baek2024researchagent}           & \checkmark             &                &                  & \checkmark     &                  &                      &                     & static,multi                           & UNL                                     \\ \cline{3-12}
                                                                                &                                                                                                  & Agent Hospital~\cite{li2024agent}                    & \checkmark             &                & \checkmark       &                &                  &                      &                     & dynamic,multi                          & UNL,SL                                  \\ \cline{3-12}
                                                                                &                                                                                                  & CulturePark~\cite{li2024culturepark}                 & \checkmark             &                & \checkmark       &                &                  & \checkmark           &                     & dynamic,single                         & UNL                                     \\ \cline{3-12}
                                                                                &                                                                                                  & SynthPAI~\cite{yukhymenko2024synthetic}              & \checkmark             &                & \checkmark       &                &                  & \checkmark           &                     & dynamic,single                         & UNL                                     \\ \cline{3-12}
                                                                                &                                                                                                  & DreamFactory~\cite{xie2024dreamfactory}              & \checkmark             &                & \checkmark       &                &                  & \checkmark           &                     & static,multi                           & UNL,SL                                  \\ \cline{3-12}
                                                                                &                                                                                                  & AutoTQA~\cite{zhuautotqa}                            & \checkmark             & \checkmark     & \checkmark       & \checkmark     & \checkmark       &                      &                     & static,multi                           & UNL                                     \\ \cline{3-12}
                                                                                &                                                                                                  & DERA~\cite{nair2023dera}                             & \checkmark             &                & \checkmark       &                &                  &                      &  \checkmark         & static,single                          & UNL                                     \\ \cline{2-12}
                                                                                & \multirow{6}{*}{\begin{tabular}[c]{@{}c@{}}Software \\ Development\end{tabular}}                 & Self-collaboration~\cite{dong2023self}               & \checkmark             &                &                  &                & \checkmark       &                      &                     & dynamic,multi                          & UNL                                     \\ \cline{3-12}
                                                                                &                                                                                                  & ChatDev~\cite{qian2024chatdev}                       & \checkmark             & \checkmark     & \checkmark       & \checkmark     &                  &                      &                     & static,multi                           & UNL,SL                                  \\ \cline{3-12}
                                                                                &                                                                                                  & MetaGPT~\cite{hong2023metagpt}                       & \checkmark             & \checkmark     & \checkmark       & \checkmark     & \checkmark       & \checkmark           &                     & static,multi                           & SL                                      \\ \cline{3-12}
                                                                                &                                                                                                  & Experiential Co-Learning~\cite{qian2023experiential} & \checkmark             & \checkmark     & \checkmark       & \checkmark     &                  &                      &                     & dynamic,multi                          & UNL,SL                                  \\ \cline{3-12}
                                                                                &                                                                                                  & AutoCodeRover~\cite{zhang2024autocoderover}          & \checkmark             &                & \checkmark       & \checkmark     &                  &                      &                     & static,multi                           & UNL,SL                                  \\ \cline{3-12}
                                                                                &                                                                                                  & IER~\cite{qian2024iterative}                         & \checkmark             &                & \checkmark       &                &                  &                      &                     & dynamic,single                         & UNL,SL                                  \\ \cline{2-12}
                                                                                & \multirow{20}{*}{\begin{tabular}[c]{@{}c@{}}Other \\ Industries\end{tabular}}                    & Blind Judgement~\cite{hamilton2023blind}             & \checkmark             &                &                  &                &                  &                      &                     & static,single                          & UNL                                     \\ \cline{3-12}
                                                                                &                                                                                                  & TradingGPT~\cite{li2023tradinggpt}                   & \checkmark             & \checkmark     & \checkmark       &                &                  &                      &                     & dynamic,single                         & UNL                                     \\ \cline{3-12}
                                                                                &                                                                                                  & Information Bazaar~\cite{weissrethinking}            & \checkmark             &                & \checkmark       &                &                  &                      &                     & static,single                          & UNL                                     \\ \cline{3-12}
                                                                                &                                                                                                  & SimuCourt~\cite{he2024simucourt}                     & \checkmark             &                & \checkmark       &                &                  & \checkmark           & \checkmark          & static,multi                           & UNL,SL                                  \\ \cline{3-12}
                                                                                &                                                                                                  & MATHVC~\cite{yue2024mathvc}                          & \checkmark             &                & \checkmark       &                & \checkmark       &                      &                     & static,multi                           & UNL                                     \\ \cline{3-12}
                                                                                &                                                                                                  & baker et al.\cite{baker2024simulating}               & \checkmark             &                & \checkmark       &                &                  &                      &                     & static,multi                           & UNL                                     \\ \cline{3-12}
                                                                                &                                                                                                  & LawLuo~\cite{sun2024lawluo}                          & \checkmark             &                & \checkmark       &                &                  & \checkmark           &                     & dynamic,multi                          & UNL                                     \\ \cline{3-12}
                                                                                &                                                                                                  & MAIC~\cite{yu2024mooc}                               & \checkmark             & \checkmark     & \checkmark       &                &                  & \checkmark           &                     & dynamic,multi                          & UNL                                     \\ \cline{3-12}
                                                                                &                                                                                                  & CAMEL~\cite{li2023camel}                             & \checkmark             &                & \checkmark       &                & \checkmark       &                      &                     & static,single                          & UNL                                     \\ \cline{3-12}
                                                                                &                                                                                                  & SwiftSage~\cite{lin2024swiftsage}                    & \checkmark             & \checkmark     & \checkmark       &                &                  &                      &                     & static,single                          & UNL                                     \\ \cline{3-12}
                                                                                &                                                                                                  & Multi-Agent Collaboration~\cite{talebirad2023multi}  & \checkmark             & \checkmark     & \checkmark       & \checkmark     &                  &                      &                     & dynamic,single                         & UNL                                     \\ \cline{3-12}
                                                                                &                                                                                                  & CoELA~\cite{zhang2023building}                       & \checkmark             & \checkmark     & \checkmark       &                &                  &                      &                     & static,multi                           & UNL                                     \\ \cline{3-12}
                                                                                &                                                                                                  & RoCo~\cite{mandi2024roco}                            & \checkmark             & \checkmark     & \checkmark       &                &                  &                      &                     & static,single                          & UNL                                     \\ \cline{3-12}
                                                                                &                                                                                                  & AgentVerse~\cite{chen2023agentverse}                 & \checkmark             & \checkmark     & \checkmark       & \checkmark     & \checkmark       &                      &                     & dynamic,multi                          & UNL                                     \\ \cline{3-12}
                                                                                &                                                                                                  & Scalable~\cite{chen2024scalable}                     & \checkmark             & \checkmark     & \checkmark       &                & \checkmark       &                      &                     & dynamic,single                         & UNL                                     \\ \cline{3-12}
                                                                                &                                                                                                  & AutoAgents~\cite{chen2023autoagents}                 & \checkmark             &                & \checkmark       & \checkmark     & \checkmark       &                      &                     & dynamic,single                         & UNL                                     \\ \cline{3-12}
                                                                                &                                                                                                  & OpenAgents~\cite{xie2023openagents}                  & \checkmark             & \checkmark     & \checkmark       & \checkmark     &                  &                      &                     & dynamic,single                         & SL                                      \\ \cline{3-12}
                                                                                &                                                                                                  & TWOSOME~\cite{tan2024true}                           & \checkmark             & \checkmark     &                  &                &                  &                      &                     & static,single                          & -                                       \\ \cline{3-12}
                                                                                &                                                                                                  & ReAd~\cite{zhang2024towards}                         & \checkmark             & \checkmark     & \checkmark       &                & \checkmark       &                      &                     & dynamic,single                         & UNL                                     \\ \cline{3-12}
                                                                                &                                                                                                  & MACNET~\cite{qian2024scaling}                        & \checkmark             &                & \checkmark       &                &                  &                      &                     & dynamic,single                         & UNL                                     \\ \bottomrule
    \end{tabular}
}
\caption{A list of representative works of scenario simulation. UNL: unstructured natural language; SL: structured language.}
\label{tab: task_summary}
\end{table*}

In the real world, individuals do not function in isolation. They frequently engage in collaborative efforts to complete tasks within specific scenarios. This raises a crucial question: can LLM-based agents cooperate like humans or even surpass human performance in achieving collective intelligence? To answer this question, researchers simulate the interactions and collaborations of multiple individuals across various scenarios~\cite{qian2023communicative,hong2023metagpt,wu2023autogen}, ranging from everyday conversations to complex professional tasks, to enhance collective intelligence and problem-solving capabilities. A scenario simulation typically starts with designing a multi-agent system that includes constructing the scenario environment, modeling agent roles, and establishing organizational structures and communication protocols to manage interactions among agents.

In this section, we begin discussing the system composition of a scenario simulation with four key aspects in~\S\ref{subsec:sce_sys}. Following this, we summarize several scenarios that have recently attracted the attention of researchers in~\S\ref{subsec:sce_sce}. Finally, we review the methods and metrics commonly used for evaluating scenario simulations in~\S\ref{subsec:sce_eval}. The overall framework is presented in Figure~\ref{fig:fm_task} and representative works are summarized in Table~\ref{tab: task_summary}.

\subsection{System}\label{subsec:sce_sys}

The diversity of scenarios presents challenges in proposing a unified system applicable to scenarios. 
Most of the current systems can be summarized as \textit{``agents organized to play roles in dedicated environments through constrained communications"}.
Based on this general description, we identify four key concepts in scenario simulations: \textbf{environment}, \textbf{role}, \textbf{organization} and \textbf{communication}.

\subsubsection{Environment}

The environment in scenario simulation defines the specific contexts in which agents operate and interact with each other. Just as humans gather information from their surroundings, agents depend on the environment to receive input from various sources. These signals guide the behaviors and strategies of agents within the system. Thus, a comprehensive understanding of the environment paves the way for agents' decision-making and task continuity. We analyze the environment of existing work by focusing on four key aspects: \textit{configuration}, \textit{state}, \textit{history} and \textit{tools}.

\paragraph{Configuration}

The environment configuration provides basic information, especially essential elements necessary for the tasks and goals in the scenario.
The system will initialize agents accordingly so that they interact with clear objectives.
More specifically, an environment configuration may include \textit{events} in the environment and \textit{profiles} of agents.

\textit{{\ul Events}} are represented as a primary focus that needs to be resolved, such as the specific cases brought before the court~\cite{hamilton2023blind,he2024simucourt,sun2024lawluo,baker2024simulating}, and the topics that serve as the basis for multi-agent debates.~\cite{xiong2023examining,du2023improving,liang2023encouraging,chan2023chateval,wu2023autogen,xia2024measuring,ma2024debate}.

\textit{{\ul Profile}} refers to personalized information relevant to the agents specific to the scenario. Different from the basic attributes described in individual simulation, this module encompasses various aspects of the agents' identities, including their interests, goals, and roles~\cite{hong2023metagpt,yukhymenko2024synthetic,zhang2024self}.
Agents can also be configured to have access to external resources, such as related research papers~\cite{baek2024researchagent}, predefined strategies~\cite{zhang2024self} or disease information~\cite{li2024agent}.

\paragraph{State}

Environment states encompass the information provided by the environment during scenario execution (configurations are fixed at the beginning instead).
They directly influence the agents' decision-making and behavior.
According to how agents receive them, states can be further divided into \textit{observation} and \textit{feedback}.

\textit{{\ul Observation}} involves changes in the environment and the current state of surrounding entities. For example, properties and spatial positions~\cite{lin2024swiftsage,tan2024true,chen2024s,chen2024scalable} of other agents are provided to agents to inform real-time decision-making. Moreover, continuously updating agents' physical states are utilized to establish real-time spatial relationships with their environment and neighboring agents~\cite{zhu2023ghost,chen2024scalable,tan2024true,zhang2024towards}.

\textit{{\ul Feedback}} consists of responses received by agents after they perform actions, which guide future strategy adjustments. Some studies~\cite{talebirad2023multi,chen2024s,sreedhar2024simulating} describe how agents’ cognitive states and strategies are modified based on feedback after each interaction, allowing them to simulate human-like adaptability. Meanwhile, feedback on market events or decisions made by others~\cite{li2023tradinggpt,sreedhar2024simulating} and execution results from external tools~\cite{qian2024chatdev,hong2023metagpt,wu2023autogen} are provided, to facilitate strategy adjustment and guide future actions.

\paragraph{History}

As the scenario runs, past states and interactions accumulate into a series of history records.
Agents can leverage them to adapt to new situations and refine strategies, ensuring more coherent and effective task performance in dynamic environments.
We summarize four widely used methods to process and utilize the history, including \textit{direct integration}, \textit{refinement}, \textit{summarization} and \textit{memory mechanisms}.

\textit{{\ul Direct integration}} appends the history to the current input without modification. Agents may retain task continuity by incorporating past dialogue directly into the current session~\cite{du2023improving,liang2023encouraging,wu2023large,wu2023autogen}. Excessive content is truncated to fit token limits while preserving key historical information~\cite{chen2024scalable,xie2023openagents}.

\textit{{\ul Refinement}} iteratively updates and enhances responses based on the history. Ma et al.~\cite{ma2024debate} uses a subgraph-focusing mechanism to refine answers, allowing agents to optimize outcomes after each reasoning step. Similarly, Weiss et al.~\cite{weissrethinking} and D'Arcy et al.~\cite{d2024marg} iteratively improves initial answers to converge to more accurate results.

\textit{{\ul Summarization}} distills essential insights from the history. This can be achieved by synthesizing core actions from multiple plans to establish a reference for diverse scenarios~\cite{zhu2023ghost}, summarizing reports from multiple agents to consolidate findings~\cite{tang2023medagents}, and sharing key solutions subtasks~\cite{qian2024chatdev} to avoid lengthy dialogue histories.

\textit{{\ul Memory mechanisms}} process the history through agents' memory modules. This dynamic approach enables agents to preserve relevant information both within and across sessions \cite{li2023tradinggpt,chen2023autoagents,liu2024skepticism,qian2024iterative,qian2024scaling,xie2024dreamfactory,lin2023agentsims,li2023metaagents}. In addition, Hong et al.~\cite{hong2023metagpt} proposed shared message pools to further enhance communication efficiency, where agents exchange structured messages directly and retrieve information in a personalized manner.

\paragraph{Tools}

External tools offer specialized functionalities related to scenario simulation tasks, enabling more accurate and precise outcomes. The spectrum of tools utilized in scenario simulation encompasses a wide range, from programming languages such as Python and SQL to APIs facilitating external interactions. Generally, Python is mainly employed to execute and verify programmes~\cite{qian2024chatdev,hong2023metagpt,wu2023autogen}. SQL~\cite{zhuautotqa} and knowledge graphs query tools ~\cite{baek2024researchagent,ma2024debate} have been harnessed to retrieve external structured data. In certain scenarios, task-related tools such as calculators, predefined tools, and APIs~\cite{chen2023autoagents,xie2023openagents}  are also utilized to provide intermediate results, simplifying the processing workflow of agents.

\subsubsection{Role}

In scenario simulations, we assign agents distinct roles based on their tasks and functionalities.
As demonstrated in Figure~\ref{fig:fm_task}, there are two groups of roles in a typical setting: \textit{participants} carry out the tasks within the scenario, and \textit{directors} manage the task execution processes while providing necessary assistance.
Each role has its own responsibility that emphasizes different aspects of the system's operations.
They collaborate to achieve the system's overall goals.

\paragraph{Participants} 

Participants are the key members that actively engaged in task execution and discussion.
Their organization and communication are the core of task completion in scenario simulations.
Participants can be further classified into \textit{communicators} and \textit{workers} according to their tasks.

\textit{{\ul Communicators}} primarily focus on communication, such as information exchange, feedback, and task guidance. Specifically, this kind of agents can process information for certain disciplines and research applications~\cite{hamilton2023blind,nair2023dera} and advocate diverse viewpoints~\cite{xiong2023examining,hao2023chatllm}, claims~\cite{liang2023encouraging} and underlying needs~\cite{ataei2024elicitron,li2024culturepark}. 

\textit{{\ul Workers}} are directly involved in task execution and operations, demonstrating specialized skills and efficiency. This typically includes the common professional roles present in each scenario, such as coder and tester in software development~\cite{dong2023self}, buyer and seller in negotiations~\cite{fu2023improving}, doctors and medical professional agents in healthcare domain~\cite{wu2023large,li2024agent}, and receptionist, lawyer, and secretary in the legal contexts~\cite{sun2024lawluo}.

\paragraph{Directors}

While participants execute most of the tasks, directors can provide essential support in crucial aspects such as planning procedures, coordinating communication, and integrating results.
We name them \textit{Planners}, \textit{Coordinators} and \textit{Integrators} respectively.

\textit{{\ul Planners}} play a vital role in task definition and strategic formulation, facilitating effective inter-agent collaboration through tasks such as defining objectives, analyzing user requirements, and optimizing execution plans. Task-specific agents~\cite{li2023camel}, central planners~\cite{chen2023agentverse}, analysts~\cite{dong2023self} and decomposer~\cite{zhu2023ghost} are responsible for breaking down requirements and dividing overarching objectives into specific sub-goals. Product managers~\cite{hong2023metagpt} contribute by creating detailed product requirements documents. Other planners can also refine execution plans according to task requirements~\cite{chen2024scalable}, optimize the process by maximizing the advantage function~\cite{chen2024s} and develop plans based on user inquiries~\cite{zhuautotqa}.

\textit{{\ul Coordinators}} are responsible for managing and coordinating the collaboration between agents to ensure effective task execution, monitor progress, and facilitate cooperation. The project managers~\cite{hong2023metagpt,zheng2023chatgpt} in software development oversee task distribution and project progress, ensuring efficient collaboration among team members throughout the development cycle. Judge assistant agents~\cite{he2024simucourt} aids in organizing information during court proceedings, and the main contact agents~\cite{li2024culturepark} manage intercultural conversations. Additionally, the secretary agents~\cite{jin2024if} manage interactions among civilization agents.
Meanwhile, coordinators also provide feedback to guide better interactions. Critic agents~\cite{fu2023improving} evaluate negotiation strategies and guide agents through iterative learning processes. Judge agents~\cite{xiong2023examining,liang2023encouraging,liang2024debatrix} serve as an authoritative evaluator, assessing arguments and performances during debates.

\textit{{\ul Integrators}} encompass various decision-making and summarization functions critical for guiding the system's trajectory. Deciders~\cite{nair2023dera} autonomously evaluate contributions from the researcher to make informed judgments on the dialogue's outcome. Summarizer agents~\cite{chan2023chateval} enhance communication clarity by providing concise summaries of discussions after each iteration, effectively integrating key points into the ongoing dialogue. In medical scenarios, medical report assistants~\cite{tang2023medagents} compile analyses into a cohesive document that supports collaborative expert discussions, while the medical decision maker ensures that final decisions reflect the collective expertise of the specialists involved. Additionally, the chief physician~\cite{fan2024ai} evaluates diagnostic performance based on accuracy and effectiveness, reinforcing the system's overall reliability. In legal contexts, the judge~\cite{he2024simucourt} oversees judicial processes, making critical decisions grounded in legal arguments and assessing the evidence presented.

\subsubsection{Organization}
Effective task execution necessitates careful coordination and scheduling of the interactions between individual agents. The organizational structures establish how each agent collaborates with others to achieve a goal. Typically, we can depict an organization schema by its \textit{mode} and \textit{structure}.

\paragraph{Mode}
The organizational structure determines whether the relationships among agents remain stable or evolve dynamically throughout the simulation process. In terms of how to organize agents, there are mainly two modes in existing research, i.e., \textit{static} and \textit{dynamic mode}.

\textit{{\ul Static mode}} refers to the organizational structure predefined based on the nature of the tasks. Agents communicate and work in an orderly manner according to these static structures. The static mode can be further divided into single-stage and multi-stage setups. In the single-stage setup, agents follow a fixed structure in multiple rounds of communication, such as structured debates~\cite{nair2023dera,li2023camel,fu2023improving,chan2023chateval}, skill training~\cite{yang2024social,yan2024social} and integrating ideas \cite{hamilton2023blind,hao2023chatllm}. In the multi-stage setup, tasks are divided into distinct stages, and the organization may change with stages. This can be found in the design, coding, and testing stages in software development scenarios following the waterfall model or standardized operating procedures~\cite{qian2024chatdev,hong2023metagpt}, and multi-stage process in judicial scenarios~\cite{he2024simucourt,baker2024simulating} and problem-solving processes~\cite{zhu2023ghost,zhang2023building,ma2024debate}.

\textit{{\ul Dynamic mode}} explores more open and adaptive organizational structures, often relying on dynamic and heuristic communication. This also includes both single-stage and multi-stage setups.  The single-stage setup emphasizes agent collaboration and adaptability in a single stage. The agents can be flexibly created and recruited~\cite{xie2023openagents,chen2023autoagents,ma2024debate,chen2023agentverse,liu2024autonomous}, coordinated through liaison agents~\cite{jin2024if,li2024culturepark}, and self-organized~\cite{chen2024s}. The multi-stage setup mainly features dynamic discussions among agents. Agents can be involved across multiple stages, but they can communicate autonomously based on the current state~\cite{dong2023self,zheng2023chatgpt,tang2023medagents,sun2024lawluo,yu2024mooc}.

\paragraph{Structure} 
The organization structure, meanwhile, reflects how agents are connected with each other. Typically, an organization can be \textit{layered}, \textit{centralized} or \textit{decentralized}.
\textit{{\ul Layered}} structures adopt a hierarchical framework, with agents assigned to distinct levels. Interactions are predominantly confined to agents within the same level or occur between adjacent layers, thereby facilitating a controlled and organized flow of information\cite{hamilton2023blind,hao2023chatllm,qian2024chatdev}.
\textit{{\ul Centralized}} structures often involve a high-level role (e.g., coordinator) that serves as the core of the organization, overseeing communication and functioning as the central hub for interactions among other agents\cite{jin2024if,li2024culturepark,fan2024ai}.
\textit{{\ul Decentralized}} structures, in contrast, is more flattened, where agents can engage in peer-to-peer interactions as needed\cite{chan2023chateval,ma2024debate,liang2023encouraging}.

\subsubsection{Communication}
The communication between agents controls the transmission of information. To better understand the internal mechanism of communication, we dissect communication from its \textit{format} and \textit{style}.

\paragraph{Format}
From the perspective of information format, there exist two common communication protocols: \textit{unstructured natural language} and \textit{structured language}.

\textit{{\ul Unstructured natural language}} is most commonly used in multi-agent communication, enabling flexible and immediate exchanges through free-form, conversational language that mirrors human dialogue~\cite{nair2023dera,li2023camel,fu2023improving,xiong2023examining,du2023improving,zheng2023chatgpt,yang2024social,yan2024social}. Communication based on natural language is diverse and flexible, but it can also suffer from issues such as ambiguity and redundancy.

\textit{{\ul Structured language}}, such as code and JSON documents, is another protocol that may alleviate the issues from natural language.
In software development, agents transit information between phases through code~\cite{hong2023metagpt,qian2024chatdev}. In the medical domain, structured summaries of reports are utilized to gain key insights~\cite{tang2023medagents}. In addition to predefined formats, agents can also autonomously choose the appropriate format during interactions to improve efficiency~\cite{chen2024beyond,chen2024optima}. Recently, more complex communication protocols using more than one language have been designed to improve communication~\cite{marro2024scalable}.

\paragraph{Style}
Communication, by nature, can be \textit{cooperative} or \textit{competitive} regarding its style.
In \textit{{\ul cooperative}} communication, agents share a common objective, aiming to optimize collective outcomes, like software development\cite{dong2023self,qian2024chatdev,hong2023metagpt}, medical diagnosis\cite{tang2023medagents,fan2024ai}, and case handling\cite{hamilton2023blind,sun2024lawluo}.
In contrast, agents in \textit{{\ul competitive}} communication typically hold differing viewpoints and positions, each striving to achieve individual objectives. Such scenarios are commonly found in settings like games\cite{xu2023exploring,du2024helmsman,wang2023avalon} and debates\cite{xiong2023examining,liang2023encouraging,fu2023improving}, where agents maintain opposing stances and seek to outmaneuver each other.

\subsection{Scenario}\label{subsec:sce_sce}
Using the collective capabilities of agents with specialized expertise, scenario simulations have been applied to various domains. Here we divide different scenarios into two groups: \textbf{dialog-driven} ones that cover social interaction and question-answering, and \textbf{task-driven} ones that focus on specialized tasks.

\subsubsection{Dialog-Driven Scenario}
Dialog-driven scenarios encompass scenarios in people's daily lives where the dialog itself is centered, such as those for social or entertainment purposes.
These scenarios share a common emphasis on tackling general goals that are not related to any specific task or domain.
We identify three primary types of dialog-driven scenarios: \textit{social interaction}, \textit{question-answering}, and \textit{game} scenarios.

\paragraph{Social Interaction}
Some works focus on task completion in simple social interaction scenarios, typically involving social tasks between two or a few agents, such as persuasion or comforting a partner. Zhou et al.~\cite{zhou2023sotopia} discusses the social intelligence of agents in social scenarios, revealing significant performance differences among models across different dimensions. The exploration in social intelligence is further extended to objective action-level evaluation~\cite{wang2024towards} and diverse scenarios and others' information reasoning\cite{mou2024agentsense}. Furthermore, some works propose interactive learning methods~\cite{yang2024social,wang2024sotopia,zhou2024real} to help learn social skills.

\paragraph{Question Answering}
Another mainstream scenario is the question answering, emphasizing collaborative processes, strategic reasoning, and integration to enhance model performance. On the one hand, some studies focus on improving reasoning through debate. FORD~\cite{xiong2023examining} facilitates a three-stage commonsense reasoning debate, demonstrating that LLMs can reach consensus even amidst inconsistencies. MAD~\cite{du2023improving}, involves agents debating under a judge's supervision, addressing the Degeneration-of-Thought problem. In addition, a ``society of minds" approach~\cite{du2023improving} is presented to guide multiple debate rounds, improving mathematical reasoning and factual accuracy while reducing hallucinations. On the other hand, some works focus on optimizing strategies in strategic reasoning and negotiation. OG-Narrator~\cite{xia2024measuring} is proposed to improve negotiation strategies, increasing the Buyers' deal success rates. Ma et al.~\cite{ma2024debate} utilize a subgraph-focusing mechanism and a multi-role debate team to improve reasoning accuracy and reliability, outperforming existing methods.

\paragraph{Game}
Games provide a unique platform for exploring scenario simulation, evolving from basic game reproduction to complex social dynamics.
Early studies, such as \cite{xu2023exploring,wang2023avalon}, introduce Werewolf and Avalon to examine LLM performance in communication games, specifically investigating how LLMs handle aspects like trust and leadership. 
Building on these complex interactions, reinforcement learning frameworks in \cite{xu2023language,wu2024enhance} allow agents to adapt their strategies, achieving near-human-level decision-making.
To explore deeper social phenomena, \cite{wu2024enhance,zhu2024player} expand on game dynamics by incorporating tools that enhance memory, reasoning, and adaptability. 
Additionally, \cite{du2024helmsman} examines the role of opinion leadership, while \cite{wu2023deciphering,shi2023cooperation,gong2023mindagent} tackle ad hoc teamwork, where agents adapt and collaborate without predefined protocols, revealing both the challenges and potential of LLM agents in team-based collaboration.

\subsubsection{Task-Driven Scenario}
In task-driven scenarios, agents role-play personas with specific functions for a certain task or task-set.
Most of these scenarios fall into one or more specific domains related to the tasks.
Here, agents are increasingly leveraged to solve complex, domain-specific problems by automating tasks and improving decision-making processes.

\paragraph{Foundational and Applied Science}
Science domains, such as medicine, mathematics, data science, and content analysis, have been popular experimental fields for scenario simulation. In the medical domain, medical reasoning and automating diagnostic processes have been refined through innovative methodologies such as chain-of-thought prompting and multi-agent collaboration\cite{wu2023large,tang2023medagents,li2024agent,bao2024piors}.
Zheng et al.\cite{zheng2023chatgpt} integrates ChatGPT with Bayesian optimization techniques to enhance research workflows in chemistry laboratories, demonstrating significant improvements in efficiency and productivity. Hassan et al.\cite{hassan2023chatgpt} introduce a conversational framework that enables seamless interaction with machine learning models, specifically for tasks like data visualization and predictive analytics. These studies demonstrate the potential of LLM-based agents to transform traditional research patterns.

\paragraph{Software Development}
Recent research has increasingly focused on harnessing agents to address complex challenges in software development and life-cycle management. Early works focus on designing frameworks for collaborative code generation. Dong et al.~\cite{dong2023self} presents a self-collaboration framework where LLM agents function as distinct ``experts," each managing specific subtasks to facilitate autonomous collaborative code generation. Building on this, ChatDev\cite{qian2024chatdev}, a chat-powered framework utilizes unified language-based communication among agents to effectively address design, coding, and testing phases. Meanwhile, Hong et al.~\cite{hong2023metagpt} enhances LLM collaborations by encoding Standardized Operating Procedures into prompts, enabling agents to verify results and produce coherent solutions through an assembly line approach. Afterward, some works focus on enabling agents to learn from past experiences and refine their processes over time
~\cite{qian2023experiential,qian2024iterative}. Further efforts focus on autonomous issue resolution and program understanding~\cite{zhang2024autocoderover}. These studies show the potential of multi-agent collaboration in software engineering, offering robust tools for automatic development and management.

\paragraph{Other Industries}
In the realm of broad social science, several studies leverage multi-agent systems to enhance decision-making processes across diverse fields, such as journalism~\cite{liu2024aipress}, judiciary, economics, and education. In the judicial field, legal consultations have been improved through LawLuo~\cite{sun2024lawluo}, which simulates collaborative discussions. Hamilton et al.~\cite{hamilton2023blind} and He et al.~\cite{he2024simucourt} design multi-agent systems to simulate U.S. Supreme Court decisions and court trials through detailed steps such as debate, resource retrieval, and decision refinement, complemented by additional benchmarks that enhance legal article generation. In the economic sector, Li et al.~\cite{li2023tradinggpt} propose a multi-agent framework with layered memory to improve LLM performance in stock trading. Additionally, Weiss et al.~\cite{weissrethinking} address the buyer's inspection paradox in information markets by simulating a marketplace where intelligent agents use LLMs to navigate information access and biases, exploring the impact of pricing and budgets on outcomes. In the education domain, MAIC~\cite{yu2024mooc}, a system simulating AI-enhanced classrooms has contributed to the development of a comprehensive AI-driven online education platform. Yue et al.~\cite{yue2024mathvc} presents MATHVC, an LLM-driven virtual classroom designed to simulate interactions among students, thereby fostering the development of mathematical skills.

\begin{figure*}[!t]
    \centering
    \includegraphics[width=\linewidth]{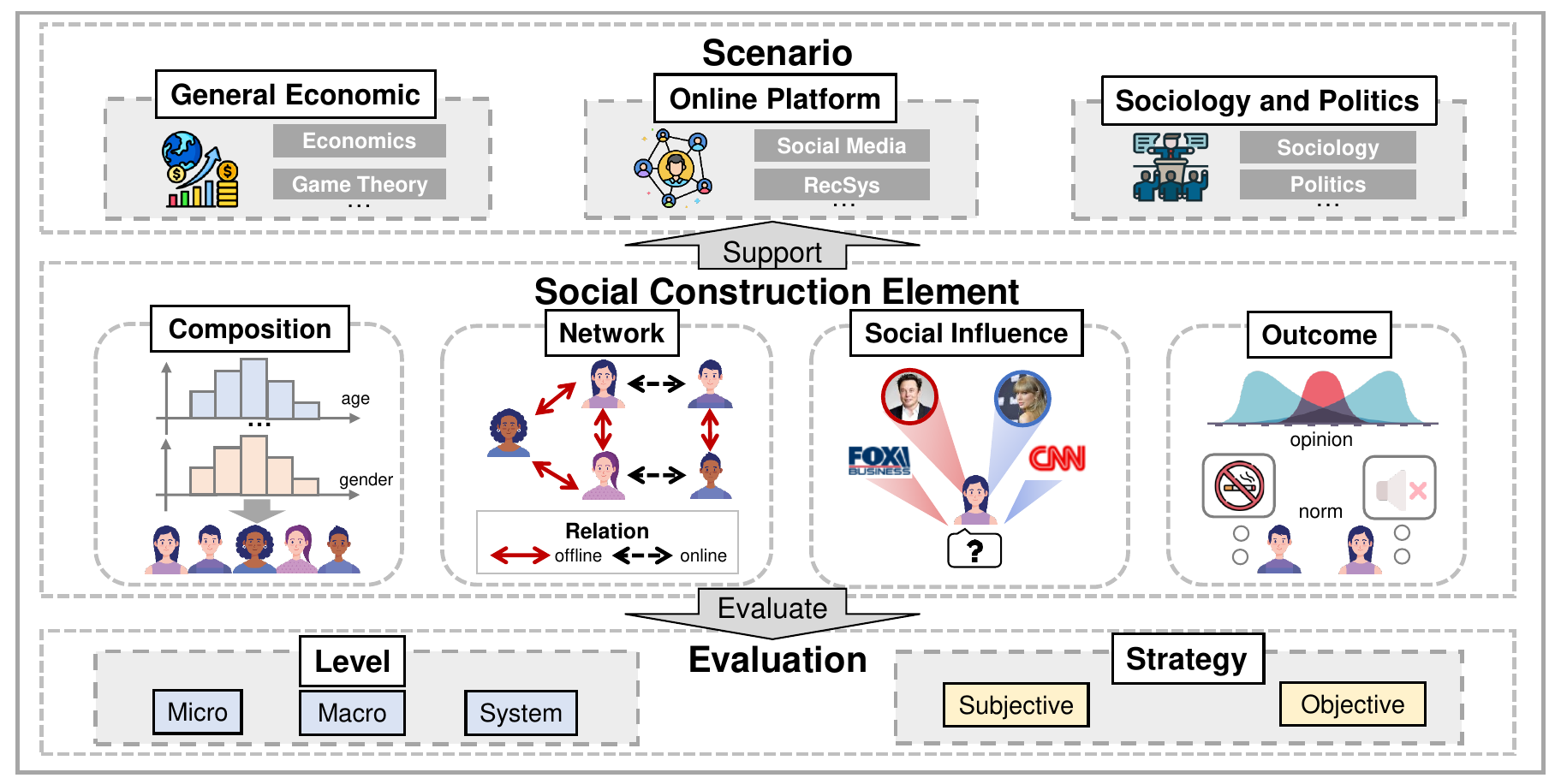}
    \caption{Illustration of society simulations. To construct society simulations, the corresponding society's {\ul construction elements}, i.e., composition, network, social influence and outcomes need to be carefully designed. Building on this, various {\ul scenarios} can be simulated. The performance of individuals and the overall performance of the system are {\ul evaluated}.}
    \label{fig:fm_soc}
\end{figure*}

\subsection{Evaluation}\label{subsec:sce_eval}
For scenario simulations, the evaluation focuses on how well the tasks of the scenarios are solved. Based on the scope of the evaluation, it can be categorized into \textbf{task evaluation}, \textbf{sub-task evaluation} and \textbf{system evaluation}, each employing various \textit{{automatic}}, \textit{{LLM-based}}, and \textit{{human}} evaluation methods to assess performance.

\paragraph{Task Evaluation}
Task Evaluation measures the overall performance of tasks assigned to the scenario. The evaluation can carried out in automatic ways or by LLMs or humans. In terms of \textit{{\ul automatic}} evaluation, predefined metrics and mathematical tools are used to objectively assess the task outcomes, such as accuracy~\cite{hamilton2023blind,xiong2023examining}, pass@k~\cite{li2023camel} for coding tasks, success rate, and coverage for exploration~\cite{zhu2023ghost}, and deal price for negotiation~\cite{fu2023improving}. These methods are efficient and scalable but may overlook complex behaviors. Thus, \textit{{\ul LLMs}}~\cite{hao2023chatllm} and \textit{{\ul human }}experts~\cite{li2023camel,liang2023encouraging} have been applied to provide more nuanced evaluation for qualitative tasks and compare solutions based on specific criteria.

\paragraph{Sub-Task Evaluation}
Sub-task Evaluation assesses the completion of sub-tasks within a scenario simulation and their impact on overall task performance. It serves as a process evaluation for the execution of complex tasks. The \textit{{\ul automatic}} evaluation uses metrics like transport rate, average steps, task success rate, re-plan attempts, and efficiency improvement to assess sub-task performance and strategy efficiency~\cite{zhang2023building,mandi2024roco}. Completeness, executability, and consistency metrics are often applied in software generation tasks~\cite{qian2024chatdev,qian2023experiential}. \textit{{\ul LLM-based}} evaluation focuses on pairwise comparisons or win rate judgments, capturing qualitative aspects of sub-task performance~\cite{qian2024chatdev}. Meanwhile, \textit{{\ul human}} evaluation relies on participants to provide subjective assessments on metrics such as executability, revision costs, or comment quality, offering practical insights into sub-task performance~\cite{hong2023metagpt,d2024marg}.

\paragraph{System Evaluation}
System Evaluation aims to capture the effectiveness and efficiency of the system in a scenario simulation as a whole. \textit{{\ul Automatic}} evaluation relies on metrics such as token consumption, task success rate, and human-likeness scores to measure the efficiency and realism of agents~\cite{tan2024true}. Additional metrics like accuracy, precision, recall, and F1 scores are used to assess system accuracy and consistency in diagnostic or predictive tasks~\cite{fan2024ai}. \textit{{\ul LLM-based}} evaluation often involves GPT-4 to assess qualitative aspects, such as human-likeness or diagnostic report quality~\cite{tan2024true,li2024agent}. \textit{{\ul Human}} evaluation typically involves subjective assessments, such as rating instructional content for tone, clarity, and supportiveness on a Likert scale~\cite{yu2024mooc}, often used to complement automatic methods and capture human perspectives on system outputs.

\section{Society Simulation}~\label{sec:soc}

\begin{table*}[!htp]
\setlength{\tabcolsep}{2pt}
\centering
\resizebox*{!}{\dimexpr\textheight-2\baselineskip\relax}{
\begin{tabular}{c|c|c|c|cccc}
\toprule
\multirow{2}{*}{\textbf{Scenario}}                                                            & \multirow{2}{*}{\textbf{Field}}                                                                                     & \multirow{2}{*}{\textbf{Paper}}                            & \multirow{2}{*}{\textbf{\# Agents}} & \multicolumn{4}{c}{\textbf{Construction Element}}                                                                                                     \\ \cline{5-8} 
                                                                                              &                                                                                                                     &                                                            &                                     & \multicolumn{1}{c|}{\textbf{Composition}} & \multicolumn{1}{c|}{\textbf{Network}} & \multicolumn{1}{c|}{\textbf{Social Influence}} & \textbf{Outcome} \\ \midrule
\multirow{17}{*}{\begin{tabular}[c]{@{}c@{}}General\\ Economic\end{tabular}}                  & \multirow{10}{*}{\begin{tabular}[c]{@{}c@{}}Game Theory\\ and\\ Strategic Interactions\end{tabular}}                & Agent-trust~\cite{xie2024can}                              & $(0, 10]$                           & \multicolumn{1}{c|}{\checkmark}           & \multicolumn{1}{c|}{}                 & \multicolumn{1}{c|}{\checkmark}                & \checkmark       \\ \cline{3-8} 
                                                                                              &                                                                                                                     & LELMA~\cite{mensfelt2024logic}                             & $(0, 10]$                           & \multicolumn{1}{c|}{}                     & \multicolumn{1}{c|}{}                 & \multicolumn{1}{c|}{\checkmark}                & \checkmark       \\ \cline{3-8} 
                                                                                              &                                                                                                                     & Economics   Arena~\cite{guo2024economics}                  & $(0, 10]$                           & \multicolumn{1}{c|}{}                     & \multicolumn{1}{c|}{}                 & \multicolumn{1}{c|}{\checkmark}                & \checkmark       \\ \cline{3-8} 
                                                                                              &                                                                                                                     & Fontana et   al.~\cite{fontana2024nicer}                   & $(0, 10]$                           & \multicolumn{1}{c|}{}                     & \multicolumn{1}{c|}{}                 & \multicolumn{1}{c|}{\checkmark}                & \checkmark       \\ \cline{3-8} 
                                                                                              &                                                                                                                     & SABM~\cite{han2023guinea}                                  & $(0, 10]$                           & \multicolumn{1}{c|}{\checkmark}           & \multicolumn{1}{c|}{\checkmark}       & \multicolumn{1}{c|}{\checkmark}                & \checkmark       \\ \cline{3-8} 
                                                                                              &                                                                                                                     & Noh and   Chang.~\cite{noh2024llms}                        & $(0, 10]$                           & \multicolumn{1}{c|}{\checkmark}           & \multicolumn{1}{c|}{}                 & \multicolumn{1}{c|}{\checkmark}                & \checkmark       \\ \cline{3-8} 
                                                                                              &                                                                                                                     & Mozikov et   al.~\cite{mozikov2024good}                    & $(0, 10]$                           & \multicolumn{1}{c|}{}                     & \multicolumn{1}{c|}{}                 & \multicolumn{1}{c|}{\checkmark}                & \checkmark       \\ \cline{3-8} 
                                                                                              &                                                                                                                     & Wu et al.~\cite{wu2024shall}                               & $(10, 100]$                         & \multicolumn{1}{c|}{}                     & \multicolumn{1}{c|}{\checkmark}       & \multicolumn{1}{c|}{\checkmark}                & \checkmark       \\ \cline{3-8} 
                                                                                              &                                                                                                                     & CompeteAI~\cite{zhaocompeteai}                             & $(10, 100]$                         & \multicolumn{1}{c|}{\checkmark}           & \multicolumn{1}{c|}{\checkmark}       & \multicolumn{1}{c|}{\checkmark}                & \checkmark       \\ \cline{3-8} 
                                                                                              &                                                                                                                     & WarAgent~\cite{hua2023war}                                 & $(10, 100]$                         & \multicolumn{1}{c|}{\checkmark}           & \multicolumn{1}{c|}{\checkmark}       & \multicolumn{1}{c|}{\checkmark}                & \checkmark       \\ \cline{2-8} 
                                                                                              & \multirow{7}{*}{\begin{tabular}[c]{@{}c@{}}Economic \\ Contexts\end{tabular}}                                       & Horton~\cite{horton2023large}                              & $(10, 100]$                         & \multicolumn{1}{c|}{\checkmark}           & \multicolumn{1}{c|}{}                 & \multicolumn{1}{c|}{}                          & \checkmark       \\ \cline{3-8} 
                                                                                              &                                                                                                                     & EconAgent~\cite{li2024econagent}                           & $(10, 100]$                         & \multicolumn{1}{c|}{\checkmark}           & \multicolumn{1}{c|}{}                 & \multicolumn{1}{c|}{}                          & \checkmark       \\ \cline{3-8} 
                                                                                              &                                                                                                                     & SRAP-Agent~\cite{ji2024srap}                               & $(10, 100]$                         & \multicolumn{1}{c|}{\checkmark}           & \multicolumn{1}{c|}{\checkmark}       & \multicolumn{1}{c|}{\checkmark}                & \checkmark       \\ \cline{3-8} 
                                                                                              &                                                                                                                     & Ghaffarzadegan et   al.\cite{ghaffarzadegan2023generative} & $(10, 100]$                         & \multicolumn{1}{c|}{\checkmark}           & \multicolumn{1}{c|}{}                 & \multicolumn{1}{c|}{\checkmark}                & \checkmark       \\ \cline{3-8} 
                                                                                              &                                                                                                                     & EC~\cite{de2023emergent}                                   & $(10, 100]$                         & \multicolumn{1}{c|}{\checkmark}           & \multicolumn{1}{c|}{\checkmark}       & \multicolumn{1}{c|}{\checkmark}                & \checkmark       \\ \cline{3-8} 
                                                                                              &                                                                                                                     & Williams et   al.~\cite{williams2023epidemic}              & $(100, \infty)$                     & \multicolumn{1}{c|}{\checkmark}           & \multicolumn{1}{c|}{\checkmark}       & \multicolumn{1}{c|}{\checkmark}                & \checkmark       \\ \cline{3-8} 
                                                                                              &                                                                                                                     & AgentTorch~\cite{chopra2024limits}                         & $(100, \infty)$                     & \multicolumn{1}{c|}{\checkmark}           & \multicolumn{1}{c|}{\checkmark}       & \multicolumn{1}{c|}{}                          & \checkmark       \\ \hline
\multirow{24}{*}{\begin{tabular}[c]{@{}c@{}}Sociology\\ and\\ Political Science\end{tabular}} & \multirow{7}{*}{\begin{tabular}[c]{@{}c@{}}Public Opinion\\ Survey\end{tabular}}                                    & Argyle et al.~\cite{Argyle_2023}                           & $(100, \infty)$                     & \multicolumn{1}{c|}{\checkmark}           & \multicolumn{1}{c|}{}                 & \multicolumn{1}{c|}{}                          & \checkmark       \\ \cline{3-8} 
                                                                                              &                                                                                                                     & Lee et al.~\cite{lee2023can}                               & $(100, \infty)$                     & \multicolumn{1}{c|}{\checkmark}           & \multicolumn{1}{c|}{}                 & \multicolumn{1}{c|}{}                          & \checkmark       \\ \cline{3-8} 
                                                                                              &                                                                                                                     & Chaudhary and   Chaudhary~\cite{chaudhary2024large}        & $(100, \infty)$                     & \multicolumn{1}{c|}{\checkmark}           & \multicolumn{1}{c|}{}                 & \multicolumn{1}{c|}{}                          & \checkmark       \\ \cline{3-8} 
                                                                                              &                                                                                                                     & ElectionSim~\cite{zhang2024electionsim}                    & $(100, \infty)$                     & \multicolumn{1}{c|}{\checkmark}           & \multicolumn{1}{c|}{}                 & \multicolumn{1}{c|}{}                          & \checkmark       \\ \cline{3-8} 
                                                                                              &                                                                                                                     & GABSS~\cite{xiao2023simulating}                            & $(100, \infty)$                     & \multicolumn{1}{c|}{\checkmark}           & \multicolumn{1}{c|}{\checkmark}       & \multicolumn{1}{c|}{\checkmark}                & \checkmark       \\ \cline{3-8} 
                                                                                              &                                                                                                                     & Park et   al.~\cite{park2024generative}                    & $(100, \infty)$                     & \multicolumn{1}{c|}{\checkmark}           & \multicolumn{1}{c|}{}                 & \multicolumn{1}{c|}{}                          & \checkmark       \\ \cline{3-8} 
                                                                                              &                                                                                                                     & Sun et   al.~\cite{sun2024randomsiliconsamplingsimulating} & $(100, \infty)$                     & \multicolumn{1}{c|}{\checkmark}           & \multicolumn{1}{c|}{}                 & \multicolumn{1}{c|}{}                          & \checkmark       \\ \cline{2-8} 
                                                                                              & \multirow{17}{*}{\begin{tabular}[c]{@{}c@{}}Individual \\ and\\ Organizational\\ Behavior Observation\end{tabular}} & Aher et al.~\cite{aher2023using}                           & $(0, 10]$                           & \multicolumn{1}{c|}{}                     & \multicolumn{1}{c|}{}                 & \multicolumn{1}{c|}{\checkmark}                & \checkmark       \\ \cline{3-8} 
                                                                                              &                                                                                                                     & Zhang et   al.~\cite{zhang2023exploring}                   & $(0, 10]$                           & \multicolumn{1}{c|}{\checkmark}           & \multicolumn{1}{c|}{}                 & \multicolumn{1}{c|}{}                          & \checkmark       \\ \cline{3-8} 
                                                                                              &                                                                                                                     & Lyfe   Agents~\cite{kaiya2023lyfe}                         & $(0, 10]$                           & \multicolumn{1}{c|}{\checkmark}           & \multicolumn{1}{c|}{\checkmark}       & \multicolumn{1}{c|}{\checkmark}                & \checkmark       \\ \cline{3-8} 
                                                                                              &                                                                                                                     & CRSEC~\cite{ren2024emergence}                              & $(0, 10]$                           & \multicolumn{1}{c|}{\checkmark}           & \multicolumn{1}{c|}{\checkmark}       & \multicolumn{1}{c|}{\checkmark}                & \checkmark       \\ \cline{3-8} 
                                                                                              &                                                                                                                     & Chuang et   al.\cite{chuang2023simulating}                 & $(0, 10]$                           & \multicolumn{1}{c|}{\checkmark}           & \multicolumn{1}{c|}{\checkmark}       & \multicolumn{1}{c|}{\checkmark}                & \checkmark       \\ \cline{3-8} 
                                                                                              &                                                                                                                     & ChoiceMates~\cite{park2023choicemates}                     & $(0, 10]$                           & \multicolumn{1}{c|}{\checkmark}           & \multicolumn{1}{c|}{\checkmark}       & \multicolumn{1}{c|}{\checkmark}                & \checkmark       \\ \cline{3-8} 
                                                                                              &                                                                                                                     & Jarrett et   al.\cite{jarrett2023language}                 & $(0, 10]$                           & \multicolumn{1}{c|}{\checkmark}           & \multicolumn{1}{c|}{}                 & \multicolumn{1}{c|}{}                          & \checkmark       \\ \cline{3-8} 
                                                                                              &                                                                                                                     & AgentReview~\cite{jin2024agentreview}                      & $(0, 10]$                           & \multicolumn{1}{c|}{\checkmark}           & \multicolumn{1}{c|}{\checkmark}       & \multicolumn{1}{c|}{}                          & \checkmark       \\ \cline{3-8} 
                                                                                              &                                                                                                                     & Generative   Agents~\cite{park2023generative}              & $(10, 100]$                         & \multicolumn{1}{c|}{\checkmark}           & \multicolumn{1}{c|}{\checkmark}       & \multicolumn{1}{c|}{\checkmark}                & \checkmark       \\ \cline{3-8} 
                                                                                              &                                                                                                                     & AGA~\cite{yu2024affordable}                                & $(10, 100]$                         & \multicolumn{1}{c|}{\checkmark}           & \multicolumn{1}{c|}{\checkmark}       & \multicolumn{1}{c|}{\checkmark}                & \checkmark       \\ \cline{3-8} 
                                                                                              &                                                                                                                     & MineLand~\cite{yu2024mineland}                             & $(10, 100]$                         & \multicolumn{1}{c|}{\checkmark}           & \multicolumn{1}{c|}{\checkmark}       & \multicolumn{1}{c|}{\checkmark}                & \checkmark       \\ \cline{3-8} 
                                                                                              &                                                                                                                     & Chuang et   al.~\cite{chuang2024wisdom}                    & $(10, 100]$                         & \multicolumn{1}{c|}{\checkmark}           & \multicolumn{1}{c|}{\checkmark}       & \multicolumn{1}{c|}{\checkmark}                & \checkmark       \\ \cline{3-8} 
                                                                                              &                                                                                                                     & CareerAgent~\cite{zhu2024generative}                       & $(10, 100]$                         & \multicolumn{1}{c|}{\checkmark}           & \multicolumn{1}{c|}{\checkmark}       & \multicolumn{1}{c|}{\checkmark}                & \checkmark       \\ \cline{3-8} 
                                                                                              &                                                                                                                     & Suzuki and   Arita~\cite{suzuki2024evolutionary}           & $(10, 100]$                         & \multicolumn{1}{c|}{\checkmark}           & \multicolumn{1}{c|}{}                 & \multicolumn{1}{c|}{\checkmark}                & \checkmark       \\ \cline{3-8} 
                                                                                              &                                                                                                                     & Chuang et   al.\cite{chuang2024beyond}                     & $(100, \infty)$                     & \multicolumn{1}{c|}{\checkmark}           & \multicolumn{1}{c|}{}                 & \multicolumn{1}{c|}{}                          & \checkmark       \\ \cline{3-8} 
                                                                                              &                                                                                                                     & Li et   al.~\cite{li2023quantifying}                       & $(100, \infty)$                     & \multicolumn{1}{c|}{}                     & \multicolumn{1}{c|}{\checkmark}       & \multicolumn{1}{c|}{\checkmark}                & \checkmark       \\ \cline{3-8} 
                                                                                              &                                                                                                                     & MATRIX~\cite{tang2024synthesizing}                         & $(100, \infty)$                     & \multicolumn{1}{c|}{\checkmark}           & \multicolumn{1}{c|}{\checkmark}       & \multicolumn{1}{c|}{}                          &                  \\ \hline
\multirow{19}{*}{\begin{tabular}[c]{@{}c@{}}Online\\ Platform\end{tabular}}                   & \multirow{14}{*}{\begin{tabular}[c]{@{}c@{}}Social\\ Platforms\end{tabular}}                                        & Cai et al.\cite{cai2024language}                           & $(0, 10]$                           & \multicolumn{1}{c|}{}                     & \multicolumn{1}{c|}{}                 & \multicolumn{1}{c|}{\checkmark}                & \checkmark       \\ \cline{3-8} 
                                                                                              &                                                                                                                     & FPS~\cite{liu2024skepticism}                               & $(10, 100]$                         & \multicolumn{1}{c|}{\checkmark}           & \multicolumn{1}{c|}{\checkmark}       & \multicolumn{1}{c|}{\checkmark}                & \checkmark       \\ \cline{3-8} 
                                                                                              &                                                                                                                     & FUSE~\cite{liu2024tiny}                                    & $(10, 100]$                         & \multicolumn{1}{c|}{\checkmark}           & \multicolumn{1}{c|}{\checkmark}       & \multicolumn{1}{c|}{\checkmark}                & \checkmark       \\ \cline{3-8} 
                                                                                              &                                                                                                                     & Wang et   al.\cite{wang2024decoding}                       & $(10, 100]$                         & \multicolumn{1}{c|}{\checkmark}           & \multicolumn{1}{c|}{\checkmark}       & \multicolumn{1}{c|}{\checkmark}                & \checkmark       \\ \cline{3-8} 
                                                                                              &                                                                                                                     & Concordia~\cite{touzel2024simulation}                      & $(10, 100]$                         & \multicolumn{1}{c|}{\checkmark}           & \multicolumn{1}{c|}{\checkmark}       & \multicolumn{1}{c|}{\checkmark}                & \checkmark       \\ \cline{3-8} 
                                                                                              &                                                                                                                     & Social   Simulacra~\cite{park2022social}                   & $(100, \infty)$                     & \multicolumn{1}{c|}{\checkmark}           & \multicolumn{1}{c|}{\checkmark}       & \multicolumn{1}{c|}{\checkmark}                & \checkmark       \\ \cline{3-8} 
                                                                                              &                                                                                                                     & $S^3$~\cite{gao2023s3}                                     & $(100, \infty)$                     & \multicolumn{1}{c|}{\checkmark}           & \multicolumn{1}{c|}{\checkmark}       & \multicolumn{1}{c|}{\checkmark}                & \checkmark       \\ \cline{3-8} 
                                                                                              &                                                                                                                     & Törnberg et   al.~\cite{tornberg2023simulating}            & $(100, \infty)$                     & \multicolumn{1}{c|}{\checkmark}           & \multicolumn{1}{c|}{\checkmark}       & \multicolumn{1}{c|}{\checkmark}                & \checkmark       \\ \cline{3-8} 
                                                                                              &                                                                                                                     & Y   Social~\cite{rossetti2024social}                       & $(100, \infty)$                     & \multicolumn{1}{c|}{\checkmark}           & \multicolumn{1}{c|}{\checkmark}       & \multicolumn{1}{c|}{\checkmark}                & \checkmark       \\ \cline{3-8} 
                                                                                              &                                                                                                                     & TIS~\cite{zhang2024large}                                  & $(100, \infty)$                     & \multicolumn{1}{c|}{\checkmark}           & \multicolumn{1}{c|}{\checkmark}       & \multicolumn{1}{c|}{\checkmark}                & \checkmark       \\ \cline{3-8} 
                                                                                              &                                                                                                                     & HiSim~\cite{mou2024unveiling}                              & $(100, \infty)$                     & \multicolumn{1}{c|}{\checkmark}           & \multicolumn{1}{c|}{\checkmark}       & \multicolumn{1}{c|}{\checkmark}                & \checkmark       \\ \cline{3-8} 
                                                                                              &                                                                                                                     & OASIS~\cite{yang2024oasis}                                 & $(100, \infty)$                     & \multicolumn{1}{c|}{\checkmark}           & \multicolumn{1}{c|}{\checkmark}       & \multicolumn{1}{c|}{\checkmark}                & \checkmark       \\ \cline{3-8} 
                                                                                              &                                                                                                                     & MindEcho~\cite{xu2024mindechoroleplayinglanguageagents}    & $(100, \infty)$                     & \multicolumn{1}{c|}{\checkmark}           & \multicolumn{1}{c|}{}                 & \multicolumn{1}{c|}{\checkmark}                &                  \\ \cline{3-8} 
                                                                                              &                                                                                                                     & BASES~\cite{ren2024bases}                                  & $(100, \infty)$                     & \multicolumn{1}{c|}{\checkmark}           & \multicolumn{1}{c|}{}                 & \multicolumn{1}{c|}{}                          &                  \\ \cline{2-8} 
                                                                                              & \multirow{5}{*}{\begin{tabular}[c]{@{}c@{}}Recommendation\\ Environments\end{tabular}}                              & InteRecAgent~\cite{huang2023recommender}                   & $(0, 10]$                           & \multicolumn{1}{c|}{\checkmark}           & \multicolumn{1}{c|}{}                 & \multicolumn{1}{c|}{}                          &                  \\ \cline{3-8} 
                                                                                              &                                                                                                                     & Rec4Agentverse~\cite{zhang2024prospect}                    & $(0, 10]$                           & \multicolumn{1}{c|}{\checkmark}           & \multicolumn{1}{c|}{\checkmark}       & \multicolumn{1}{c|}{}                          &                  \\ \cline{3-8} 
                                                                                              &                                                                                                                     & RecAgent~\cite{wang2023recagent}                           & $(10, 100]$                         & \multicolumn{1}{c|}{\checkmark}           & \multicolumn{1}{c|}{\checkmark}       & \multicolumn{1}{c|}{\checkmark}                & \checkmark       \\ \cline{3-8} 
                                                                                              &                                                                                                                     & Agent4Rec~\cite{Zhang2024on}                               & $(100, \infty)$                     & \multicolumn{1}{c|}{\checkmark}           & \multicolumn{1}{c|}{\checkmark}       & \multicolumn{1}{c|}{\checkmark}                & \checkmark       \\ \cline{3-8} 
                                                                                              &                                                                                                                     & AgentCF~\cite{zhang2024agentcf}                            & $(100, \infty)$                     & \multicolumn{1}{c|}{\checkmark}           & \multicolumn{1}{c|}{\checkmark}       & \multicolumn{1}{c|}{\checkmark}                & \checkmark       \\ \bottomrule
\end{tabular}
}
\caption{A list of representative works of society simulation.}
\label{tab:soc}
\end{table*}

While scenarios discuss multi-agent interactions in relatively focused and small-scale contexts and provide solutions within specific domains, society is more complex than a simple scenario. Its complexity lies in many aspects, such as the diversity of its components, the variety of structures, and nonlinear effects~\cite{squazzoni2014social}. Considering this, a series of studies focus on society simulation. In terms of research topic, society simulation generally hopes to investigate societal and macro-level results. In terms of research purpose, society simulation does not aim to solve a task or problem, instead, it focuses on revealing and explaining emergent behaviors and the outcomes of interactions among numerous agents. Society simulations have been a vital tool for theoretical validation and predicting social dynamics.

In this section, we summarize the components of social construction to capture the key features reflected in society simulations in~\S\ref{subsec:soc_element}. Then, we present the different categories of scenarios in society simulation in~\S\ref{subsec:soc_scenario}. After that, we introduce the evaluation of society simulation in~\S\ref{subsec:soc_eval}. The overall framework is illustrated in Figure~\ref{fig:fm_soc} and representative works are summarized in Table~\ref{tab:soc}.

\subsection{Social Construction Elements}
\label{subsec:soc_element}
Considering the complexity of society, a major challenge in society simulation is bridging the gap between individual and societal scales. Some core elements serve as the foundation for modeling social systems. We outline four key dimensions that underpin societal structures and dynamics: \textbf{composition}, \textbf{network},\textbf{ social influence}, and \textbf{outcomes}.

\subsubsection{Composition}
Society is composed of massive and diverse individuals. This diversity, also referred to as heterogeneity~\cite{squazzoni2014social} in social science, encompasses a wide range of beliefs, preferences, behaviors, normative values, and positions within social structures. Modeling this diversity is essential for capturing the varied behavioral patterns and complex social dynamics that emerge from individual differences within a social system.

\paragraph{Individual Composition} 
To model a diverse society, the composition of individuals in society needs to be determined. There are three main approaches to determining the composition of individuals in a system simulating a microcosm of society. Some works rely on \textit{{\ul virtual individual synthesis}}, often not focused on alignment with the real world, aiming to ensure that the system includes users with a variety of attributes, typically by generating virtual individuals with the help of LLMs or humans~\cite{chuang2024wisdom,binz2023turning}. Other works utilize \textit{{\ul existing datasets}}, such as MovieLens-1M~\cite{wang2023recagent,Zhang2024on}, to define user composition within a simulated recommendation platform. Agents are initialized on the basis of the user information within these datasets, reflecting the distribution of users in that context. Recently, an increasing number of studies have focused on \textit{{\ul real-world distribution replication}}, such as the composition of users on social platforms~\cite{yang2024oasis} or the distribution of voters in surveys~\cite{zhang2024electionsim}. For small-scale individual sets, individual data are typically collected manually~\cite{park2023choicemates,park2024generative}. In cases where large-scale populations are required or obtaining real data is difficult, individuals may be sampled based on real-world macro distributions or generated by LLMs to match desired attribute distribution~\cite{Argyle_2023,lee2023can,zhang2024electionsim}.

\paragraph{Trade-off between Simulation Precision and Scale}
When simulating individuals in society simulations, many studies adopt detailed role modeling to enhance the authenticity of agent behavior. Beyond common demographic attributes, this may include factors such as an individual’s past statements and interaction history~\cite{park2023generative,wang2023recagent,Zhang2024on,zhaocompeteai,fontana2024nicer}. However, as the number of individuals increases, such fine-grained modeling becomes expensive. Consequently, a trade-off often arises between the precision of individual modeling and the scale of the simulation. In large-scale simulations, to reduce computational costs, the details of each agent are typically simplified, by retaining only the most essential and common characteristics~\cite{williams2023epidemic,chopra2024limits} or compressing auxiliary dialogue information into shared memory~\cite{yu2024affordable}.

\paragraph{Special Modeling on Outliers} 
As previously mentioned, the composition of individuals in society is diverse. However, not all individuals play an equally significant role. Some individuals, whose attributes or behaviors significantly deviate from the majority, are referred to as outliers~\cite{squazzoni2014social}. Compared to average individuals, outliers often introduce variability and unpredictability to society. Examples include celebrities and opinion leaders~\cite{zhang2024large,xu2024mindechoroleplayinglanguageagents}, who frequently hold prominent positions within social structures and amplify their influence. In situations with limited resources, some studies~\cite{mou2024unveiling} prioritize detailed modeling of these core content producers, while simplifying the modeling for the majority. Meanwhile, intervention policies based on simulation results often focus on these key nodes in networks~\cite{li2024large}, aiming to influence the overall system's behavior by blocking or interfering with them.

\subsubsection{Network}
Social interactions are often conducted through social networks, which can be described using graph structures where nodes represent individuals and edges represent their relations. The network determines the direction of information and influence dissemination. In social science, it has been observed that homophily of individuals can increase the likelihood of communication. Highly similar individuals are more likely to establish connections compared to those with greater differences~\cite{brown1987social,kossinets2009origins}. This principle also informs the construction of networks in society simulations. The methods for constructing social networks vary across different scenarios. Here, we divide them into offline networks and online networks.

\paragraph{Offline Network}
An offline network represents connections formed through in-person interactions, such as face-to-face communication or the spread of opinions and diseases in physical settings. On the one hand, some studies aim to simulate interactions in virtual worlds, thus determining the connections between agents in a \textit{{\ul random or predefined}} manner~\cite{park2023generative,yu2024affordable,ren2024emergence}. On the other hand, when some studies aim to simulate the spread of a disease or event information in the real world, considering the difficulty of obtaining real data, they often estimate the social relations using \textit{{\ul external algorithms}} or agents themselves~\cite{xiao2023simulating,williams2023epidemic}. However, in studies with a large scale of agents, the network relationships between individuals are sometimes ignored, and individuals are treated as independent~\cite{zhang2024electionsim}. Alternatively, some studies provide rough information, such as community statistics, in place of specific details about the agents' neighbors~\cite{chopra2024limits}.

\paragraph{Online Network}
An online network is a digital structure where individuals or entities interact through platforms, such as online social platforms and recommendation platforms, forming connections based on activities, relationships, or shared interests. At the beginning, some studies \textit{{\ul randomly}} initialize the social relations for users existing datasets~\cite{wang2023recagent} or synthesized users~\cite{liu2024skepticism}, while other efforts have focused on crawling \textit{{\ul authentic}} social relationships from social media platforms like Weibo~\cite{gao2023s3} and Twitter~\cite{mou2024unveiling}. However, as the scale of individuals increase, it may be challenging to obtain all of their authentic relationships. Therefore, some studies construct networks using a small portion of real relationship data combined with a large amount of \textit{{\ul synthetic}} relationship data~\cite{yang2024oasis}, or connect similar users based on the assumption of homophily~\cite{tang2024synthesizing}.

\subsubsection{Social Influence}
Social influence refers to the influence agents have on others and the influence they receive from others during interactions. This is also known as embeddedness in social sciences~\cite{squazzoni2014social}, which suggests that individuals behavior and decisions are influenced by their environment. When conducting society simulations, it is necessary to consider the modeling of such social influence.

\paragraph{Influence Received by the Influencee}
The same information may produce different effects when received by individuals with different traits. Currently, most studies have modeled how the influence received by the recipient varies based on their profile~\cite{gao2023s3,liu2024skepticism,yang2024oasis}. This can be easily achieved by integrating the individual's profile, memory and the information received from others into the same context. Building this, a few works further induce additional mechanisms such as cognitive bias~\cite{chuang2023simulating} and reflection on norms~\cite{ren2024emergence} to enhance agents' understanding and perception of the received messages.

\paragraph{Influence Exerted by the Influencer}
The same message conveyed by different individuals can result in varying social impacts. The Pareto distribution and the Matthew Effect~\cite{wang2023recagent,mou2024unveiling} indicate that information, influence, or attention tends to concentrate on a small group of individuals who are already dominant in the community. Therefore, when simulating social interactions, the identity, status, and reputation of the information sender are also crucial. Some studies start with real-world data to conduct detailed modeling of opinion leaders~\cite{xu2024mindechoroleplayinglanguageagents,zhang2024large}. Other studies, instead of focusing on the role of the influencer, model the influence exerted by the influencer by incorporating the relation information such as social impression memory~\cite{yu2024affordable} and share party affiliation~\cite{chuang2024wisdom}. In addition to the influence exerted by individuals, research has found that as group size increases, the impact of a single influencer may diminish. However, the influence of the group on individuals often drives them to align their behavior with the group, leading to the emergence of the herd effect~\cite{yang2024oasis}.

\subsubsection{Outcomes}
Social emergence suggests that the collective behaviors or phenomena arise from individual interactions are not a linear sum of individual actions but rather complex patterns emerge from the interactions~\cite{schelling1971dynamic,squazzoni2014social}. 
These interaction outcomes may be measurable macro results, such as voting results and public opinion levels, or they may also be qualitative social phenomena and norms. Next, we will discuss these two types of outcomes separately.

\paragraph{Macro Statistical Results}
Macro statistical results are typically the focus of existing studies, as they are closely related to predefined research objectives such as market research, election predictions, and public opinion forecasting. These studies often aim to calculate the sum or average of the choices or opinions of all agents in the system. To get a static opinion distribution, some studies overlook the social interactions and instead directly sum up individual choices to obtain macro outcomes~\cite{sun2024randomsiliconsamplingsimulating,zhang2024electionsim}, simplifying the complexity of social dynamics. Another line of research focuses on the change of indicators by modeling multiple rounds of interactions among the agents over a period of time and then statistically analyzing the results~\cite{gao2023s3,tornberg2023simulating,han2023guinea,li2024econagent,wu2024shall}.

\paragraph{Formation of Social Phenomena and Social Norms}
In addition to the quantifiable macro results, some social phenomena and social norms are also important outcomes of social interactions. On the one hand, some studies have identified the bubble effect in recommendation systems~\cite{Zhang2024on}, echo chambers in social media~\cite{mou2024unveiling,yang2024oasis,wang2024decoding},  Matthew effect in competitive agent interactions~\cite{zhaocompeteai}, and spontaneous cooperation of competing agents~\cite{wu2024shall} by calculating additional metrics or observing the trends of primary indicators. On the other hand, some studies examine social norms as an important byproduct of social interactions. This includes simulating and testing whether community rules can shape desired social norms~\cite{park2022social}, constructing normative architecture to observe the emergence of social norms~\cite{ren2024emergence}, studying how social media language evolves in the presence of regulatory constraints~\cite{cai2024language}, and observing changes in social norms in real-world scenarios such as autonomous driving~\cite{wang2024can}.

\subsection{Scenario}
\label{subsec:soc_scenario}
Society simulation has been widely applied to various scenarios related to human society. These scenarios cover different aspects of daily human life, and existing studies can be categorized into three primary areas: \textbf{general economics}, \textbf{sociology and political science}, as well as \textbf{online platforms}. 

\subsubsection{General Economics}
Simulations in general economics analyze decision-making and behaviors related to resource allocation and competition. These studies primarily investigate how agents make decisions influenced by economic incentives, market rules and resource constraints, while also examining how interactions among groups shape broader economic trends.

\paragraph{Game Theory and Strategic Interactions}
Some research mainly focuses on game theory and strategic interaction. These scenarios typically involve small groups of agents, with a primary focus on the complex interactions between agents. Some works use classic game theory games, such as the Prisoner's Dilemma, to explore agent behavior in game-theoretic scenarios, including trust behavior~\cite{xie2024can}, logic reasoning and decision-making~\cite{mensfelt2024logic}, rationality and strategic reasoning ability~\cite{guo2024economics}, cooperation tendencies~\cite{fontana2024nicer} and how emotional states can disrupt rational decision-making~\cite{mozikov2024good}. Other studies focus on real-world scenarios other than the games, such as spontaneous cooperation in competitive environments~\cite{wu2024shall}, complex market behaviors in firm competition~\cite{han2023guinea}, and competition between restaurant and customer agents~\cite{zhaocompeteai}. Overall, the former kind of scenarios simplifies the environment, making it easier to conduct controlled research on agent behavior, while the latter provides more insights for real-world applications.

\paragraph{Economic Contexts}
In addition to close studies on game theory and strategic interactions, some studies focus on the use of agents and their interactions within economic environments. Horton~\cite{horton2023large} examines economic agents driven by LLMs in various experiments to replicate human behavior in economic scenarios. EconAgent~\cite{li2024econagent} introduces agents for macroeconomic simulation, emphasizing the influence of macroeconomic trends. SRAP-Agent~\cite{ji2024srap} proposes a framework for simulating and optimizing scarce resource allocation in economics, specifically in public housing allocation scenarios. Besides, some studies involve broader macroeconomic domains, using agents to simulate and predict the spread of diseases and the change in unemployment rates~\cite{williams2023epidemic,chopra2024limits}.

\subsubsection{Sociology and Political Science}
Society simulation has been widely used in sociological and political science research.  These studies range from small-scale laboratory experiments that validate theories and hypotheses to large-scale social surveys aimed at understanding public choices. The goal is to leverage agents as substitutes for humans in studying human behavior within sociological and political contexts.

\paragraph{Public Opinion Survey} A mainstream application of society simulation is public opinion survey, which aims to predict the perspectives of specific groups toward a given subject through simulation and aggregate their opinions to support advanced needs such as election forecasting and public administration. Argyle et al.~\cite{Argyle_2023} first propose that LLMs could serve as silicon samples of humans, through several large-scale surveys conducted in the United States. Building on this, some studies have expanded their focus to scenarios of opinion surveys~\cite{lee2023can,chaudhary2024large,chuang2024beyond}, such as election polls~\cite{zhang2024electionsim} and response to public administration crisis~\cite{xiao2023simulating}, delving deeper into issues like population complexity and algorithmic bias. Recently, agents have demonstrated the potential to replicate participants' responses in individual interviews~\cite{park2024generative}. These studies lay the foundation for new tools to investigate individual and collective behavior.

\paragraph{Individual and Organizational Behavior Observation}
Other studies focus on observing individual or organizational behavior in common or specific settings. Some works do not specify a particular scenario but instead observe agents' social interactions and potential phenomena in daily life within a sandbox environment~\cite{park2023generative,kaiya2023lyfe,ren2024emergence,yu2024mineland}. 
Other studies aim to validate theories or hypotheses in specific scenarios, such as the wisdom of partisan crowds~\cite{chuang2024wisdom}, information management~\cite{park2023choicemates}, organizational behavior management~\cite{zhu2024generative}, and the evolution of personality traits~\cite{suzuki2024evolutionary}.

\subsubsection{Online Platform}
Online Platforms are a vital component of society simulation, offering a practical means to study complex social phenomena in digital environments. These platforms, ranging from social media to online communities, allow agents to simulate real-world interactions and study dynamics such as opinion formation, information spread, and collective behaviors. 
\paragraph{Social Platforms}
Online social platforms have long served as an important testing ground for studying the propagation of information and the evolution of opinions. These studies typically recreate environments similar to popular social platforms, such as Twitter, Reddit, and Weibo, with action spaces that include behaviors like sharing, commenting, and liking. By simulating these scenarios, researchers can model the spread of information and track changes in user attitudes following events, covering a wide range of topics such as general news, rumors, and the role of opinion leaders \cite{liu2024skepticism,gao2023s3,cai2024language,rossetti2024social,zhang2024large,liu2024tiny}. In such scenarios, the roles and relationships of agents play a critical role in ensuring realistic simulations. Initially, many studies relied on real-world data scraped from platforms to maintain consistency \cite{mou2024unveiling,gao2023s3}. However, as the scale of these simulations grew and data acquisition became more challenging, researchers began exploring the use of synthetic data \cite{yang2024oasis}. Furthermore, to accommodate the increasing demand for simulating larger numbers of agents, some studies have developed large-scale society simulation platforms \cite{gao2024agentscope,pan2024very}, employing parallel processing and other strategies to enhance simulation efficiency.

\paragraph{Recommendation Environments}
Another widely studied scenario is the recommendation environment, where these works use agents to simulate user responses in order to validate and improve recommendation algorithms~\cite{huang2023recommender,zhang2024prospect}. A key feature across these studies is the use of agents to emulate personalized behaviors such as item selection, preferences, and emotional responses, often integrating user memory and contextual factors~\cite{wang2023recagent,zhang2024agentcf,Zhang2024on}. Additionally, some approaches incorporate external knowledge or self-reflection mechanisms, allowing agents to adapt and learn from their interactions over time~\cite{wang2023recmind}. These studies collectively show how LLMs can bridge the gap between traditional recommender systems and more interactive, human-like behavior simulations, offering new ways to improve recommendation accuracy and better understand user dynamics.

\subsection{Evaluation}
\label{subsec:soc_eval}

For society simulations, the evaluation primarily focuses on the comparison between the simulation results and real-world data, with assessments centered on micro level, macro level and system level.

\paragraph{Micro-level Evaluation}
Individual simulation accuracy is key to society simulation. Therefore, micro-level evaluation of society simulation has received widespread attention. Initially, evaluations in non-real-world simulations draw on the Turing test, assessing agent behavior's resemblance to human behavior, often \textit{\underline{subjectively}} by humans or LLMs~\cite{park2023generative,liang2023leveraging,yu2024affordable}. For specific scenarios, metrics like partisan bias and human likeness index are proposed~\cite{chuang2024wisdom}. When simulations target real-world scenarios with available empirical data, automated metrics like emotion, attitude, behavior consistency, and user taste alignment can be designed for more \textit{\underline{objective}} evaluations by comparing simulation content with real-world data~\cite{gao2023s3,mou2024unveiling,Zhang2024on}.

\paragraph{Macro-level Evaluation}
Social interactions often lead to collective outcomes, so it is important to evaluate whether macro-level outcomes show patterns and trends that are consistent with the real world. For sociology and online platforms, attention is typically given to whether the scale of propagation, the distribution and trends of collective opinions and traits align with those of the real world. In addition to qualitative methods such as \textit{\underline{subjective}} evaluation~\cite{gao2023s3,Zhang2024on}, some studies have proposed quantitative metrics, such as fitted parameters, correlation coefficients and change of toxicity of community content to measure this differences \textit{\underline{objectively}}~\cite {tornberg2023simulating,liu2024skepticism,mou2024unveiling,yang2024oasis}. Similarly, in economic simulation, the evaluation of simulated economic systems depends on whether they can reproduce the most representative macroeconomic laws~\cite{li2024econagent}.

\paragraph{System-level Evaluation}
System-level evaluation is concerned with assessing the overall performance of a simulation system, irrespective of the specific content being simulated. With the growing number of agents in simulation, the focus of contemporary research has been on system efficiency and associated costs. Efficiency is assessed through various metrics, such as the time it takes to run a simulation, the resources that are utilized during the process, and how well the simulation can scale with an increasing number of agents~\cite{wang2023recagent,yang2024oasis,pan2024very}. These metrics are crucial for understanding how well the system can handle complexity and the demands of larger simulations. On the cost side, evaluations often center on the number of tokens consumed during the simulation or the financial expenditure incurred~\cite{yu2024affordable}.

\begin{table*}[!htp]
\resizebox{\textwidth}{!}{ 
\setlength{\tabcolsep}{2pt} 
\centering

\resizebox{\textwidth}{!}{\begin{tabular}{c|c|c|c|c|c|c|c}
\toprule
\textbf{Domain} & \textbf{Dataset} & \textbf{Type} & \textbf{Source} & \textbf{\# individual num} & \textbf{\# dialogue num} & \textbf{Paper} & \textbf{Link}\\
\midrule
\multirow{23}{*}{Characters} & Final Dialogue Dataset & Dialogue & Wikipedia & /  & 22,311 & \cite{dinan2019wizardwikipediaknowledgepoweredconversational} & \href{https://parl.ai/projects/wizard_of_wikipedia/}{\textcolor{blue}{Link}} \\
\cline{2-8}
   & P-weibo Dataset & Dialogue/Description & Weibo & /  & 2,000,000 & \cite{li2021dialoguehistorymatterspersonalized} & /  \\
\cline{2-8}
   & P-Ubuntu dialogue corpus & Dialogue/Description & Corpus & /   & 2,000,000 & \cite{li2021dialoguehistorymatterspersonalized} & /  \\
\cline{2-8}
   & LISCU Dataset & Description & Books, Summaries & 9,499 & /  & \cite{brahman2021letcharacterstellstory} & \href{https://github.com/huangmeng123/lit_char_data_wayback}{\textcolor{blue}{Link}} \\
\cline{2-8}
   & FoCus Dataset & Description & Wikipedia & /  & 86,712 & \cite{jang2022customizedconversationcustomizedconversation} & \href{https://drive.google.com/file/d/1YmEW12HqjAjlEfZ05g8VLRux8kyUjdcI/view}{\textcolor{blue}{Link}} \\
\cline{2-8}
   & ConvAI2 benchmark dataset & Description & Human & /  & 18,878 & \cite{cho2022personalizeddialoguegeneratorimplicit} & /  \\
\cline{2-8}
   & HPD Benchmark & Dialogue/Description & Books & 1 & about 2,500 & \cite{chen2023largelanguagemodelsmeet} & \href{https://nuochenpku.github.io/HPD.github.io/}{\textcolor{blue}{Link}} \\
\cline{2-8}
   & LaMP Benchmark & Description & /  & /  & /  & \cite{salemi2024lamplargelanguagemodels} & \href{http://lamp-benchmark.github.io/}{\textcolor{blue}{Link}} \\
\cline{2-8}
   & Multimodal Persona Chat & Image/Dialogue & Reddit & /  & 15,000 & \cite{ahn-etal-2023-mpchat} & \href{https://github.com/ahnjaewoo/mpchat}{\textcolor{blue}{Link}} \\
\cline{2-8}
   & LiveChat & Description/Dialogue & Douyin & 351 & 1,330,000 & \cite{gao2023livechatlargescalepersonalizeddialogue} & \href{https://github.com/gaojingsheng/LiveChat}{\textcolor{blue}{Link}} \\
\cline{2-8}
   & COMSET & Dialogue & Strips & 13 & 53,903 & \cite{agrawal-etal-2023-multimodal} & \href{https://github.com/dair-iitd/MPdialog}{\textcolor{blue}{Link}} \\
\cline{2-8}
   & ChatHaruhi Dataset & Dialogue & Movies, Script & 32 & 54,000 & \cite{li2023chatharuhirevivinganimecharacter} & \href{https://github.com/LC1332/Chat-Haruhi-Suzumiya}{\textcolor{blue}{Link}} \\
\cline{2-8}
   & RoleBench & Dialogue & Scripts & 100 & 168,093 & \cite{wang2024rolellmbenchmarkingelicitingenhancing} & \href{https://github.com/InteractiveNLP-Team/RoleLLM-public}{\textcolor{blue}{Link}} \\
\cline{2-8}
   & Character-LLM Dataset & Description & /  & 9 & 14,400 & \cite{shao2023characterllmtrainableagentroleplaying} & \href{https://github.com/choosewhatulike/trainable-agents}{\textcolor{blue}{Link}} \\
\cline{2-8}
   & PersonaChat Dataset & Description & /  & /  & /  & \cite{zhang2018personalizingdialogueagentsi} & \href{https://github.com/datasets-mila/datasets--personachat}{\textcolor{blue}{Link}} \\
\cline{2-8}
   & CharacterDial & Description/Dialogue & Literary Resources,LLM ,Human  & 250 & 1,034 & \cite{zhou2023characterglmcustomizingchineseconversational} & \href{https://github.com/thu-coai/CharacterGLM-6B}{\textcolor{blue}{Link}} \\
\cline{2-8}
   & Synthetic Persona Chat & Description/Dialogue & LLM & 10,371 & 21,907 & \cite{jandaghi2023faithfulpersonabasedconversationaldataset} & \href{https://github.com/google-research-datasets/Synthetic-Persona-Chat}{\textcolor{blue}{Link}} \\
\cline{2-8}
   & RoleEval Dataset & Description & Wikipedia, Baidu, Fandom, Moegirlpedia & 300 & 6,000 & \cite{shen2024roleevalbilingualroleevaluation} & \href{https://github.com/Magnetic2014/RoleEval}{\textcolor{blue}{Link}} \\
\cline{2-8}
   & CharacterEval Dataset & Description/Dialogue & Novels,Scripts & 77 & 1,785 & \cite{tu2024characterevalchinesebenchmarkroleplaying} & \href{https://github.com/morecry/CharacterEval}{\textcolor{blue}{Link}} \\
\cline{2-8}
   & Life Choice Dataset & Description & Books & 1,401 & /  & \cite{xu2024characterdestinylargelanguage} & /  \\
\cline{2-8}
   & Cross Dataset & Description & Books & /  & /  & \cite{yuan2024evaluatingcharacterunderstandinglarge} & \href{https://github.com/Joanna0123/character_profiling}{\textcolor{blue}{Link}} \\
\cline{2-8}
   & MMRole-Data & Description/Dialogue/Image & Wikipedia,Baidu & 85 & 14000 & \cite{dai2024mmrolecomprehensiveframeworkdeveloping} & \href{https://huggingface.co/datasets/YanqiDai/MMRole_dataset}{\textcolor{blue}{Link}} \\
\cline{2-8}
   & RP Dataset & Dialogue & Novels,Scripts & 331 & 3552 & \cite{yu2024dialogueprofiledialoguealignmentframework} & \href{https://github.com/yuyouyu32/BeyondDialogue}{\textcolor{blue}{Link}} \\
\cline{2-8}
    & MPI dataset & Description & /  & /   & /  & \cite{jiang2023evaluatinginducingpersonalitypretrained} & \href{https://github.com/jianggy/MPI/tree/main/inventories}{\textcolor{blue}{Link}} \\
\hline
\multirow{7}{*}{Demographics} & Who is GPT3 Dataset & /  & /  & /  & /  & \cite{miotto2022gpt3explorationpersonalityvalues} & \href{https://github.com/ben-aaron188/}{\textcolor{blue}{Link}} \\
\cline{2-8}
& Dataset Movielens 1M & /  & /  & /  & /  & \cite{wang2024userbehaviorsimulationlarge} & \href{https://grouplens.org/datasets/movielens/1m/}{\textcolor{blue}{Link}} \\
\cline{2-8}
    & EmotionBench & /  & /  & /  & /  & \cite{huang2024emotionallynumbempatheticevaluating} & \href{https://github.com/CUHK-ARISE/EmotionBench}{\textcolor{blue}{Link}} \\
\cline{2-8}
    & OpinionQA Dataset & /  & Surveys & /  & /  & \cite{li2024steerabilitylargelanguagemodels} & \href{https://github.com/tatsu-lab/opinions_qa}{\textcolor{blue}{Link}} \\
\cline{2-8}
    & CultureLLM Dataset & Dialogue & Survey & /  & /  & \cite{li2024culturellmincorporatingculturaldifferences} & \href{https://github.com/Scarelette/CultureLLM}{\textcolor{blue}{Link}} \\
\cline{2-8}
    & PersonaHub Dataset & Description & LLM & 200,000 & 375,000 & \cite{ge2024scalingsyntheticdatacreation} & \href{https://github.com/tencent-ailab/persona-hub}{\textcolor{blue}{Link}} \\

\bottomrule
\end{tabular}}
}
\vspace{-3mm}
\caption{Summary of commonly used datasets for individual simulation.}
\label{tab:indi_data}
\end{table*}

\section{Datasets and Benchmarks}~\label{sec:data}

\subsection{Individual Simulation}
We summarize commonly used datasets for scenario simulation in Table~\ref{tab:indi_data}. Datasets for individual simulation can be classified into two types: description datasets and dialogue datasets. Description datasets include individual-specific information, such as life experiences, relationships, and basic demographic details like career, age, and gender, often sourced from literature summaries or search engines like Baidu and Wikipedia. Dialogue datasets consist of single-turn or multi-turn conversations in specific scenarios, created by extracting relevant plots for targeted characters or gathering utterances from social media. Some datasets are designed specifically for evaluation, combining basic personal information with customized questions or tasks to assess simulation performance.

\subsection{Scenario Simulation}
We summarize commonly used datasets for scenario simulation in Table~\ref{tab:task_data}, comprising dialog-driven and task-driven scenarios. The datasets cover a wide range of formats, including QA, multiple-choice, rating, code, and game. We observed that QA and multiple-choice formats dominate the data types, while domain-specific datasets like judicial, game, and media prefer to preserve domain-tailored data type.
Based on task complexity, datasets are categorized into three levels: easy, medium, and hard.
Additionally, according to the collection methods, datasets are classified as human-annotated, real-world, or synthetic.
\begin{table*}[!htp]
\setlength{\tabcolsep}{2pt} 
\centering
\resizebox*{!}{\dimexpr\textheight-2\baselineskip\relax}{  
\begin{tabular}{c|c|c|c|c|c|c|c}
\toprule
\textbf{Domain}                                                                   & \textbf{Datasets}      & \textbf{Type}           & \textbf{Complexity} & \textbf{\# case} & \textbf{Collection} & \textbf{Used by}                                                                                                                                                                & \textbf{Data Link}                                                                                                                                                \\ \midrule
\multirow{33}{*}{\begin{tabular}[c]{@{}c@{}}Dialog- \\ Driven\end{tabular}}     & MiniWob++              & Web Interaction         & Hard                & /                & human               & \cite{wu2023autogen}                                                                                                                                                            & \href{https://miniwob.farama.org/}{\textcolor{blue}{Link}}                                                                                                        \\ \cline{2-8}
                                                                                  & SOTOPIA                & Open-Ended Environment  & Hard                & /                & human               & \cite{zhou2023sotopia}                                                                                                                                                          & \href{https://huggingface.co/datasets/cmu-lti/sotopia}{\textcolor{blue}{Link}}                                                                                    \\ \cline{2-8}
                                                                                  & WebQuestions           & QA                      & Easy                & 5,810            & human               & \cite{ma2024debate}                                                                                                                                                             & \href{https://worksheets.codalab.org/worksheets/0xba659fe363cb46e7a505c5b6a774dc8a}{\textcolor{blue}{Link}}                                                       \\ \cline{2-8}
                                                                                  & WebQSP                 & QA                      & Easy                & 4,737            & human               & \cite{ma2024debate}                                                                                                                                                             & \href{https://www.microsoft.com/en-us/research/publication/the-value-of-semantic-parse-labeling-for-knowledge-base-question-answering-2/}{\textcolor{blue}{Link}} \\ \cline{2-8}
                                                                                  & CWQ                    & QA                      & Easy                & 34,689           & human               & \cite{ma2024debate}                                                                                                                                                             & \href{https://github.com/alontalmor/WebAsKB?tab=readme-ov-file}{\textcolor{blue}{Link}}                                                                           \\ \cline{2-8}
                                                                                  & GrailQA                & QA                      & Easy                & 64,331           & human               & \cite{ma2024debate}                                                                                                                                                             & \href{https://dki-lab.github.io/GrailQA/}{\textcolor{blue}{Link}}                                                                                                 \\ \cline{2-8}
                                                                                  & Natural Questions      & QA                      & Easy                & 323,045          & human               & \cite{wu2023autogen}                                                                                                                                                            & \href{https://ai.google.com/research/NaturalQuestions}{\textcolor{blue}{Link}}                                                                                    \\ \cline{2-8}
                                                                                  & FairEval               & QA                      & Medium              & 80               & human               & \cite{chan2023chateval}                                                                                                                                                         & \href{https://github.com/i-Eval/FairEval}{\textcolor{blue}{Link}}                                                                                                 \\ \cline{2-8}
                                                                                  & MMLU                   & Multiple-Choice         & Hard                & 115,700          & real world          & \cite{du2023improving,tang2023medagents,tan2024true,yukhymenko2024synthetic,zhang2023exploring} & \href{https://github.com/hendrycks/test}{\textcolor{blue}{Link}}                                                                                                  \\ \cline{2-8}
                                                                                  & BIG-bench              & /                       & Hard                & /                & human               & \cite{du2023improving,chen2023agentverse,zhang2023exploring}                                                        & \href{https://github.com/google/BIG-bench}{\textcolor{blue}{Link}}                                                                                                \\ \cline{2-8}
                                                                                  & MetaQA                 & QA                      & Medium              & 407,513          & real world, human   & \cite{ma2024debate}                                                                                                                                                             & \href{https://github.com/yuyuz/MetaQA}{\textcolor{blue}{Link}}                                                                                                    \\ \cline{2-8}
                                                                                  & AmazonHistoryPrice     & Product Info            & Hard                & 930              & real world          & \cite{xia2024measuring}                                                                                                                                                         & \href{https://github.com/TianXiaSJTU/AmazonPriceHistory}{\textcolor{blue}{Link}}                                                                                  \\ \cline{2-8}
                                                                                  & MATH                   & Math Problem            & Medium              & 12,500           & real world          & \cite{wu2023autogen}                                                                                                                                                            & \href{https://github.com/hendrycks/math/}{\textcolor{blue}{Link}}                                                                                                 \\ \cline{2-8}
                                                                                  & Arithmetic             & Math Expression         & Easy                & /                & human               & \cite{du2023improving}                                                                                                                                                          & \href{https://github.com/composable-models/llm_multiagent_debate}{\textcolor{blue}{Link}}                                                                         \\ \cline{2-8}
                                                                                  & Counter-Intuitive AR   & Reasoning Problem       & Easy                & 200              & human               & \cite{liang2023encouraging}                                                                                                                                                     & \href{https://github.com/Skytliang/Multi-Agents-Debate}{\textcolor{blue}{Link}}                                                                                   \\ \cline{2-8}
                                                                                  & CommonMT               & Translation Triple      & Medium              & 1,200            & human               & \cite{liang2023encouraging}                                                                                                                                                     & \href{https://github.com/tjunlp-lab/CommonMT}{\textcolor{blue}{Link}}                                                                                             \\ \cline{2-8}
                                                                                & Overcooked-AI          & Game                    & Medium              & /                & human               & \cite{zhang2024towards}                                                                                                                                                         & \href{https://github.com/HumanCompatibleAI/overcooked_ai}{\textcolor{blue}{Link}}                                                                                 \\ \cline{2-8}
                                                                                  & AVALONBENCH            & Game                    & Easy                & /                & human               & \cite{light2023text}                                                                                                                                                            & \href{https://github.com/jonathanmli/Avalon-LLM}{\textcolor{blue}{Link}}                                                                                          \\ \cline{2-8}
                                                                                  & Jubensha               & Game                    & Medium              & 1,115            & real world          & \cite{wu2023deciphering}                                                                                                                                                        & \href{https://github.com/jackwu502/ThinkThrice}{\textcolor{blue}{Link}}                                                                                           \\ \cline{2-8}
                                                                                  & FanLang-9              & Game                    & Easy                & 18,800           & real world          & \cite{wu2024enhance}                                                                                                                                                            & \href{https://github.com/boluoweifenda/werewolf}{\textcolor{blue}{Link}}                                                                                          \\ \cline{2-8}
                                                                                  & WellPlay               & QA                      & Hard                & 1,482            & human               & \cite{zhu2024player}                                                                                                                                                            & \href{https://github.com/alickzhu/PLAYER}{\textcolor{blue}{Link}}                                                                                                 \\ \cline{2-8}
                                                                                  & WWQA                   & QA                      & Medium              & 2,053            & synthetic           & \cite{du2024helmsman}                                                                                                                                                           & \href{https://github.com/doslim/Evaluate-the-Opinion-Leadership-of-LLMs}{\textcolor{blue}{Link}}                                                                  \\ \cline{2-8}
                                                                                  & Biographies            & Biographies             & Easy                & 524              & real world          & \cite{du2023improving}                                                                                                                                                          & \href{https://github.com/composable-models/llm_multiagent_debate}{\textcolor{blue}{Link}}                                                                         \\ \cline{2-8}
                                                                                  & ALFWorld               & Embodied Environment    & Medium              & 3,827            & human               & \cite{wu2023autogen}                                                                                                                                                            & \href{https://github.com/alfworld/alfworld/tree/master/alfworld/data}{\textcolor{blue}{Link}}                                                                     \\ \cline{2-8}
                                                                                  & ED dataset             & Conversational          & Hard                & 24,850           & human               & \cite{zhang2024self}                                                                                                                                                            & \href{https://github.com/facebookresearch/EmpatheticDialogues}{\textcolor{blue}{Link}}                                                                            \\ \cline{2-8}
                                                                                  & Topical-Chat           & Conversational          & Medium              & 10,784           & human               & \cite{chan2023chateval}                                                                                                                                                         & \href{https://github.com/alexa/Topical-Chat}{\textcolor{blue}{Link}}                                                                                              \\ \cline{2-8}
                                                                                  & COPA                   & Multiple-Choice         & Easy                & 500              & real world          & \cite{xiong2023examining}                                                                                                                                                       & \href{http://www.ict.usc.edu/~gordon/copa.html}{\textcolor{blue}{Link}}                                                                                           \\ \cline{2-8}
                                                                                  & $\alpha$NLI            & Multiple-Choice         & Easy                & 1,507            & human               & \cite{xiong2023examining}                                                                                                                                                       & \href{http://abductivecommonsense.xyz}{\textcolor{blue}{Link}}                                                                                                    \\ \cline{2-8}
                                                                                  & CSQA                   & Multiple-Choice         & Easy                & 1,221            & human               & \cite{xiong2023examining}                                                                                                                                                       & \href{https://www.tau-nlp.org/commonsenseqa}{\textcolor{blue}{Link}}                                                                                              \\ \cline{2-8}
                                                                                  & Social IQa             & Multiple-Choice         & Easy                & 1,935            & human               & \cite{xiong2023examining}                                                                                                                                                       & \href{https://huggingface.co/datasets/allenai/social_i_qa}{\textcolor{blue}{Link}}                                                                                \\ \cline{2-8}
                                                                                  & PIQA                   & Multiple-Choice         & Easy                & 1,838            & human               & \cite{xiong2023examining,tan2024true}                                                                                         & \href{https://huggingface.co/datasets/ybisk/piqa}{\textcolor{blue}{Link}}                                                                                         \\ \cline{2-8}
                                                                                  & StrategyQA             & Multiple-Choice         & Easy                & 2,290            & human               & \cite{xiong2023examining}                                                                                                                                                       & \href{https://github.com/eladsegal/strategyqa}{\textcolor{blue}{Link}}                                                                                            \\ \cline{2-8}
                                                                                  & e-CARE                 & Multiple-Choice         & Easy                & 2,122            & human               & \cite{xiong2023examining}                                                                                                                                                       & \href{https://github.com/waste-wood/e-care}{\textcolor{blue}{Link}}                                                                                               \\ \hline

\multirow{41}{*}{\begin{tabular}[c]{@{}c@{}}Task- \\ Driven\end{tabular}} & WiKiTQ                 & QA                      & Easy                & 22,033           & real world          & \cite{zhuautotqa}                                                                                                                                                               & \href{https://ppasupat.github.io/WikiTableQuestions/}{\textcolor{blue}{Link}}                                                                                     \\ \cline{2-8}
                                                                                  & TabFact                & QA                      & Hard                & 118,275          & real world, human   & \cite{zhuautotqa}                                                                                                                                                               & \href{https://tabfact.github.io/}{\textcolor{blue}{Link}}                                                                                                         \\ \cline{2-8}
                                                                                  & FeTaQA                 & QA                      & Hard                & 10,330           & real world, human   & \cite{zhuautotqa}                                                                                                                                                               & \href{https://github.com/Yale-LILY/FeTaQA}{\textcolor{blue}{Link}}                                                                                                \\ \cline{2-8}
                                                                                  & HumanEval              & Code                    & Easy                & 164              & real world          & \cite{dong2023self,hong2023metagpt,chen2023agentverse,yukhymenko2024synthetic}                            & \href{https://github.com/openai/human-eval}{\textcolor{blue}{Link}}                                                                                               \\ \cline{2-8}
                                                                                  & MBPP                   & Code                    & Easy                & 974              & real world          & \cite{dong2023self,hong2023metagpt}                                                                                           & \href{https://github.com/google-research/google-research/tree/master/mbpp}{\textcolor{blue}{Link}}                                                                \\ \cline{2-8}
                                                                                  & APPS                   & Code                    & Easy                & /                & real world          & \cite{dong2023self}                                                                                                                                                             & \href{https://github.com/hendrycks/apps}{\textcolor{blue}{Link}}                                                                                                  \\ \cline{2-8}
                                                                                  & Code                   & Conversational          & Hard                & 50,000           & synthetic           & \cite{li2023camel}                                                                                                                                                              & \href{https://huggingface.co/camel-ai}{\textcolor{blue}{Link}}                                                                                                    \\ \cline{2-8}
                                                                                  & CoderEval              & Code                    & Medium              & 230              & real world          & \cite{dong2023self}                                                                                                                                                             & \href{https://github.com/CoderEval/CoderEval}{\textcolor{blue}{Link}}                                                                                             \\ \cline{2-8}
                                                                                  & SRDD                   & Software Requirement    & Medium              & 1,200            & synthetic           & \cite{qian2024chatdev,qian2023experiential,qian2024iterative,yukhymenko2024synthetic}                     & \href{https://github.com/OpenBMB/ChatDev/tree/main/SRDD}{\textcolor{blue}{Link}}                                                                                  \\ \cline{2-8}
                                                                                  & SoftwareDev            & Task Prompt             & Hard                & 70               & human               & \cite{hong2023metagpt}                                                                                                                                                          & \href{https://github.com/geekan/MetaGPT}{\textcolor{blue}{Link}}                                                                                                  \\ \cline{2-8}
                                                                                  & SWE-bench              & Code                    & Easy                & 2,294            & real world          & \cite{zhang2024autocoderover}                                                                                                                                                   & \href{https://github.com/princeton-nlp/SWE-bench}{\textcolor{blue}{Link}}                                                                                         \\ \cline{2-8}
                                                                                  & AI Society             & Conversational          & Easy                & 25,000           & synthetic           & \cite{li2023camel}                                                                                                                                                              & \href{https://huggingface.co/camel-ai}{\textcolor{blue}{Link}}                                                                                                    \\ \cline{2-8}
                                                                                  & SynthPAI               & Comment                 & Hard                & 7,823            & synthetic           & \cite{qian2024scaling}                                                                                                                                                          & \href{https://huggingface.co/datasets/RobinSta/SynthPAI}{\textcolor{blue}{Link}}                                                                                  \\ \cline{2-8}
                                                                                  & ScienceWorld           & Interactive Environment & Hard                & /                & human               & \cite{lin2024swiftsage}                                                                                                                                                         & \href{https://github.com/allenai/ScienceWorld}{\textcolor{blue}{Link}}                                                                                            \\ \cline{2-8}
                                                                                  & Science                & QA                      & Medium              & 60,000           & synthetic           & \cite{li2023camel}                                                                                                                                                              & \href{https://huggingface.co/camel-ai}{\textcolor{blue}{Link}}                                                                                                    \\ \cline{2-8}
                                                                                  & TriviaQA               & QA                      & Easy                & 650,000          & real world          & \cite{chen2023autoagents}                                                                                                                                                       & \href{https://nlp.cs.washington.edu/triviaqa/}{\textcolor{blue}{Link}}                                                                                            \\ \cline{2-8}
                                                                                  & MT-bench               & QA                      & Medium              & 80               & human               & \cite{chen2023autoagents}                                                                                                                                                       & \href{https://github.com/lm-sys/FastChat/tree/main/fastchat/llm_judge}{\textcolor{blue}{Link}}                                                                    \\ \cline{2-8}
                                                                                  & RoCoBench-Text         & QA                      & Medium              & 269              & human               & \cite{mandi2024roco}                                                                                                                                                            & \href{https://github.com/MandiZhao/robot-collab}{\textcolor{blue}{Link}}                                                                                          \\ \cline{2-8}
                                                                                  & PubMedQA               & QA                      & Medium              & 273,500          & human, synthetic    & \cite{tang2023medagents}                                                                                                                                                        & \href{https://pubmedqa.github.io/}{\textcolor{blue}{Link}}                                                                                                        \\ \cline{2-8}
                                                                                  & MedQA                  & Multiple-Choice         & Medium              & 61,097           & real world          & \cite{nair2023dera,tang2023medagents}                                                                                         & \href{https://drive.google.com/file/d/1ImYUSLk9JbgHXOemfvyiDiirluZHPeQw/view}{\textcolor{blue}{Link}}                                                             \\ \cline{2-8}
                                                                                  & DDXPlus                & Medical Record          & Hard                & 1,300,000        & synthetic           & \cite{wu2023large}                                                                                                                                                              & \href{https://figshare.com/articles/dataset/DDXPlus_Dataset/20043374}{\textcolor{blue}{Link}}                                                                     \\ \cline{2-8}
                                                                                  & MedMCQA                & Multiple-Choice         & Hard                & 194,000          & real world          & \cite{tang2023medagents}                                                                                                                                                        & \href{https://medmcqa.github.io/}{\textcolor{blue}{Link}}                                                                                                         \\ \cline{2-8}
                                                                                  & MVME                   & Medical Record          & Medium              & 506              & real world          & \cite{fan2024ai}                                                                                                                                                                & \href{https://github.com/LibertFan/AI_Hospital}{\textcolor{blue}{Link}}                                                                                           \\ \cline{2-8}
                                                                                  & ARIES                  & Review Comment          & Easy                & 3,900            & human, synthetic    & \cite{d2024marg}                                                                                                                                                                & \href{https://github.com/allenai/aries}{\textcolor{blue}{Link}}                                                                                                   \\ \cline{2-8}
                                                                                  & Reviewer2              & Review                  & Easy                & 99,727           & human, synthetic    & \cite{gao2024reviewer2}                                                                                                                                                         & \href{https://huggingface.co/datasets/GitBag/Reviewer2_PGE_cleaned}{\textcolor{blue}{Link}}                                                                       \\ \cline{2-8}
                                                                                  & GSM8K                  & Math Problem            & Easy                & 8,500            & human               & \cite{du2023improving,yue2024mathvc}                                                                                          & \href{https://huggingface.co/datasets/openai/gsm8k}{\textcolor{blue}{Link}}                                                                                       \\ \cline{2-8}
                                                                                  & MGSM                   & Math Problem            & Hard                & 2,750            & human               & \cite{chen2023agentverse}                                                                                                                                                       & \href{https://github.com/google-research/url-nlp}{\textcolor{blue}{Link}}                                                                                         \\ \cline{2-8}
                                                                                  & Math                   & QA                      & Hard                & 50,000           & synthetic           & \cite{li2023camel,zhang2023exploring}                                                                                         & \href{https://huggingface.co/camel-ai}{\textcolor{blue}{Link}}                                                                                                    \\ \cline{2-8}
                                                                                  & SimuCourt              & Legal Cases             & Medium              & 420              & real world          & \cite{he2024simucourt}                                                                                                                                                          & \href{https://github.com/Zhitao-He/SimuCourt?utm_source=catalyzex.com}{\textcolor{blue}{Link}}                                                                    \\ \cline{2-8}
                                                                                  & KINLED                 & Conversational          & Medium              & 10,546           & human, synthetic    & \cite{sun2024lawluo}                                                                                                                                                            & \href{https://github.com/nefujing/lawluo}{\textcolor{blue}{Link}}                                                                                                 \\ \cline{2-8}
                                                                                  & Supreme Court Database & Legal Cases             & Easy                & 9,095            & real world          & \cite{hamilton2023blind}                                                                                                                                                        & \href{http://scdb.wustl.edu}{\textcolor{blue}{Link}}                                                                                                              \\ \cline{2-8}
                                                                                  & TDW-MAT                & Embodied Environment    & Medium              & /                & human               & \cite{zhang2023building}                                                                                                                                                        & \href{https://github.com/UMass-Foundation-Model/Co-LLM-Agents/}{\textcolor{blue}{Link}}                                                                           \\ \cline{2-8}
                                                                                  & C-WAH                  & Embodied Environment    & Medium              & /                & human               & \cite{zhang2023building}                                                                                                                                                        & \href{https://github.com/UMass-Foundation-Model/Co-LLM-Agents/}{\textcolor{blue}{Link}}                                                                           \\ \cline{2-8}
                                                                                  & RoCoBench              & Embodied Environment    & Medium              & /                & human               & \cite{mandi2024roco}                                                                                                                                                            & \href{https://github.com/MandiZhao/robot-collab}{\textcolor{blue}{Link}}                                                                                          \\ \cline{2-8}
                                                                                  & FED                    & Dialogue Response       & Medium              & 4,712            & human               & \cite{chen2023agentverse}                                                                                                                                                       & \href{http://shikib.com/fed_data.json}{\textcolor{blue}{Link}}                                                                                                    \\ \cline{2-8}
                                                                                  & CulturePark            & Conversational          & Medium              & 41,000           & synthetic           & \cite{li2024culturepark}                                                                                                                                                        & \href{https://github.com/Scarelette/CulturePark}{\textcolor{blue}{Link}}                                                                                          \\ \cline{2-8}
                                                                                  & CommonGen-Hard         & Concept                 & Easy                & 200              & human               & \cite{chen2023agentverse,yukhymenko2024synthetic}                                                                             & \href{https://github.com/madaan/self-refine}{\textcolor{blue}{Link}}                                                                                              \\ \cline{2-8}
                                                                                  & ARC Challenge          & Multiple-Choice         & Easy                & 2,590            & human               & \cite{tan2024true}                                                                                                                                                              & \href{https://huggingface.co/datasets/allenai/ai2_arc}{\textcolor{blue}{Link}}                                                                                    \\ \cline{2-8}
                                                                                  & HellaSwag              & Multiple-Choice         & Easy                & 70,000           & synthetic           & \cite{tan2024true}                                                                                                                                                              & \href{https://rowanzellers.com/hellaswag/}{\textcolor{blue}{Link}}                                                                                                \\ \cline{2-8}
                                                                                  & UCF101                 & Video Clip              & Medium              & 7,000            & human               & \cite{xie2024dreamfactory}                                                                                                                                                      & \href{https://www.crcv.ucf.edu/data/UCF101.php}{\textcolor{blue}{Link}}                                                                                           \\ \cline{2-8}
                                                                                  & HMDB51                 & Video Clip              & Medium              & 13,320           & real world          & \cite{xie2024dreamfactory}                                                                                                                                                      & \href{https://serre-lab.clps.brown.edu/resource/hmdb-a-large-human-motion-database/}{\textcolor{blue}{Link}}                                                      \\ \bottomrule

\end{tabular}
}
\vspace{-3mm}
\caption{Summary of commonly used datasets for scenario simulation.}
\label{tab:task_data}
\end{table*}

\subsection{Social Simulation}
We summarize commonly used datasets or benchmarks for social simulations in Table~\ref{tab:soc_data}. In social simulations, datasets often consist of two parts: those for initialization of agents and those for evaluation. Data used for agent initialization typically contain profiles and potential relations between agents, to help initialize the simulation settings. In contrast, datasets for evaluation provide the reference data of behaviors of real-world individuals. These datasets are sourced in various ways, such as public surveys, existing datasets like MovieLens and Amazon-Book, and crawling from online platforms like Twitter. 

\begin{table*}[!htp]
\setlength{\tabcolsep}{2pt} 
\centering
\resizebox{0.9\textwidth}{!}{
\begin{tabular}{c|c|c|c|c|c|c|c|c}
\toprule
\textbf{Scenario}                                                                             & \textbf{Dataset}             & \textbf{Init.} & \textbf{Eval.} & \textbf{Content}                                                         & \textbf{\# case}          & \textbf{Simulation Objectives}                                        & \textbf{Used by}                                                                & \textbf{Data Link}                                                                                                            \\ \midrule
\multirow{7}{*}{\begin{tabular}[c]{@{}c@{}}General\\ Economics\end{tabular}}                  & 2018 U.S. population         & \checkmark              &                     & profile                                                                  & 100 people             & macroeconomic activities                                              & ~\cite{li2024econagent}                                                         & \href{https://github.com/automoto/big-five-data}{\textcolor{blue}{Link}}                                                      \\ \cline{2-9} 
                                                                                              & public government data       &                         & \checkmark          & rent information                                                         & 51 users               & resource allocation                                                   & ~\cite{ji2024srap}                                                              & \href{https://www.bphc.com.cn/home}{\textcolor{blue}{Link}}                                                                   \\ \cline{2-9} 
                                                                                              & names-dataset 3.1.0          & \checkmark              &                     & profile                                                                  & 1,000 people           & epidemic modeling                                                     & ~\cite{williams2023epidemic}                                                    & \href{https://github.com/philipperemy/name-dataset}{\textcolor{blue}{Link}}                                                   \\ \cline{2-9} 
                                                                                              & big-five-data                & \checkmark              &                     & profile                                                                  & 1,000 people           & epidemic modeling                                                     & ~\cite{williams2023epidemic}                                                    & \href{https://github.com/automoto/big-five-data}{\textcolor{blue}{Link}}                                                      \\ \cline{2-9} 
                                                                                              & American Community Survey    & \checkmark              &                     & profile                                                                  & 8.4M people            & epidemic modeling                                                     & ~\cite{chopra2024limits}                                                        & \href{https://www.nyc.gov/site/planning/planning-level/nyc-population/american-community-survey.page}{\textcolor{blue}{Link}} \\ \cline{2-9} 
                                                                                              & Bureau of Labor Statistics   &                         & \checkmark          & labor statistics                                                         & 8.4M people            & unemployment rate                                                     & ~\cite{chopra2024limits}                                                        & \href{https://www.bls.gov/charts/employment-situation/civilian-labor-force-participation-rate.htm}{\textcolor{blue}{Link}}    \\ \cline{2-9} 
                                                                                              & CDC                          &                         & \checkmark          & infection rate                                                           & 8.4M people            & epidemic modeling                                                     & ~\cite{chopra2024limits}                                                        & \href{https://www.cdc.gov/}{\textcolor{blue}{Link}}                                                                           \\ \hline
\multirow{16}{*}{\begin{tabular}[c]{@{}c@{}}Sociology\\ and\\ Political Science\end{tabular}} & ANES                         & \checkmark              & \checkmark          & profile,answer                                                           & 15,626 responses       & voting                                                                & ~\cite{Argyle_2023,zhang2024electionsim,sun2024randomsiliconsamplingsimulating} & \href{https://electionstudies.org/about-us/}{\textcolor{blue}{Link}}                                                          \\ \cline{2-9} 
                                                                                              & Pigeonholing Partisans       & \checkmark              & \checkmark          & profile,answer                                                           & 2,107 responses        & partisan bias                                                         & ~\cite{Argyle_2023}                                                             & \href{https://dataverse.harvard.edu/dataset.xhtml?persistentId=doi:10.7910/DVN/U23L09   }{\textcolor{blue}{Link}}             \\ \cline{2-9} 
                                                                                              & Global Warming               & \checkmark              & \checkmark          & profile,answer                                                           & 2,310 responses        & opinion                                                               & ~\cite{lee2023can}                                                              & /                                                                                                                             \\ \cline{2-9} 
                                                                                              & Twitter                      & \checkmark              &                     & statements                                                               & 1,006,517 users        & voting                                                                & ~\cite{zhang2024electionsim}                                                    & /                                                                                                                             \\ \cline{2-9} 
                                                                                              & Interview                    & \checkmark              & \checkmark          & profile,answer                                                           & 1,002 users            & opnion and behavior                                                   & ~\cite{park2024generative}                                                      & \href{https://github.com/joonspk-research/genagents}{\textcolor{blue}{Link}}                                                  \\ \cline{2-9} 
                                                                                              & Name                         & \checkmark              &                     & name                                                                     & 500 names              & /                                                                     & ~\cite{aher2023using}                                                           & \href{https://github.com/microsoft/turing-experiments/tree/main}{\textcolor{blue}{Link}}                                      \\ \cline{2-9} 
                                                                                              & Ultimatum Game               &                         & \checkmark          & money allocation                                                         & 10,000 pairs & money allocation                                                      & ~\cite{aher2023using}                                                           & \href{https://github.com/microsoft/turing-experiments/tree/main}{\textcolor{blue}{Link}}                                      \\ \cline{2-9} 
                                                                                              & Garden Path Sentences        &                         & \checkmark          & garden path sentences                                                    & 96 sentences           & language parsing                                                      & ~\cite{aher2023using}                                                           & \href{https://github.com/microsoft/turing-experiments/tree/main}{\textcolor{blue}{Link}}                                      \\ \cline{2-9} 
                                                                                              & Wisdom of Crowds             &                         & \checkmark          & answers to questions                                                     & 15,000 answers         & wisdom of crowds                                                      & ~\cite{aher2023using}                                                           & \href{https://github.com/microsoft/turing-experiments/tree/main}{\textcolor{blue}{Link}}                                      \\ \cline{2-9} 
                                                                                              & Milgram Shock Experiment     &                         & \checkmark          & behavior records                                                         & 100 people             & obedience behavior                                                    & ~\cite{aher2023using}                                                           & \href{https://github.com/microsoft/turing-experiments/tree/main}{\textcolor{blue}{Link}}                                      \\ \cline{2-9} 
                                                                                              & 15 Topics                    & \checkmark              &                     & profile, opinion                                                         & 10 users               & opinion dynamics                                                      & ~\cite{chuang2023simulating}                                                    & \href{https://github.com/yunshiuan/llm-agent-opinion-dynamics}{\textcolor{blue}{Link}}                                        \\ \cline{2-9} 
                                                                                              & Formative Study              & \checkmark              & \checkmark          & profile, interview                                                       & 14 users               & \begin{tabular}[c]{@{}c@{}}information\\ management\end{tabular}      & ~\cite{park2023choicemates}                                                     & /                                                                                                                             \\ \cline{2-9} 
                                                                                              & User Study                   & \checkmark              & \checkmark          & profile, interview                                                       & 36 users               & \begin{tabular}[c]{@{}c@{}}information\\ management\end{tabular}      & ~\cite{park2023choicemates}                                                     & /                                                                                                                             \\ \cline{2-9} 
                                                                                              & collective decision-making   & \checkmark              & \checkmark          & profile, opinion                                                         & 2,290 users            & \begin{tabular}[c]{@{}c@{}}collective \\ decision-making\end{tabular} & ~\cite{jarrett2023language}                                                     & /                                                                                                                             \\ \cline{2-9} 
                                                                                              & Becker-2019                  & \checkmark              & \checkmark          & profile, answers                                                         & 1,120 users            & wisdom of crowds                                                      & ~\cite{chuang2024wisdom}                                                        & \href{https://github.com/joshua-a-becker/wisdom-of-partisan-crowds   }{\textcolor{blue}{Link}}                                \\ \cline{2-9} 
                                                                                              & Controversial Beliefs Survey & \checkmark              & \checkmark          & profile, opinion                                                         & 564 users              & opinion                                                               & ~\cite{chuang2024beyond}                                                        & /                                                                                                                             \\ \hline
\multirow{20}{*}{\begin{tabular}[c]{@{}c@{}}Online\\ Platforms\end{tabular}}                  & FPS                          & \checkmark              &                     & /                                                                        & 6 topics               & opinion dynamics                                                      & ~\cite{liu2024skepticism}                                                       & /                                                                                                                             \\ \cline{2-9} 
                                                                                              & Echo Chambers                & \checkmark              &                     & profile                                                                  & 3 networks             & opinion polarization                                                  & ~\cite{wang2024decoding}                                                        & /                                                                                                                             \\ \cline{2-9} 
                                                                                              & Gender Discrimination        & \checkmark              & \checkmark          & profile, opinion                                                         & 8,563 users            & opinion dynamics                                                      & ~\cite{gao2023s3}                                                               & /                                                                                                                             \\ \cline{2-9} 
                                                                                              & Nuclear Energy               & \checkmark              & \checkmark          & profile, opinion                                                         & 17,945 users           & opinion dynamics                                                      & ~\cite{gao2023s3}                                                               & /                                                                                                                             \\ \cline{2-9} 
                                                                                              & ANES                         & \checkmark              &                     & profile                                                                  & 500 users              & partisan bias                                                         & ~\cite{tornberg2023simulating}                                                  & \href{https://electionstudies.org/about-us/}{\textcolor{blue}{Link}}                                                          \\ \cline{2-9} 
                                                                                              & SAGraph                      & \checkmark              & \checkmark          & profile, interaction                                                     & 40~300 influencers     & influencer selection                                                  & ~\cite{zhang2024large}                                                          & /                                                                                                                             \\ \cline{2-9} 
                                                                                              & Metoo                        & \checkmark              & \checkmark          & profile, opinion                                                         & 1,000 users            & opinion dynamics                                                      & ~\cite{mou2024unveiling}                                                        & \href{https://drive.google.com/file/d/1qQzQAvDH-eLtg1jPTKe6NkToF7Aq1EAA/view?usp=sharing}{\textcolor{blue}{Link}}             \\ \cline{2-9} 
                                                                                              & Roe                          & \checkmark              & \checkmark          & profile, opinion                                                         & 1,000 users            & opinion dynamics                                                      & ~\cite{mou2024unveiling}                                                        & \href{https://drive.google.com/file/d/13dkJ_P2JzbrDdJkYdwred260Ps-ym-64/view?usp=sharing}{\textcolor{blue}{Link}}             \\ \cline{2-9} 
                                                                                              & BLM                          & \checkmark              & \checkmark          & profile, opinion                                                         & 1,000 users            & opinion dynamics                                                      & ~\cite{mou2024unveiling}                                                        & \href{https://drive.google.com/file/d/1HymVETg5SgLJqL1O3bPiT-RcBVSMGEhT/view?usp=sharing}{\textcolor{blue}{Link}}             \\ \cline{2-9} 
                                                                                              & Twitter15                    & \checkmark              & \checkmark          & profile, behavior                                                        & 198 news               & rumor propagation                                                     & ~\cite{yang2024oasis}                                                           & \href{https://www.dropbox.com/s/7ewzdrbelpmrnxu/rumdetect2017.zip?dl=0}{\textcolor{blue}{Link}}                               \\ \cline{2-9} 
                                                                                              & Twitter16                    & \checkmark              & \checkmark          & profile, behavior                                                        & 198 news               & rumor propagation                                                     & ~\cite{yang2024oasis}                                                           & \href{https://www.dropbox.com/s/7ewzdrbelpmrnxu/rumdetect2017.zip?dl=0}{\textcolor{blue}{Link}}                               \\ \cline{2-9} 
                                                                                              & Reddit                       & \checkmark              & \checkmark          & profile, comment                                                         & 116,932 comments       & herd effect                                                           & ~\cite{yang2024oasis}                                                           & /                                                                                                                             \\ \cline{2-9} 
                                                                                              & MindEcho                     & \checkmark              & \checkmark          & profile, comment                                                         & 14 KOL                 & key opinion leader                                                    & ~\cite{xu2024mindechoroleplayinglanguageagents}                                 & /                                                                                                                             \\ \cline{2-9} 
                                                                                              & WARRIORS                     & \checkmark              & \checkmark          & \begin{tabular}[c]{@{}c@{}}profile,\\ search behavior\end{tabular}       & 100,000 users          & search behavior                                                       & ~\cite{ren2024bases}                                                            & /                                                                                                                             \\ \cline{2-9} 
                                                                                              & Amazon Beauty                & \checkmark              & \checkmark          & \begin{tabular}[c]{@{}c@{}}profile,\\ user-item interaction\end{tabular} & 15,577 users           & \begin{tabular}[c]{@{}c@{}}user-item\\ interaction\end{tabular}       & ~\cite{huang2023recommender}                                                    & \href{https://jmcauley.ucsd.edu/data/amazon/links.html}{\textcolor{blue}{Link}}                                               \\ \cline{2-9} 
                                                                                              & Steam                        & \checkmark              & \checkmark          & \begin{tabular}[c]{@{}c@{}}profile,\\ user-item interaction\end{tabular} & 281,205 users          & \begin{tabular}[c]{@{}c@{}}user-item\\ interaction\end{tabular}       & ~\cite{huang2023recommender,Zhang2024on}                                        & \href{https://github.com/LehengTHU/Agent4Rec}{\textcolor{blue}{Link}}                                                         \\ \cline{2-9} 
                                                                                              & MovieLens                    & \checkmark              & \checkmark          & \begin{tabular}[c]{@{}c@{}}profile,\\ user-item interaction\end{tabular} & 298,074 users          & \begin{tabular}[c]{@{}c@{}}user-item\\ interaction\end{tabular}       & ~\cite{huang2023recommender,wang2023recagent,Zhang2024on}                       & \href{https://github.com/LehengTHU/Agent4Rec}{\textcolor{blue}{Link}}                                                         \\ \cline{2-9} 
                                                                                              & Amazon Book                  & \checkmark              & \checkmark          & \begin{tabular}[c]{@{}c@{}}profile,\\ user-item interaction\end{tabular} & /                      & \begin{tabular}[c]{@{}c@{}}user-item\\ interaction\end{tabular}       & ~\cite{Zhang2024on}                                                             & \href{https://github.com/LehengTHU/Agent4Rec}{\textcolor{blue}{Link}}                                                         \\ \cline{2-9} 
                                                                                              & Amazon Review CD             & \checkmark              & \checkmark          & \begin{tabular}[c]{@{}c@{}}profile,\\ user-item interaction\end{tabular} & 100 users              & \begin{tabular}[c]{@{}c@{}}user-item\\ interaction\end{tabular}       & ~\cite{zhang2024agentcf}                                                        & \href{https://amazon-reviews-2023.github.io/main.html}{\textcolor{blue}{Link}}                                                \\ \cline{2-9} 
                                                                                              & Amazon Review Office         & \checkmark              & \checkmark          & \begin{tabular}[c]{@{}c@{}}profile,\\ user-item interaction\end{tabular} & 100 users              & \begin{tabular}[c]{@{}c@{}}user-item\\ interaction\end{tabular}       & ~\cite{zhang2024agentcf}                                                        & \href{https://amazon-reviews-2023.github.io/main.html}{\textcolor{blue}{Link}}                                                \\ \bottomrule
\end{tabular}
}
\vspace{-3mm}
\caption{Summary of commonly used datasets for society simulation. \textit{Init.} means the data provides profile to initialize agents, and \textit{Eval.} means it provides data to validate the simulation effectiveness.}
\label{tab:soc_data}
\end{table*}

\section{Trend of Social Simulations}~\label{sec:trend}

\subsection{Trend of Individual Simulation}

\begin{figure*}[t]
    \centering
    \includegraphics[width=\linewidth]{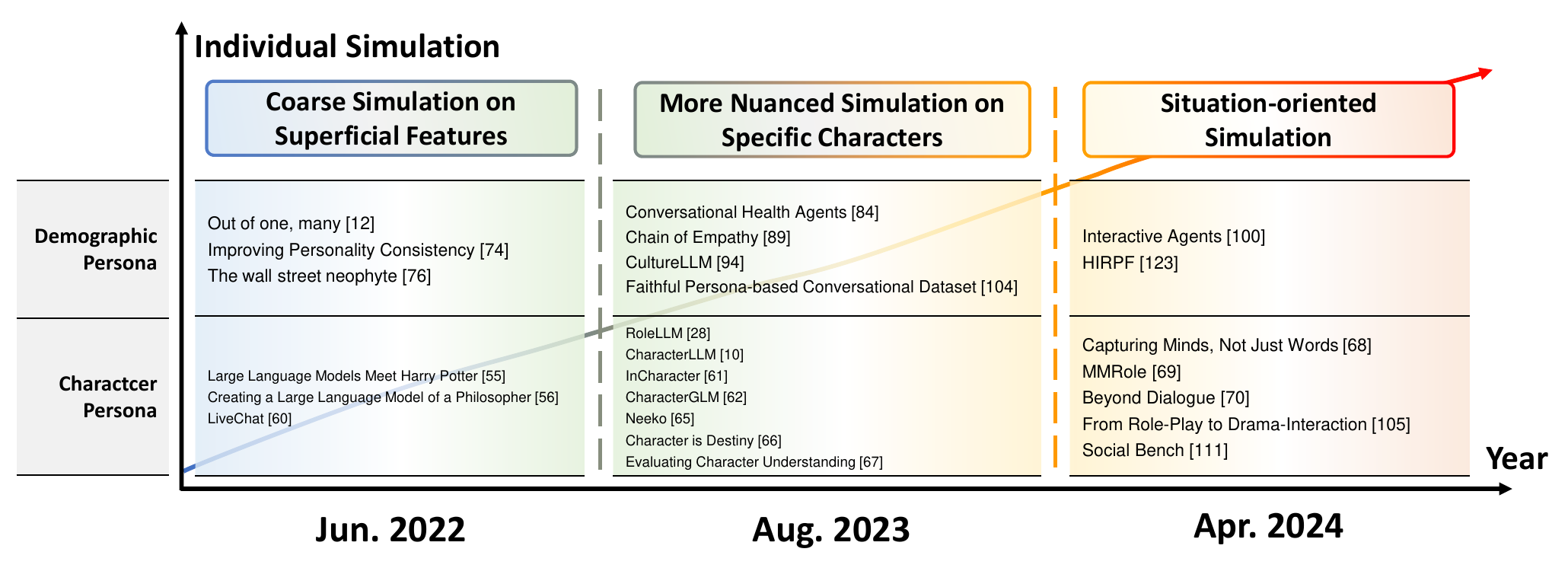}
    \caption{Illustration of individual simulation trend, which goes through {\ul coarse simulation}, {\ul more nuanced simulation}, and {\ul situation-oriented simulation}.}
    \label{fig:indi_trend}
\end{figure*}

\begin{figure*}[t]
    \centering
    \includegraphics[trim=8mm 0mm 0mm 0mm, width=\linewidth]{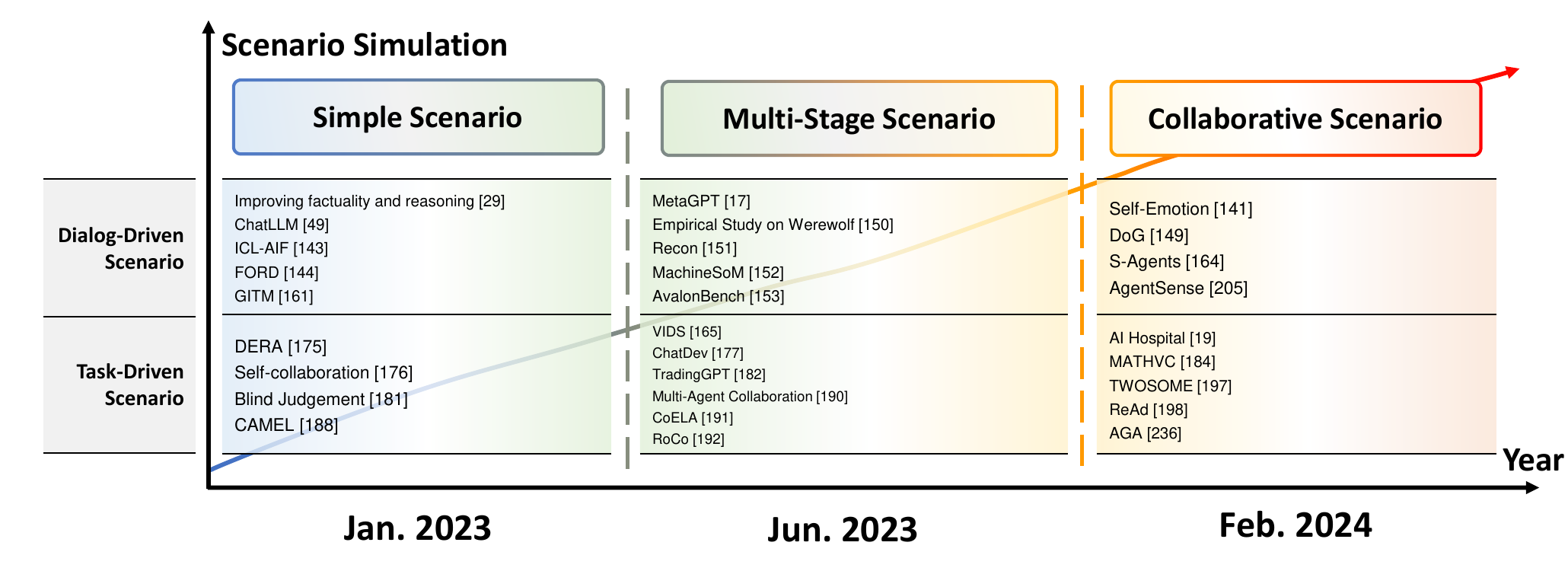}
    \caption{Illustration of scenario simulation trend, which goes through {\ul simple scenario}, {\ul multi-stage scenario}, and {\ul collaborative scenario}.}
    \label{fig:task_trend}
\end{figure*}

\begin{figure*}[t]
    \centering
    \includegraphics[width=\linewidth]{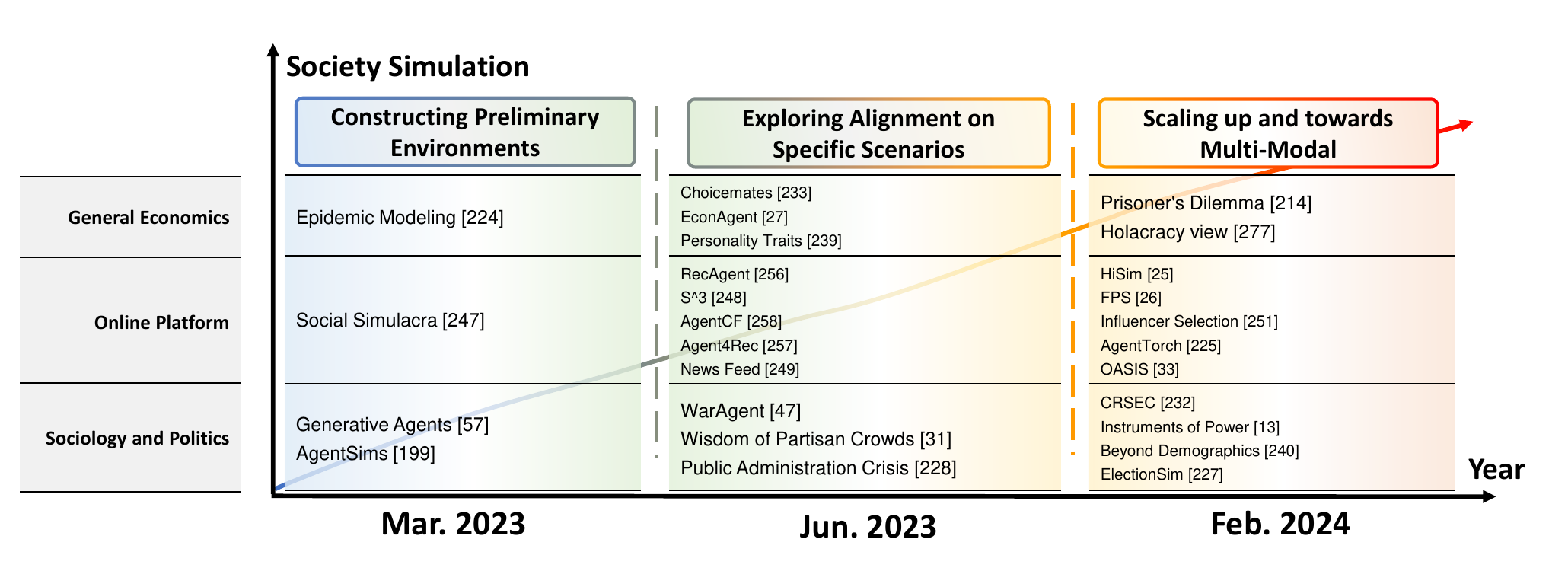}
    \caption{Illustration of society simulation trend, which goes through three stages: {\ul constructing preliminary environments}, {\ul exploring alignment on specific scenarios}, and {\ul scaling up} while moving {\ul towards multi-modal}.}
    \label{fig:soc_trend}
\end{figure*}

Evolving from social science, individual simulation powered by LLMs has progressed through three distinct stages, namely \textbf{coarse simulation}, \textbf{more nuanced simulation}, and \textbf{situation-oriented simulation}, which is depicted in Figure~\ref{fig:indi_trend}. Since \textit{{\ul June 2022}}, researchers started to focus on coarse simulations, especially for superficial traits like testing the personalities of LLMs and simulating well-known characters~\cite{serapiogarcía2023personalitytraitslargelanguage,liu2023agentbenchevaluatingllmsagents}. After \textit{{\ul August 2023}}, the trends shifted towards more refined simulations of specific individuals, with studies evaluating the cognitive aspects of simulated models~\cite{wang2024incharacterevaluatingpersonalityfidelity,yuan2024evaluatingcharacterunderstandinglarge} and improving their simulation capabilities~\cite{abbasian2024conversationalhealthagentspersonalized,yu2024neekoleveragingdynamiclora}. By \textit{{\ul May 2024}}, researchers began conducting individual simulations in specific scenarios~\cite{chen2024socialbenchsocialityevaluationroleplaying,yu2024dialogueprofiledialoguealignmentframework}, further expanding the complexity and realism of these simulations.

\subsubsection{Coarse Simulation on Superficial Features}
Many individual simulation works born since June 2022, the majority of which initially focus on simulating superficial features implied in human behaviors. A significant portion of the effort was dedicated to collecting and standardizing character-related information to build persona-based datasets~\cite{chen2023largelanguagemodelsmeet,schwitzgebel2023creatinglargelanguagemodel}. Additionally, eliciting the underlying demographic personalities of prevailing LLMs posed a challenge in this early stage~\cite{serapiogarcía2023personalitytraitslargelanguage,pan2023llmspossesspersonalitymaking}. The early trials on coarse individual simulations shed light on LLMs' attributes during simulation, including hallucinations, inherent biases, and stereotypes, which are proven to be crucial for future simulations.

\subsubsection{More Nuanced Simulation on Specific Characters}
As individual simulation methods advanced, the precision of simulations significantly improved. 
More nuanced aspects of the individual simulation gained growing attention. Some works implement new functionalities and refine the models' architecture, such as incorporating memory and planning modules~\cite{abbasian2024conversationalhealthagentspersonalized,xu2024characterdestinylargelanguage}, while others focus on designing specific tasks for training and evaluation, like multi-dimensional interviews~\cite{wang2024incharacterevaluatingpersonalityfidelity} and simulation with rich information from scene descriptions and experiential memories~\cite{wang2024rolellmbenchmarkingelicitingenhancing}.


\subsubsection{Situation-Oriented Simulation}
Situation-oriented individual simulations begin within game environments~\cite{light2023avalonbenchevaluatingllmsplaying}, where LLMs are required to make appropriate decisions based on predefined rules. In more complex environments, simulated individuals are supposed to interact dynamically with their surroundings, responding to real-time environmental feedback~\cite{chen2024socialbenchsocialityevaluationroleplaying,qiu2024interactiveagentssimulatingcounselorclient}. Beyond traditional simulations like dialogue, situation-oriented simulations expand into areas such as dramatic performances~\cite{wu2024roleplaydramainteractionllmsolution}, digital game exploration~\cite{wang2023voyageropenendedembodiedagent}, and 3D task execution~\cite{huang2024embodiedgeneralistagent3d}. As the complexity of these simulations grows, the demands on the underlying architecture grow as well.


\subsection{Trend of Scenario Simulation}
The development of scenario simulation has progressed through several distinct stages. 
Starting from \textit{{\ul January 2023}}, different researches focused primarily on simple scenarios concerning single objectives and facilitated basic contextual interactions~\cite{nair2023dera,li2023camel,hamilton2023blind,xiong2023examining}.
By \textit{{\ul June 2023}}, the emphasis changed to multi-stage scenarios, incorporating multi-step tasks that enabled agents to engage in sequential decision-making and adaptive responses across varied contexts to achieve the more complex goal~\cite{talebirad2023multi,mandi2024roco,li2023tradinggpt,hassan2023chatgpt}.
By \textit{{\ul February 2024}}, research has increasingly focused on multi-agent collaborative scenarios, emphasizing agents' capabilities to cooperate and adapt within complex, high-order simulations~\cite{yu2024affordable,chen2024s,ma2024debate,yue2024mathvc}.

\subsubsection{Simple Scenario}
In the initial phase of scenario simulation, researchers focused on constructing simple scenarios that supported foundational agent interactions. Much of this work concentrated on dialogue-driven decision-making frameworks, which facilitated structured information exchange and agent alignment~\cite{nair2023dera,li2023camel,hao2023chatllm}. Additionally, studies explored the collaborative potentials of agents through multi-agent debate frameworks, employing debate and critical feedback to assess cooperative reasoning and performance enhancement in LLMs~\cite{fu2023improving,xiong2023examining,du2023improving}. Simultaneously, other studies applied scenario simulations within specific domains—such as law, software development, scientific analysis, and recommendation systems—demonstrating the versatility of task-based simulations in achieving domain-specific objectives~\cite{hamilton2023blind,dong2023self,zhu2023ghost}.

\subsubsection{Multi-Stage Scenario}
Different from simple task-oriented scenarios, multi-stage scenarios are no longer limited to mere agent interactions. Instead, they emphasize the fine-grained construction of scenarios. 
This stage introduces multiple roles and task decomposition as central elements, enabling agents to collaborate not merely on single tasks but through incremental task breakdowns that require coordinated effort~\cite{zhang2023building,mandi2024roco}. 
In software development, ~\cite{qian2024chatdev,hong2023metagpt} decomposed the development process into multiple stages like design, coding and testing to enhance the capacity for achieving complex objectives and improving software quality.
Additionally, communication games were introduced to investigate human behavior within complex conversational scenarios, adding depth to interaction analysis~\cite{xu2023exploring,wang2023avalon,zhang2023exploring,light2023text}.

\subsubsection{Collaborative Scenario}
With the growing interest in scenario simulation, research shifted toward collaborative scenarios, emphasizing advanced social dynamics and cooperative strategies in agent interactions. \cite{tan2024true,zhang2024towards} introduce reinforcement learning to align LLM with embodied environments. 
To build efficient scenario simulations, \cite{yu2024affordable} focused on reducing LLM inference costs by modeling social relationships while \cite{chen2024s} utilized dynamic ``agent trees'' in environments like Minecraft, enabling asynchronous task execution for efficient resource gathering. 
In addition, \cite{fan2024ai,yan2024social} simulated collaborative environments in the real world, reflecting complex social interactions such as medical processes and the development of social skills, with agents handling evolving multistep tasks.

\subsection{Trend of Society Simulation}
Since the concept of social simulation was first introduced by Park et al. \cite{park2022social}, numerous notable studies have emerged. Broadly, the development of this field can be categorized into three phases. Prior to \textit{{\ul June 2023}}, researchers concentrated on constructing preliminary environments \cite{park2023generative,williams2023epidemic,lin2023agentsims}. By \textit{{\ul February 2024}}, the focus shifted toward exploring alignment within specific scenarios, such as persona modeling and targeted environments, marking the first significant surge of publications \cite{gao2023s3,wang2023avalons,li2024econagent}. Most recently, the trend has moved towards scaling up and incorporating multi-modal approaches. In this phase, large-scale precise modeling has gained recognition, with other modalities such as vision and voice being integrated into simulations \cite{mou2024unveiling,ren2024emergence,wu2024enhance,wang2024grutopia}.

The main characteristics can be summarized as:
\subsubsection{Constructing Preliminary Environments}
The complexity of society simulation, to a certain extent, stems from the complexity of the environment involved. Society simulation usually involve multiple interacting individuals (such as people, organizations, groups, etc.), which act in a specific environment (such as cities, markets, cyberspace, etc.). Therefore, the pioneer work focuses on how to design a specific environment to support society simulation. \cite{park2023generative} built an interactive sandbox environment by extending a LLM to store a complete record of an agent's experience and dynamically synthesizing memory to plan behavior. \cite{williams2023epidemic} built an epidemic spread simulation environment that simulates human behavior at the individual level to reproduce the spread of an epidemic in a simulated environment. \cite{lin2023agentsims} created an easy-to-use infrastructure that allows researchers to build evaluation tasks by adding agents and buildings, providing a visual and program-based platform for testing LLMs.

\subsubsection{Exploring Alignment on Specific Scenarios}
With the development of simulation environment technology, society simulation has basically become operational. At this time, to test the credibility of simulation, evaluating the alignment performance of agents with real situations on specific tasks has gradually become an important research direction. \cite{gao2023s3} use real social network data to measure the accuracy of simulation by evaluating the behavior and decision-making of agents at the individual and group levels in a simulated social network environment. \cite{li2024econagent} evaluate the decision rationality of LLM agents by simulating macroeconomic activities and comparing the performance of LLM agents with traditional rule-based agents or language agents in generating classic macroeconomic phenomena such as inflation and unemployment.

\subsubsection{Scaling Up and towards Multi-Modal}
\paragraph{Scaling up} Before LLM-based agents became widely adopted for society simulation, researchers predominantly relied on agent-based modeling (ABM) methods, where agents were typically programmed to react based on predefined algorithms. With the advent of LLM providing glimpses of human-like intelligence \cite{browning2023personhood}, LLM-based agents entered the spotlight. Given the good performance of LLM-based agents in a series of specific scenarios, researchers began to expand the scale of simulation. \cite{mou2024unveiling, ren2024emergence} involve the core elements of large-scale society simulation and study the interaction between agents and the generation of behavioral norms. \cite{wu2024enhance} proposed a proving ground for assessing advanced reasoning capabilities of LLM agents in a large-scale society simulation context.

\paragraph{Multi-Modal} With the development of language models, using language agents for society simulation has become a hot topic in research. It incorporates other modal information elements such as vision in life into the simulation through text descriptions. However, with a series of advances in the field of Vision-Language Model(VLM)~\cite{Radford2021, liu2024visual, achiam2023gpt}, researchers began to incorporate VLM-based agents into society simulation research. \cite{wang2024grutopia} provide rich multi-modal interaction information and detailed annotations in large-scale scenarios. \cite{yu2024mineland} focus on simulating the perceptual limitations and physical demands of the real world to facilitate more realistic social interactions.

\section{Conclusion}~\label{sec:con}
In this paper, we categorize LLM-driven social simulations into three types: individual, scenario, and society simulation, highlighting their progression from modeling individual behaviors to replicating complex social dynamics. By systematically reviewing architectures, methods, and evaluations across these categories, we provide a structured framework for advancing research in this field. This work aims to guide the development of LLM-based simulations and foster interdisciplinary studies to address real-world challenges and support decision-making.

\bibliographystyle{unsrt}
\bibliography{ijcai24}

\end{document}